%% file: main.tex
\documentclass{article}

\makeatletter
\def\@fnsymbol#1{\ensuremath{\ifcase#1\or \dagger\or \ddagger\or
\mathsection\or \mathparagraph\or \|\or **\or \dagger\dagger
\or \ddagger\ddagger \else\@ctrerr\fi}}
\makeatother
    
\usepackage{times}

\newcommand\iclr{iclr}
\newcommand\metalearn{metalearn}
\newcommand\drl{drl}

\newcommand\venue{iclr}

\ifx\venue\iclr
    \usepackage{iclr2022_conference}
    \iclrfinalcopy
\fi
\ifx\venue\metalearn
    \usepackage[final]{metalearn_workshop}
\fi
\ifx\venue\drl
    \usepackage{deeprl_workshop}
    \iclrfinalcopy
\fi

\usepackage[utf8]{inputenc} %
\usepackage[T1]{fontenc}    %
\usepackage[colorlinks=True, citecolor=black, linkcolor=red]{hyperref}       %
\usepackage{url}            %
\usepackage{booktabs}       %
\usepackage{amsfonts}       %
\usepackage{nicefrac}       %
\usepackage{microtype}      %

\usepackage{microtype}
\usepackage{graphicx}
\usepackage[utf8]{inputenc} %
\usepackage[T1]{fontenc}    %
\usepackage{url}            %
\usepackage{booktabs}       %
\usepackage{amsfonts}       %
\usepackage{nicefrac}       %
\usepackage{color}
\usepackage{mathtools}
\usepackage{amsmath,amssymb}
\usepackage{bm}
\usepackage{siunitx}
\usepackage{wrapfig}
\usepackage{algorithm}
\usepackage[noend]{algorithmic}
\usepackage{lipsum}
\usepackage{caption}
\usepackage{upgreek}
\usepackage[small]{subfigure}

\usepackage[nottoc]{tocbibind}
\usepackage{minitoc}

\input{math_commands.tex}

\input{macro.tex}

\title{Skill-based Meta-Reinforcement Learning}

\input{tex/author}

\begin{document}

\doparttoc %
\faketableofcontents %

\maketitle

\input{sections/00_abstract}
\input{sections/01_introduction}

\input{sections/02_related_work}

\input{sections/03_prelim}

\ifx\venu\iclr
    \input{sections/04_approach}
\else
    \input{sections_metalearn/04_approach}
\fi

\ifx\venue\metalearn
    \input{sections_metalearn/05_experiments}
\else
    \input{sections/05_experiments}
\fi
\input{sections/06_conclusion}

\clearpage

\input{sections/07_ack}

\renewcommand{\bibname}{References}
\bibliography{main}
\bibliographystyle{iclr2022_conference}

\clearpage

\input{sections/appendix.tex}

\end{document}

%% file: math_commands.tex
\usepackage{amsmath,amsfonts,bm}

\def\eqref#1{equation~\ref{#1}}

\def\1{\bm{1}}

\DeclareMathAlphabet{\mathsfit}{\encodingdefault}{\sfdefault}{m}{sl}
\SetMathAlphabet{\mathsfit}{bold}{\encodingdefault}{\sfdefault}{bx}{n}

\DeclarePairedDelimiterX{\norm}[1]{\lVert}{\rVert}{#1}

%% file: macro.tex
\newcommand{\sun}[1]{{\color{blue}{\small\bf\sf [Sun: #1]}}}

\newcommand{\Skip}[1]{}

\newcommand{\vspacesection}[1]{\vspace{-0.05cm}
\section{#1}
\vspace{-0.05cm}}
\newcommand{\vspacesubsection}[1]{\vspace{-0.05cm}
\subsection{#1}
\vspace{-0.05cm}}
\newcommand{\vspacesubsubsection}[1]{\vspace{-0.05cm}
\subsubsection{#1}
\vspace{-0.05cm}}

\newcommand{\method}[1]{SiMPL}

\newcommand{\ie}{\textit{i}.\textit{e}.\ }
\newcommand{\eg}{\textit{e}.\textit{g}.\ }

\newcommand{\myfig}[1]{Figure~\ref{#1}}

\newcommand{\mysecref}[1]{Section~\ref{#1}}

\newcommand{\dotieconcat}[2]{%
  \text{\raisebox{.8ex}{$\smallfrown$}}%
}

\makeatletter
\newcommand\dslfontsize{\@setfontsize\dslfontsize\@viipt\@viiipt}
\makeatother

\newcommand\blfootnote[1]{%
  \begingroup
  \renewcommand\thefootnote{}\footnote{#1}%
  \addtocounter{footnote}{-1}%
  \endgroup
}

\usepackage{color}
\usepackage{xcolor}
\definecolor{codegray}{rgb}{0.5,0.5,0.5}

\usepackage{enumitem}
\setlist[itemize]{leftmargin=10mm} %
\newenvironment{compactwrapfigure}
     {\setlength{\intextsep}{.4\intextsep}%
      \wrapfloat{figure}}
     {\endwrapfloat}

%% file: tex/author.tex
\newcommand{\kaist}[0]{\textsuperscript{1}}
\newcommand{\usc}[0]{\textsuperscript{2}}
\newcommand{\aitrics}[0]{\textsuperscript{3}}
\newcommand{\naver}[0]{\textsuperscript{4}}

\author{Taewook Nam\kaist, Shao-Hua Sun\usc, Karl Pertsch\usc, Sung Ju Hwang\kaist\textsuperscript{,}\aitrics, Joseph J. Lim\kaist\thanks{Work done while at USC}\hspace{3.5pt}\textsuperscript{,}\hspace{0pt}\naver\thanks{AI advisor at Naver AI Lab} \\
Korea Advanced Institute of Science and  Technology\kaist\\
University of Southern California\usc, AITRICS\aitrics, Naver AI Lab\naver \\
\texttt{namsan@kaist.ac.kr,\{shaohuas,pertsch\}@usc.edu,
}\\
\texttt{sjhwang82@kaist.ac.kr,joe.lim@kaist.ac.kr}
}

%% file: sections/00_abstract.tex
\begin{abstract}
While deep reinforcement learning methods have shown impressive results in robot learning, their sample inefficiency makes the learning of complex, long-horizon behaviors with real robot systems infeasible. To mitigate this issue, meta-reinforcement learning methods aim to enable fast learning on novel tasks by learning how to learn. Yet, the application has been limited to short-horizon tasks with dense rewards. To enable learning long-horizon behaviors, recent works have explored leveraging prior experience in the form of offline datasets without reward or task annotations. While these approaches yield improved sample efficiency, millions of interactions with environments are still required to solve complex tasks. In this work, we devise a method that enables meta-learning on long-horizon, sparse-reward tasks, allowing us to solve unseen target tasks with orders of magnitude fewer environment interactions. Our core idea is to leverage prior experience extracted from offline datasets during meta-learning. Specifically, we propose to (1) extract reusable skills and a skill prior from offline datasets, (2) meta-train a high-level policy that learns to efficiently compose learned skills into long-horizon behaviors, and (3) rapidly adapt the meta-trained policy to solve an unseen target task. Experimental results on continuous control tasks in navigation and manipulation demonstrate that the proposed method can efficiently solve long-horizon novel target tasks by combining the strengths of meta-learning and the usage of offline datasets, while prior approaches in RL, meta-RL, and multi-task RL require substantially more environment interactions to solve the tasks.
\blfootnote{
Project page: \url{https://namsan96.github.io/SiMPL}
}
\end{abstract}

%% file: sections/01_introduction.tex
\vspacesection{Introduction}
\label{sec:intro}

In recent years, deep reinforcement learning methods have achieved impressive results in robot learning~\citep{gu2017deep, openai2018learning,kalashnikov2021mt}. 
Yet, %
existing approaches are sample inefficient, 
thus rendering the learning of complex behaviors through trial and error learning infeasible, especially on real robot systems. 
In contrast, humans are capable of effectively learning a variety of complex skills in only a few trials. 
This can be greatly attributed to our ability to 
learn how to learn new tasks quickly by
efficiently utilizing previously acquired skills.

\input{fig/teaser}

Can machines likewise learn to how to learn by efficiently utilizing learned skills like humans?
Meta-reinforcement learning (meta-RL) holds the promise of allowing RL agents to acquire novel tasks with improved efficiency
by learning to learn from a distribution of tasks~\citep{finn2017model, rakelly2019efficient}.
In spite of recent advances in the field,
most existing meta-RL algorithms are restricted to short-horizon, dense-reward tasks.
To facilitate efficient learning on long-horizon, sparse-reward tasks,
recent works 
aim to leverage experience from prior tasks 
in the form of offline datasets 
without additional reward and task annotations
~\citep{lynch2020learning,pertsch2020spirl,chebotar2021actionable}. 
While these methods can solve complex tasks 
with substantially improved sample efficiency over methods learning from scratch,
millions of interactions with environments are still required 
to acquire long-horizon skills.

In this work, we aim to take a step towards combining the capabilities of 
\emph{both} learning how to quickly learn new tasks while \emph{also} 
leveraging prior experience in the form of unannotated offline data (see Figure~\ref{fig:teaser}). 
Specifically, 
we aim to devise a method that enables 
meta-learning on complex, long-horizon tasks 
and can solve unseen target tasks with orders of magnitude fewer environment interactions than prior works.

We propose to leverage the offline experience by extracting reusable \emph{skills} --
short-term behaviors that can be composed to solve unseen long-horizon tasks. 
We employ a hierarchical meta-learning scheme in which we meta-train a high-level 
policy to learn how to quickly reuse the extracted skills.
To efficiently explore the learned skill space during meta-training,
the high-level policy is guided by a skill prior
which is also acquired from the offline experience data.

We evaluate our method and prior approaches 
in deep RL, skill-based RL, meta-RL, and multi-task RL 
on
two challenging continuous control environments: maze navigation and kitchen manipulation, 
which require long-horizon control and only provides sparse rewards. 
Experimental results show that %
our method 
can
efficiently solve unseen tasks 
by exploiting meta-learning tasks and offline datasets,
while prior approaches 
require substantially more samples
or fail to solve the tasks. 

In summary, the main contributions of this paper are threefold:
\begin{itemize}
    \item To the best of our knowledge, this is the first work to combine meta-reinforcement learning algorithms with task-agnostic offline datasets that do not contain reward or task annotations.
    \item We propose a method that combines meta-learning with offline data by extracting learned skills and a skill prior as well as meta-learning a hierarchical skill policy regularized by the skill prior.
    \item We empirically show that our method is significantly more efficient at learning long-horizon sparse-reward tasks 
    compared to prior methods in deep RL, skill-based RL, meta-RL, and multi-task RL.
\end{itemize}

%% file: fig/teaser.tex
\begin{compactwrapfigure}[15]{OR}{0.5\textwidth}
    \centering
    \includegraphics[width=0.97\linewidth]{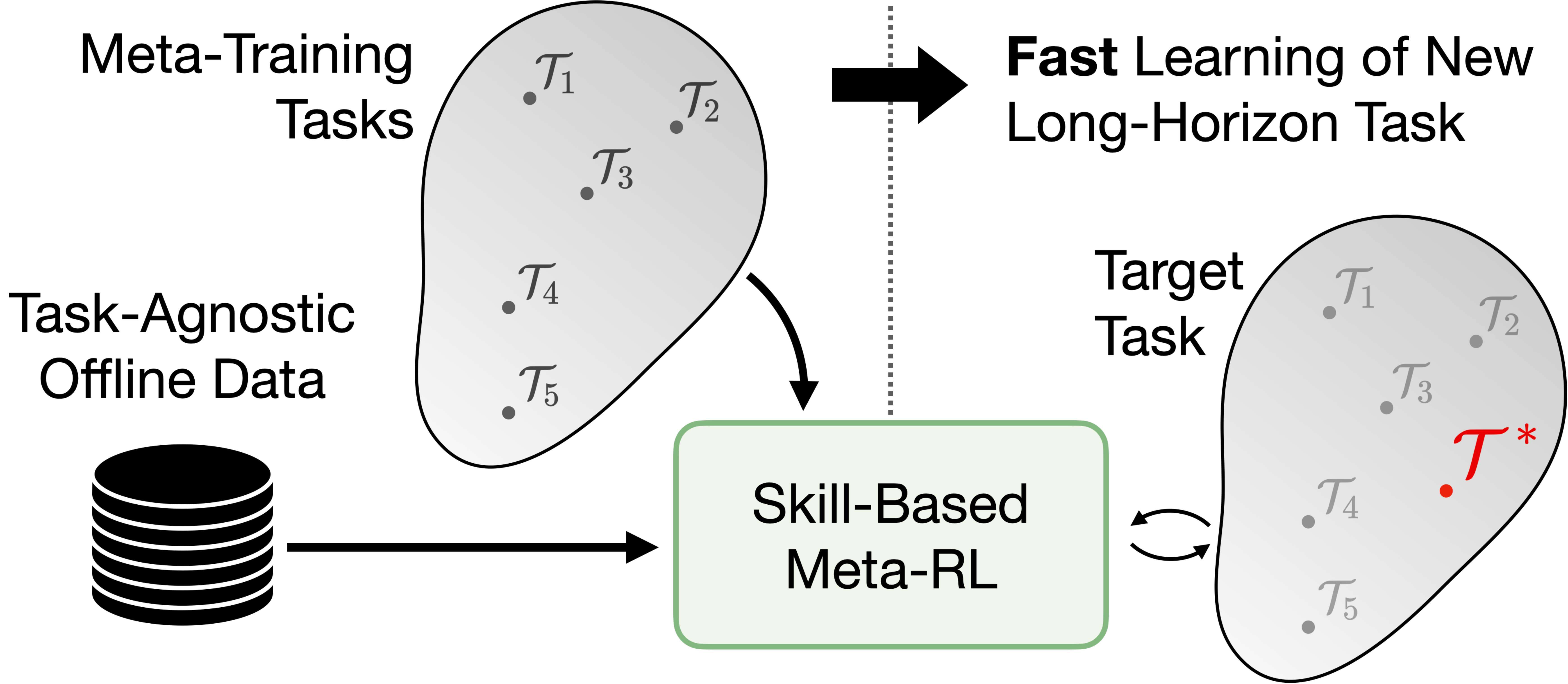}
    \vspace{-0.1cm}
    \caption{\small
    We propose a method that jointly leverages 
    (1) a large offline dataset of prior experience collected across many tasks without reward or task annotations and 
    (2) a set of meta-training tasks to 
    learn how to quickly solve unseen long-horizon tasks. 
    Our method extracts reusable skills from the offline dataset 
    and meta-learn a policy to quickly use them for solving new tasks.
    }
    \vspace{-0.55cm}
    \label{fig:teaser}
\end{compactwrapfigure}

%% file: sections/02_related_work.tex
\vspacesection{Related Work}
\label{sec:related}

\noindent \textbf{Meta-Reinforcement Learning.}
Meta-RL approaches~\citep{duan2016rl, wang2016learning, finn2017model, yu2018one, rothfuss2018promp, NEURIPS2018_4de75424, vuorio2018toward, nagabandi2018deep, clavera2018learning, clavera18a,
rakelly2019efficient,
vuorio2019multimodal, yang2019norml, zintgraf2019fast, humplik2019meta, varibad,  liu2021decoupling} hold the promise of 
allowing learning agents to quickly adapt to novel tasks
by learning to learn from a distribution of tasks.
Despite the recent advances in the field,
most existing meta-RL algorithms are limited to 
short-horizon and dense-reward tasks.
In contrast, we aim to develop a method that can meta-learn to solve 
long-horizon tasks with sparse rewards
by leveraging offline datasets.

\Skip{
--> learns fast but only on easy tasks
    \begin{itemize}
        \item meta-learning (older-than-MAML stuff)
        \item meta-RL (MAML, PEARL)
    \end{itemize}
}

\noindent \textbf{Offline datasets.}
Recently, many works have investigated the usage of offline datasets for agent training. In particular, the field of \emph{offline reinforcement learning}~\citep{levine2020offline,siegel2020keep,kumar2020conservative,yu2021combo} aims to devise methods that can perform RL fully offline from pre-collected data, without the need for environment interactions. However, these methods require target task reward annotations on the offline data for every new tasks that should be learned. These reward annotations can be challenging to obtain, especially if the offline data is collected from a diverse set of prior tasks. In contrast, our method is able to leverage offline datasets without any reward annotations.

\noindent \textbf{Offline Meta-RL.}
Another recent line of research aims to \emph{meta-learn} from static, pre-collected datasets including reward annotations ~\citep{mitchell2021offline, pong2021offline, dorfman2020offline}. After meta-training with the offline datasets, these works aim to quickly adapt to a new task with only a small amount of data from that new task. In contrast to the aforementioned offline RL methods these works aim to adapt to \emph{unseen} tasks and assume access to only \emph{limited data} from the new tasks. However, in addition to reward annotations, these approaches often require that the offline training data is split into separate datasets for each training tasks, further limiting the scalability. %
\Skip{
\begin{itemize}
        \item does meta-RL but assumes reward-labeled offline data --> this is not as scalable as our assumption of reward-free offline data
    \end{itemize}
}

\noindent \textbf{Skill-based Learning.}
An alternative approach for leveraging offline data that does not require reward or task annotations is through the extraction of skills -- reusable short-horizon behaviors. Methods for skill-based learning recombine these skills for learning unseen target tasks and converge substantially faster than methods that learn from scratch~\citep{lee2018composing, hausman2018learning, sharma2019dynamics, sun2022program}. 
When trained from diverse datasets these approaches can extract a wide repertoire of skills and learn complex, long-horizon tasks~\citep{merel2019reusable,lynch2020learning,pertsch2020spirl,ajay2020opal,chebotar2021actionable,pertsch2021skild}. Yet, although they are more efficient than training from scratch, they still require a large number of environment interactions to learn a new task. Our method instead combines skills extracted from offline data with meta-learning, leading to significantly improved sample efficiency.

\Skip{
learns complex tasks but requires many steps or demos
    \begin{itemize}
        \item pre-defined skills (YW's skill composition, SH's program stuff)
        \item learned skills (Karol's paper, SPiRL, OPAL) + demos (SkiLD, FIST)
    \end{itemize}
}

%% file: sections/03_prelim.tex
\vspacesection{Problem Formulation and Preliminaries}
\label{sec:prelim}

Our approach builds on prior work for meta-learning and learning from offline datasets and aims to combine the best of both worlds. In the following we will formalize our problem setup and briefly summarize relevant prior work.

\noindent\textbf{Problem Formulation.} Following prior work on learning from large offline datasets~\citep{lynch2020learning,pertsch2020spirl,pertsch2021skild}, we assume access to a dataset of state-action trajectories $\mathbf{D} = \{s_t, a_t, ...\}$ which is collected either across a wide variety of tasks or as ``play data'' with no particular task in mind. We thus refer to this dataset as \emph{task-agnostic}. 
With a large number of data collection tasks, 
the dataset covers a wide variety of behaviors 
and can be used to accelerate learning on diverse tasks. 
Such data can be collected at scale, \eg through autonomous exploration~\citep{hausman2018learning,sharma2019dynamics,dasari2019robonet}, human teleoperation~\citep{schaal2005motion,gupta2019relay,mandlekar2018roboturk,lynch2020learning}, or from previously trained agents~\citep{fu2020d4rl,gulcehre2020rl}. We additionally assume access to a set of meta-training tasks $\mathbf{T} = \{\mathcal{T}_1, \dots, \mathcal{T}_N\}$, where each task is represented as a Markov decision process (MDP) defined by a tuple $\{\mathcal{S}, \mathcal{A}, \mathcal{P}, r, \rho, \gamma\}$ of states, actions, transition probability, reward, initial state distribution, and discount factor.

Our goal is to leverage both, the offline dataset $\mathbf{D}$ and the meta-training tasks $\mathbf{T}$, to accelerate the training of a policy $\pi(a \vert s)$ on a target task $\mathcal{T}^*$ which is also represented as an MDP. Crucially, we do not assume that $\mathcal{T}^*$ is a part of the set of training tasks $\mathbf{T}$, nor that $\mathbf{D}$ contains demonstrations for solving $\mathcal{T}^*$. 
Thus, 
we aim to design an algorithm that can leverage offline data and meta-training tasks for learning how to quickly compose known skills for solving an unseen target task. 
Next, we will describe existing approaches that \emph{either} leverage offline data \emph{or} meta-training tasks to accelerate target task learning.
Then, we describe how our approach takes advantage of the best of both worlds.

\noindent\textbf{Skill-based RL.} One successful approach for leveraging task-agnostic datasets for accelerating the learning of unseen tasks is though the transfer of reusable \emph{skills}, 
\ie short-horizon behaviors that can be composed to solve long-horizon tasks. 
Prior work in skill-based RL called Skill-Prior RL (SPiRL, \citet{pertsch2020spirl}) proposes an effective way to implement this idea. 
Specifically, SPiRL uses a task-agnostic dataset to learns two models: 
(1)~a skill policy $\pi(a \vert s, z)$ that decodes a latent skill representation $z$ into a sequence of executable actions and 
(2)~a prior over latent skill variables $p(z \vert s)$ which can be leveraged to guide exploration in skill space. 
SPiRL uses these skills for learning new tasks efficiently by training a high-level skill policy $\pi(z \vert s)$ that acts over the space of learned skills instead of primitive actions. 
The target task RL objective extends Soft Actor Critic (SAC, \citet{haarnoja2018sac}), a popular off-policy RL algorithm, by guiding the high-level policy with the learned skill prior:
\begin{equation}
\label{eq:spirl_objective}
    \max_\pi
    \sum_{t}
        \mathbb{E}_{(s_t, z_t) \sim \rho_\pi} \big[
            r(s_t, z_t) -
            \alpha D_{\text{KL}} \big(
                \pi(z|s_t), p(z|s_t)
            \big)
        \big].
\end{equation}
Here $D_\text{KL}$ denotes the Kullback-Leibler divergence between the policy and skill prior, and $\alpha$ is a weighting coefficient.

\noindent\textbf{Off-Policy Meta-RL.} 
\citet{rakelly2019efficient} introduced an off-policy meta-RL algorithm called probabilistic embeddings for actor-critic RL (PEARL) that 
leverages a set of training tasks $\mathbf{T}$ to enable quick learning of new tasks. 
\Skip{
\sun{discuss here: we dont want to say pearl+spirl is trivial. we can instead say pearl is sample efficient so we use it.}
While there is a wide range of meta-RL algorithms (see Section~\ref{sec:related}), PEARL is a natural fit for our approach since analogous to SPiRL it extends the SAC algorithm and thus eases a combination of both methods in our method. 
}
Specifically, PEARL leverages the meta-training tasks for learning a task encoder $q(e \vert c)$. This encoder takes in a small set of state-action-reward transitions $c$ and produces a task embedding $e$. This embedding is used to condition the actor $\pi(a \vert s, z)$ and critic $Q(s, a, e)$. In PEARL, actor, critic and task encoder are trained by jointly maximizing the obtained reward and the policy's entropy~$\mathcal{H}$~\citep{haarnoja2018sac}:
\begin{equation}
\label{eq:pearl_objective}
    \max_\pi \mathbb{E}_{\mathcal{T} \sim p_{\mathcal{T}}, e \sim q(\cdot \vert c^{\mathcal{T}})} \bigg[
        \sum_{t}
        \mathbb{E}_{(s_t, a_t) \sim \rho_{\pi \vert e} } \big[
            r_{\mathcal{T}}(s_t, a_t) +
            \alpha \mathcal{H}\big(\pi(a \vert s_t, e)\big)
        \big]
    \bigg].
\end{equation}
Additionally, the task embedding output of the task encoder is regularized towards a constant prior distribution $p(e)$.

%% file: sections/04_approach.tex
\vspacesection{Approach}
\label{sec:approach}

\begin{figure}[t]
    \centering
    \includegraphics[width=1\linewidth]{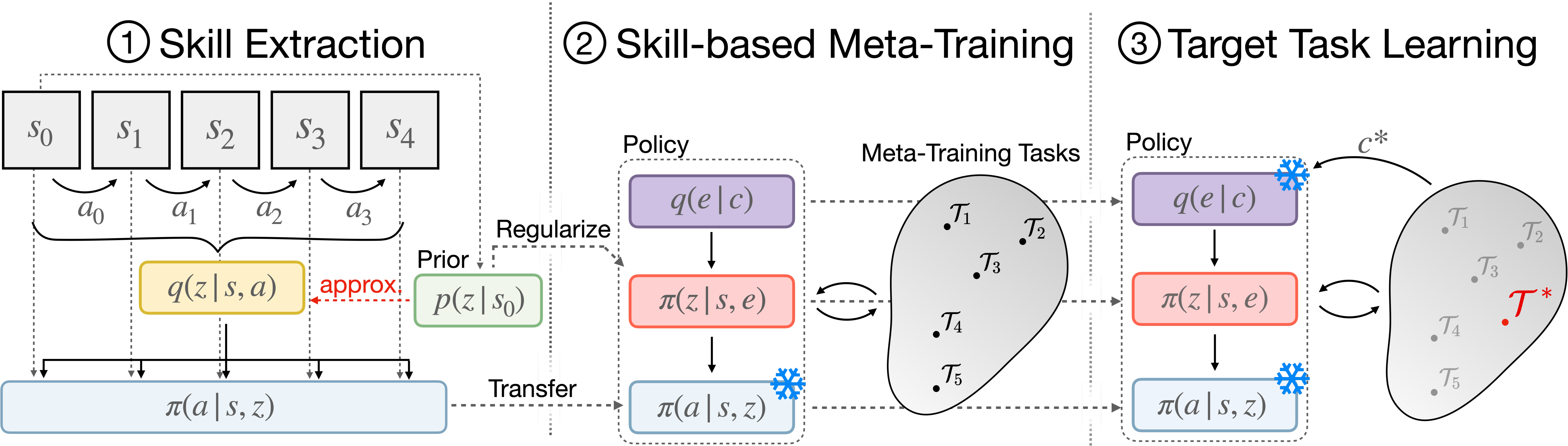}
    \vspace{-0.3cm}
    \caption{\small
    \textbf{Method Overview.}
    Our proposed skill-based meta-RL method has three phases.
    \textbf{(1) Skill Extraction}: learns reusable skills from snippets of task-agnostic offline data through a skill extractor (\textcolor[HTML]{F4C430}{\textbf{yellow}}) and low-level skill policy (\textcolor[HTML]{0F52BA}{\textbf{blue}}). Also trains a prior distribution over skill embeddings (\textcolor[HTML]{4CA577}{\textbf{green}}). \textbf{(2) Skill-based Meta-training}: Meta-trains a high-level skill policy (\textcolor[HTML]{FF6350}{\textbf{red}}) and task encoder (\textcolor[HTML]{79678F}{\textbf{purple}}) while using the pre-trained low-level policy. The pre-trained skill prior is used to regularize the high-level policy during meta-training and guide exploration. \textbf{(3) Target Task Learning}: Leverages the meta-trained hierarchical policy for quick learning of an unseen target task. After conditioning the policy by encoding a few transitions $c^*$ from the target task $\mathcal{T}^*$, we continue fine-tuning the high-level skill policy on the target task while regularizing it with the pre-trained skill prior.
    }
    \label{fig:model}
\end{figure}

We propose
\textbf{S}k\textbf{i}ll-based \textbf{M}eta-\textbf{P}olicy \textbf{L}earning (\method\\),
an algorithm for jointly leveraging offline data as well as a set of meta-training tasks to accelerate the learning of unseen target tasks. 
Our method has three phases: 
\textbf{(1) skill extraction}: 
we extract reusable skills and a skill prior from the offline data (\mysecref{sec:extraction}), 
\textbf{(2) skill-based meta-training}: 
we utilize the meta-training tasks to learn how to leverage the extracted skills and skill prior to efficiently solve new tasks
(\mysecref{sec:meta-training}), 
\textbf{(3) target task learning}: 
we fine-tune the meta-trained policy to rapidly adapt to solve an unseen target task (\mysecref{sec:target}).
An illustration of the proposed method is shown in~\myfig{fig:model}.

\vspacesubsection{Skill Extraction}
\label{sec:extraction}
To acquire a set of reusable skills from the offline dataset $\mathbf{D}$,
we leverage the skill extraction approach proposed in~\citet{pertsch2020spirl}.
Concretely, we jointly train (1)~a skill encoder $q(z \vert s_{0:K}, a_{0:K-1})$ that embeds an $K$-steps trajectory randomly cropped from the sequences in $\mathbf{D}$ into a low-dimensional skill embedding $z$, and 
(2)~a low-level skill policy $\pi(a_t \vert s_t, z)$ that is trained with behavioral cloning to reproduce the action sequence $a_{0:K-1}$ given the skill embedding. 
To learn a smooth skill representation, we regularize the output of the skill encoder with a unit Gaussian prior distribution, and weight this regularization by a coefficient $\beta$~\citep{higgins2017beta}:
\begin{equation}
\label{eq:ours_objective_pretrain}
    \max_{q, \pi} \mathbb{E}_{z \sim q} \bigg[
        \underbrace{\prod_{t=0}^{K-1} \log \pi(a_t \vert s_t, z)}_{\text{behavioral cloning}} -
        \underbrace{\vphantom{\prod_{t=1}^{K-1}} \beta D_\text{KL}\big( q(z \vert s_{0:K}, a_{0:K-1}), \mathcal{N}(0, I)\big)}_{\text{embedding regularization}}
    \bigg].
\end{equation}
Additionally, we follow \citet{pertsch2020spirl} and learn a skill prior $p(z \vert s)$ that captures the distribution of skills likely to be executed in a given state under the training data distribution. The prior is trained to match the output of the skill encoder: $\min_p D_\text{KL}\big(\lfloor q(z \vert s_{0:K}, a_{0:K-1})\rfloor, p(z\vert s_0)\big)$. 
Here $\lfloor\cdot\rfloor$ indicates that gradient flow is stopped into the skill encoder for training the skill prior.

\vspacesubsection{Skill-based Meta-Training}
\label{sec:meta-training}
We aim to learn a policy that can quickly learn to leverage the extracted skills to solve new tasks. 
We leverage off-policy meta-RL (see Section~\ref{sec:prelim}) to learn such a policy using our set of meta-training tasks $\mathbf{T}$. 
Similar to PEARL~\citep{rakelly2019efficient}, we train a task-encoder that takes in a set of sampled transitions and produces a task embedding $e$. Crucially, we leverage our learned skills by training a task-embedding-conditioned policy over \emph{skills} instead of primitive actions: $\pi(z \vert s, e)$, thus equipping the policy with a set of useful pre-trained behaviors and reducing the meta-training task to learning how to combine these behaviors instead of learning them from scratch. We find that this usage of offline data through learned skills is crucial for enabling meta-training on complex, long-horizon tasks (see Section~\ref{sec:experiments}).

Prior work has shown that the efficiency of RL on learned skill spaces can be substantially improved by guiding the policy with a learned skill prior~\citep{pertsch2020spirl,ajay2020opal}. Thus, instead of regularizing with a maximum entropy objective as done in prior work on off-policy meta-RL~\citep{rakelly2019efficient}, we propose to regularize the meta-training policy with our pre-trained skill prior, leading to the following meta-training objective:
\begin{equation}
\label{eq:ours_objective_metarl}
    \max_\pi \mathbb{E}_{\mathcal{T} \sim p_{\mathcal{T}}, e \sim q(\cdot \vert c^{\mathcal{T}})} \bigg[
        \sum_{t}
            \mathbb{E}_{
                (s_t, z_t) \sim \rho_{\pi \vert e}
            } \big[
                r_{\mathcal{T}}(s_t, z_t) -
                \alpha D_\text{KL}\big(\pi(z \vert s_t, e), p(z \vert s_t)\big)
            \big]
    \bigg].
\end{equation}

where $\alpha$ 
determines the strength of the prior regularization. We automatically tune $\alpha$ via dual gradient descent by choosing a target divergence $\delta$ between policy and prior~\citep{pertsch2020spirl}. 

To compute the task embedding $e$, we used multiple different sizes of $c$.
We found that we can increase training stability by adjusting the strength of the prior regularization to the size of the conditioning set. Intuitively, when the high-level policy is conditioned on only a few transitions, i.e. when the set $c$ is small, it has only little information about the task at hand and should thus be regularized stronger towards the task-agnostic skill prior. Conversely, when $c$ is large, the policy likely has more information about the target task and thus should be allowed to deviate from the skill prior more to solve the task, i.e. have a weaker regularization strength.

To implement this intuition,
we employ a simple approach: 
we define \emph{two} target divergences $\delta_1$ and $\delta_2$ and associated auto-tuned coefficients $\alpha_1$ and $\alpha_2$ with $\delta_1 < \delta_2$. We regularize the policy using the larger coefficient $\alpha_1$ with small  conditioning transition set and otherwise we regularize using the smaller coefficient $\alpha_2$.
We found this technique simple yet sufficient in our experiments and leave the investigation of more sophisticated regularization approaches for future work.

\vspacesubsection{Target Task Learning}
\label{sec:target}
When a target task is given,
we aim to leverage the meta-trained policy for quickly learning how to solve it. Intuitively, the policy should first explore different skill options to learn about the task at hand and then rapidly narrow its output distribution to those skills that solve the task.
We implement this intuition by first collecting a small set of conditioning transitions $c^*$ from the target task by exploring with the meta-trained policy. 
Since we have no information about the target task at this stage, we explore the environment by conditioning our pre-trained policy with task embeddings sampled from the task prior $p(e)$. 
Then, we encode this set of transitions into a target task embedding $e^* \sim q(e \vert c^*)$. 
By conditioning our meta-trained high-level policy on this encoding, we can rapidly narrow its skill distribution to skills that solve the given target task: $\pi(z \vert s, e^*)$. 

Empirically, we find that this policy is often already able to achieve high success rates on the target task. 
Note that only very few interactions with the environment for collecting $c^*$ are required for learning a complex, long-horizon and unseen target task with sparse reward. 
This is substantially more efficient than prior approaches such as SPiRL that require orders of magnitude more target task interactions for achieving comparable performance.

To further improve the performance on the target task, 
we fine-tune the conditioned policy with target task rewards while guiding its exploration with the pre-trained skill prior\footnote{Other regularization distributions are possible during fine-tuning, e.g. the high-level policy conditioned on task prior samples $p(z \vert s, e \sim p(e))$ or the target task embedding conditioned policy $p(z \vert s, e^*)$ \emph{before finetuning}. Yet, we found the regularization with the pre-trained task-agnostic skill prior to work best in our experiments.}:

\begin{equation}
\label{eq:pearl_objective}
    \max_\pi 
    \mathbb{E}_{
    e^* \sim q(\cdot \vert c^*)} \bigg[
        \sum_{t}
        \mathbb{E}_{
            (s_t, z_t) \sim \rho_{\pi \vert e^*}
        } \big[
            r_{\mathcal{T}^*}(s_t, z_t) -
            \alpha D_\text{KL}\big(\pi(z \vert s_t, e^*), p(z \vert s_t)\big)
        \big]
    \bigg].
\end{equation}

In practice, we propose several techniques for stabilizing meta-training and fine-tuning:
(1)~adaptively regularizing the policy based on the size of the conditioning trajectory set as described in Section~\ref{sec:meta-training},
(2)~parameterizing the policy as a residual policy that outputs differences to the pre-trained skill prior instead of the approach from~\citet{pertsch2020spirl} that directly fine-tunes the skill prior, and
(3)~initializing the Q-function and $\alpha$ parameter during fine-tuning with meta-trained parameters instead of randomly initialized networks. 
We discuss these techniques in detail in~\mysecref{sec:implementation_details}. 

%% file: sections_metalearn/04_approach.tex
\vspacesection{Approach}
\label{sec:approach}

\begin{figure}[t]
    \centering
    \includegraphics[width=1\linewidth]{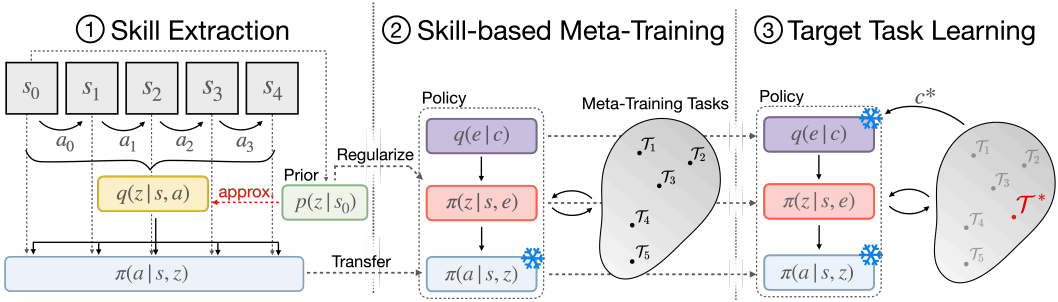}
    \vspace{-0.3cm}
    \caption{\small
    \textbf{Method Overview.}
    Our proposed skill-based meta-RL method has three phases.
    \textbf{(1) Skill Extraction}: learns reusable skills from snippets of task-agnostic offline data through a skill extractor (\textcolor[HTML]{F4C430}{\textbf{yellow}}) and low-level skill policy (\textcolor[HTML]{0F52BA}{\textbf{blue}}). Also trains a prior distribution over skill embeddings (\textcolor[HTML]{4CA577}{\textbf{green}}). \textbf{(2) Skill-based Meta-training}: Meta-trains a high-level skill policy (\textcolor[HTML]{FF6350}{\textbf{red}}) and task encoder (\textcolor[HTML]{79678F}{\textbf{purple}}) while using the pre-trained low-level policy. The pre-trained skill prior is used to regularize the high-level policy during meta-training and guide exploration. \textbf{(3) Target Task Learning}: Leverages the meta-trained hierarchical policy for quick learning of an unseen target task. After conditioning the policy by encoding a few transitions $c^*$ from the target task $\mathcal{T}^*$, we continue fine-tuning the high-level skill policy on the target task while regularizing it with the pre-trained skill prior.
    }
    \label{fig:model}
\end{figure}

We propose
\textbf{S}k\textbf{i}ll-based \textbf{M}eta-\textbf{P}olicy \textbf{L}earning (\method\\),
an algorithm for jointly leveraging offline data as well as a set of meta-training tasks to accelerate the learning of unseen target tasks. 
Our method has three phases: 
\textbf{(1) skill extraction}: 
we extract reusable skills and a skill prior from the offline data (\mysecref{sec:extraction}), 
\textbf{(2) skill-based meta-training}: 
we utilize the meta-training tasks to learn how to leverage the extracted skills and skill prior to efficiently solve new tasks
(\mysecref{sec:meta-training}), 
\textbf{(3) target task learning}: 
we fine-tune the meta-trained policy to rapidly adapt to solve an unseen target task (\mysecref{sec:target}).
An illustration of the proposed method is shown in~\myfig{fig:model}.

\vspacesubsection{Skill Extraction}
\label{sec:extraction}
To acquire a set of reusable skills from the offline dataset $\mathbf{D}$,
we leverage the skill extraction approach proposed in~\citet{pertsch2020spirl}.
Concretely, we jointly train (1)~a skill encoder $q(z \vert s_{0:K}, a_{0:K-1})$ that embeds an $K$-steps trajectory randomly cropped from the sequences in $\mathbf{D}$ into a low-dimensional skill embedding $z$, and 
(2)~a low-level skill policy $\pi(a_t \vert s_t, z)$ that is trained with behavioral cloning to reproduce the action sequence $a_{0:K-1}$ given the skill embedding. 
To learn a smooth skill representation, we regularize the output of the skill encoder with a unit Gaussian prior distribution, and weight this regularization by a coefficient $\beta$~\citep{higgins2017beta}:
\begin{equation}
\label{eq:ours_objective_pretrain}
    \max_{q, \pi} \mathbb{E}_{z \sim q} \bigg[
        \underbrace{\prod_{t=0}^{K-1} \log \pi(a_t \vert s_t, z)}_{\text{behavioral cloning}} -
        \underbrace{\vphantom{\prod_{t=1}^{K-1}} \beta D_\text{KL}\big( q(z \vert s_{0:K}, a_{0:K-1}), \mathcal{N}(0, I)\big)}_{\text{embedding regularization}}
    \bigg].
\end{equation}
Additionally, we follow \citet{pertsch2020spirl} and learn a skill prior $p(z \vert s)$ that captures the distribution of skills likely to be executed in a given state under the training data distribution. The prior is trained to match the output of the skill encoder: $\min_p D_\text{KL}\big(\lfloor q(z \vert s_{0:K}, a_{0:K-1})\rfloor, p(z\vert s_0)\big)$. 
Here $\lfloor\cdot\rfloor$ indicates that gradient flow is stopped into the skill encoder for training the skill prior.

\vspacesubsection{Skill-based Meta-Training}
\label{sec:meta-training}
We aim to learn a policy that can quickly learn to leverage the extracted skills to solve new tasks. 
We leverage off-policy meta-RL (see Section~\ref{sec:prelim}) to learn such a policy using our set of meta-training tasks $\mathbf{T}$. 
Similar to PEARL~\citep{rakelly2019efficient}, we train a task-encoder that takes in a set of sampled transitions and produces a task embedding $e$. Crucially, we leverage our learned skills by training a task-embedding-conditioned policy over \emph{skills} instead of primitive actions: $\pi(z \vert s, e)$, thus equipping the policy with a set of useful pre-trained behaviors and reducing the meta-training task to learning how to combine these behaviors instead of learning them from scratch. We find that this usage of offline data through learned skills is crucial for enabling meta-training on complex, long-horizon tasks (see Section~\ref{sec:experiments}).

Prior work has shown that the efficiency of RL on learned skill spaces can be substantially improved by guiding the policy with a learned skill prior~\citep{pertsch2020spirl,ajay2020opal}. Thus, instead of regularizing with a maximum entropy objective as done in prior work on off-policy meta-RL~\citep{rakelly2019efficient}, we propose to regularize the meta-training policy with our pre-trained skill prior, leading to the following meta-training objective:
\begin{equation}
\label{eq:ours_objective_metarl}
    \max_\pi \mathbb{E}_{\mathcal{T} \sim p_{\mathcal{T}}, e \sim q(\cdot \vert c^{\mathcal{T}})} \bigg[
        \sum_{t}
            \mathbb{E}_{
                (s_t, z_t) \sim \rho_{\pi \vert e}
            } \big[
                r_{\mathcal{T}}(s_t, z_t) -
                \alpha D_\text{KL}\big(\pi(z \vert s_t, e), p(z \vert s_t)\big)
            \big]
    \bigg].
\end{equation}

where $\alpha$ 
determines the strength of the prior regularization. We automatically tune $\alpha$ via dual gradient descent by choosing a target divergence $\delta$ between policy and prior~\citep{pertsch2020spirl}. 

To compute the task embedding $e$, we used multiple different sizes of $c$.
We found that we can increase training stability by adjusting the strength of the prior regularization to the size of the conditioning set. Intuitively, when the high-level policy is conditioned on only a few transitions, i.e. when the set $c$ is small, it has only little information about the task at hand and should thus be regularized stronger towards the task-agnostic skill prior. Conversely, when $c$ is large, the policy likely has more information about the target task and thus should be allowed to deviate from the skill prior more to solve the task, i.e. have a weaker regularization strength.

To implement this intuition,
we employ a simple approach: 
we define \emph{two} target divergences $\delta_1$ and $\delta_2$ and associated auto-tuned coefficients $\alpha_1$ and $\alpha_2$ with $\delta_1 < \delta_2$. We regularize the policy using the larger coefficient $\alpha_1$ with small  conditioning transition set and otherwise we regularize using the smaller coefficient $\alpha_2$.
We found this technique simple yet sufficient in our experiments and leave the investigation of more sophisticated regularization approaches for future work.

\vspacesubsection{Target Task Learning}
\label{sec:target}
When a target task is given,
we aim to leverage the meta-trained policy for quickly learning how to solve it. Intuitively, the policy should first explore different skill options to learn about the task at hand and then rapidly narrow its output distribution to those skills that solve the task.
We implement this intuition by first collecting a small set of conditioning transitions $c^*$ from the target task by exploring with the meta-trained policy. 
Since we have no information about the target task at this stage, we explore the environment by conditioning our pre-trained policy with task embeddings sampled from the task prior $p(e)$. 
Then, we encode this set of transitions into a target task embedding $e^* \sim q(e \vert c^*)$. 
By conditioning our meta-trained high-level policy on this encoding, we can rapidly narrow its skill distribution to skills that solve the given target task: $\pi(z \vert s, e^*)$. 

Empirically, we find that this policy is often already able to achieve high success rates on the target task. 
Note that only very few interactions with the environment for collecting $c^*$ are required for learning a complex, long-horizon and unseen target task with sparse reward. 
This is substantially more efficient than prior approaches such as SPiRL that require orders of magnitude more target task interactions for achieving comparable performance.

To further improve the performance on the target task, 
we fine-tune the conditioned policy with target task rewards while guiding its exploration with the pre-trained skill prior\footnote{Other regularization distributions are possible during fine-tuning, e.g. the high-level policy conditioned on task prior samples $p(z \vert s, e \sim p(e))$ or the target task embedding conditioned policy $p(z \vert s, e^*)$ \emph{before finetuning}. Yet, we found the regularization with the pre-trained task-agnostic skill prior to work best in our experiments.}:

\begin{equation}
\label{eq:pearl_objective}
    \max_\pi 
    \mathbb{E}_{
    e^* \sim q(\cdot \vert c^*)} \bigg[
        \sum_{t}
        \mathbb{E}_{
            (s_t, z_t) \sim \rho_{\pi \vert e^*}
        } \big[
            r_{\mathcal{T}^*}(s_t, z_t) -
            \alpha D_\text{KL}\big(\pi(z \vert s_t, e^*), p(z \vert s_t)\big)
        \big]
    \bigg].
\end{equation}

More implementation details on our method can be found in~\mysecref{sec:implementation_details}.

%% file: sections_metalearn/05_experiments.tex
\vspacesection{Experiments}
\label{sec:experiments}

Our experiments aim to answer the following questions:
(1)~Can our proposed method learn to efficiently 
solve long-horizon, sparse reward tasks? 
(2)~Is it crucial to utilize offline datasets to achieve this?
(3)~How can we best leverage the training tasks for efficient learning of target tasks?

\vspacesubsection{Experimental Setup}

We evaluate our approach in two challenging continuous control environments: maze navigation and kitchen manipulation environment,
as illustrated in Figure~\ref{fig:env_fig}. 
While meta-RL algorithms are typically evaluated on tasks that span only a few dozen time steps and provide dense rewards~\citep{finn2017model,rothfuss2018promp,rakelly2019efficient, varibad}, our tasks require to learn long-horizon behaviors over hundreds of time steps from sparse reward feedback and thus pose a new challenge to meta-learning algorithms.

\input{fig/env_fig}

\vspacesubsubsection{Maze Navigation}
\label{sec:maze}

\noindent\textbf{Environment.} 
This 2D maze navigation domain based on the maze navigation problem in~\citet{fu2020d4rl}
requires long-horizon control with hundreds of steps for a successful episode and only provides sparse reward feedback upon reaching the goal. 
The observation space of the agent consists of its 2D position and velocity and it acts via planar, continuous velocity commands.

\noindent\textbf{Offline Dataset \& Meta-training / Target Tasks.} Following~\citet{fu2020d4rl}, we collect a task-agnostic offline dataset by randomly sampling start-goal locations in the maze and using a planner to generate a trajectory that reaches from start to goal. 
Note that the trajectories are not annotated with any reward or task labels (\ie which start-goal location is used for producing each trajectory).
To generate a set of meta-training and target tasks,
we fix the agent's initial position in the center of the maze and sample 
\SI{40}{}
random goal locations for meta-training 
and another set of \SI{10}{} goals for target tasks.
All meta-training and target tasks use the same sparse reward formulation.
More details can be found in~\mysecref{sec:app_maze_task_dist}.

\vspacesubsubsection{Kitchen Manipulation}
\label{sec:kitchen}

\noindent\textbf{Environment.} 
The FrankaKitchen environment of~\citet{gupta2019relay} requires the agent to control a 7-DoF robot arm via continuous joint velocity commands and complete a sequence of manipulation tasks like opening the microwave or turning on the stove.
Successful episodes span 300-500 steps and the agent is only provided a sparse reward signal upon successful completion of a subtask.

\noindent\textbf{Offline Dataset \& Meta-training / Target Tasks.} We leverage a dataset of 600 human-teleoperated manipulation sequences of~\citet{gupta2019relay} for offline pre-training. 
In each trajectory, the robot executes a sequence of four subtasks. 
We then define a set of 23 meta-training tasks 
and 10 target tasks that in turn require 
the consecutive execution of four subtasks 
(see Figure~\ref{fig:env_fig} for examples).
Note that each task consists of a unique combination of subtasks.
More details can be found in~\mysecref{sec:app_kitchen_task_dist}.

\vspacesubsection{Baselines} 
\label{sec:baselines}
We compare 
\method\\
to prior approaches
in RL, skill-based RL, meta-RL, and multi-task RL.
\begin{itemize}
    \item \textbf{SAC}~\citep{haarnoja2018sac} 
    is a state of the art deep RL algorithm.
    It learns to solve a target task from scratch
    without leveraging the offline dataset nor the meta-training tasks.
    
    \item \textbf{SPiRL}~\citep{pertsch2020spirl} 
    is a method designed to leverage offline data through the transfer of learned skills. 
    It acquires skills and a skill prior from the offline dataset
    but does not utilize the meta-training tasks.
    This investigates the benefits our method can obtain from leveraging the meta-training tasks. 

    \item \textbf{PEARL}~\citep{rakelly2019efficient} 
    is a state of the art off-policy meta-RL algorithm that 
    learns a policy which can quickly adapt to unseen test tasks.
    It learns from the meta-training tasks but does not use the offline dataset.
    This examines the benefits of using learned skills in meta-RL.
    
    \item \textbf{PEARL-ft} 
    demonstrates the performance of 
    a PEARL~\citep{rakelly2019efficient} model further fine-tuned on a target task
    using SAC~\citep{haarnoja2018sac}.
    
    \item 
    \textbf{Multi-task RL (MTRL)}
    is a multi-task RL baseline
    which learns from the meta-training tasks
    by distilling individual policies specialized in each task
    into a shared policy, similar to Distral~\citep{distral}.
    Each individual policy is trained using SPiRL by leveraging skills extracted from the offline dataset. 
    Therefore, it utilizes both the meta-training tasks and offline dataset similar to our method.
    This provides a direct comparison of 
    multi-task learning (MTRL) from the training tasks
    vs. meta-learning using them (ours).
\end{itemize}

More implementation details on the baselines can be found in~\mysecref{sec:app_baseline}.

\vspacesubsection{Results}

We present 
the quantitative results in~\myfig{fig:main_curve} 
and the qualitative results on the maze navigation domain in~\myfig{fig:maze_qual}.
In~\myfig{fig:main_curve}, 
\method\\ demonstrates much better sample efficiency 
for learning the unseen target tasks compared to all the baselines.
Without leveraging the offline dataset and meta-training tasks,
SAC is not able to make learning progress on most of the target tasks. 
While PEARL is first trained on the meta-training tasks,
it still achieves poor performance on the target tasks
and fine-tuning it (PEARL-ft) does not yield significant improvement.
We believe this is because both environments provide only sparse rewards 
yet require the model to exhibit long-horizon and complex behaviors,
which is known to be difficult for meta-RL methods~\citep{mitchell2021offline}.

\input{fig/main_curve}
\input{fig/maze_qual}

On the other hand, 
by first extracting skills and acquiring a skill prior from the offline dataset,
SPiRL's performance consistently improves with more samples from the target tasks.
Yet, it requires significantly more environment interactions 
than our method to solve the target tasks
since the policy is optimized using vanilla RL, 
which is not designed to learn to quickly learn new tasks.
While the multi-task RL (MTRL) baseline first learns a multi-task policy 
from the meta-training tasks,
its sample efficiency is similar to SPiRL on target task learning,
which highlights the strength of our proposed method -- meta-learning from the meta-training tasks for fast target task learning.

Compared to the baselines, our method learns the target tasks much quicker.
Within only a few episodes the policy converges to solve 
more than $80\%$ 
of the target tasks in the maze environment and two out of four subtasks in the kitchen manipulation environment. 
The prior-regularized fine-tuning then continues to improve performance. 
The rapidly increasing performance and the overall faster convergence show the benefits of leveraging meta-training tasks in addition to learning from offline data: 
by first learning to learn how to quickly solve tasks using the extracted skills and the skill prior, 
our policy can efficiently solve the target tasks.

The qualitative results presented in~\myfig{fig:maze_qual} 
show that all the methods that leverage the offline dataset 
(\ie \method\\, SPiRL, and MTRL) effectively 
explore the maze in the first episode.
Then, \method\\ converges with much fewer episodes compared to 
SPiRL and MTRL,
underlining the effectiveness of meta-training.
In contrast, PEARL-ft is not able to make learning progress,
justifying the necessity of employing offline datasets
for acquiring long-horizon, complex behaviors.

%% file: fig/env_fig.tex
\begin{figure}[t]
    \centering
    \includegraphics[width=1\linewidth]{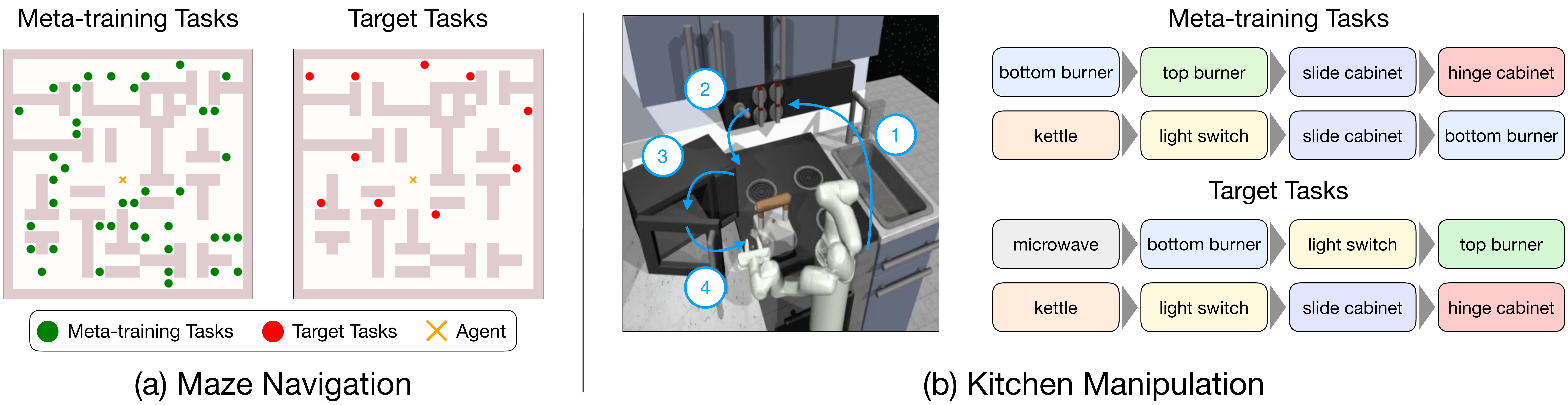}
    \vspace{-0.3cm}
    \caption{\small
        \textbf{Environments.}
        We evaluate our proposed framework in two domains
        that require the learning of complex, long-horizon behaviors from sparse rewards. These environments are substantially more complex than those typically used to evaluate meta-RL algorithms. 
        (a) \textbf{Maze Navigation}: The agent needs to navigate for hundreds of steps to reach unseen target goals and only receives a binary reward upon task success.
        (b) \textbf{Kitchen Manipulation}: The 7DoF agent needs to execute an unseen sequence of four subtasks, spanning hundreds of time steps, and only receives a sparse reward upon completion of each subtask.
    }
    \label{fig:env_fig}
\end{figure}

%% file: fig/main_curve.tex
\begin{figure}
\Skip{
\centering

\subfigure[Maze Navigation]
    {\label{fig:maze_density}\includegraphics[width=0.48\textwidth]{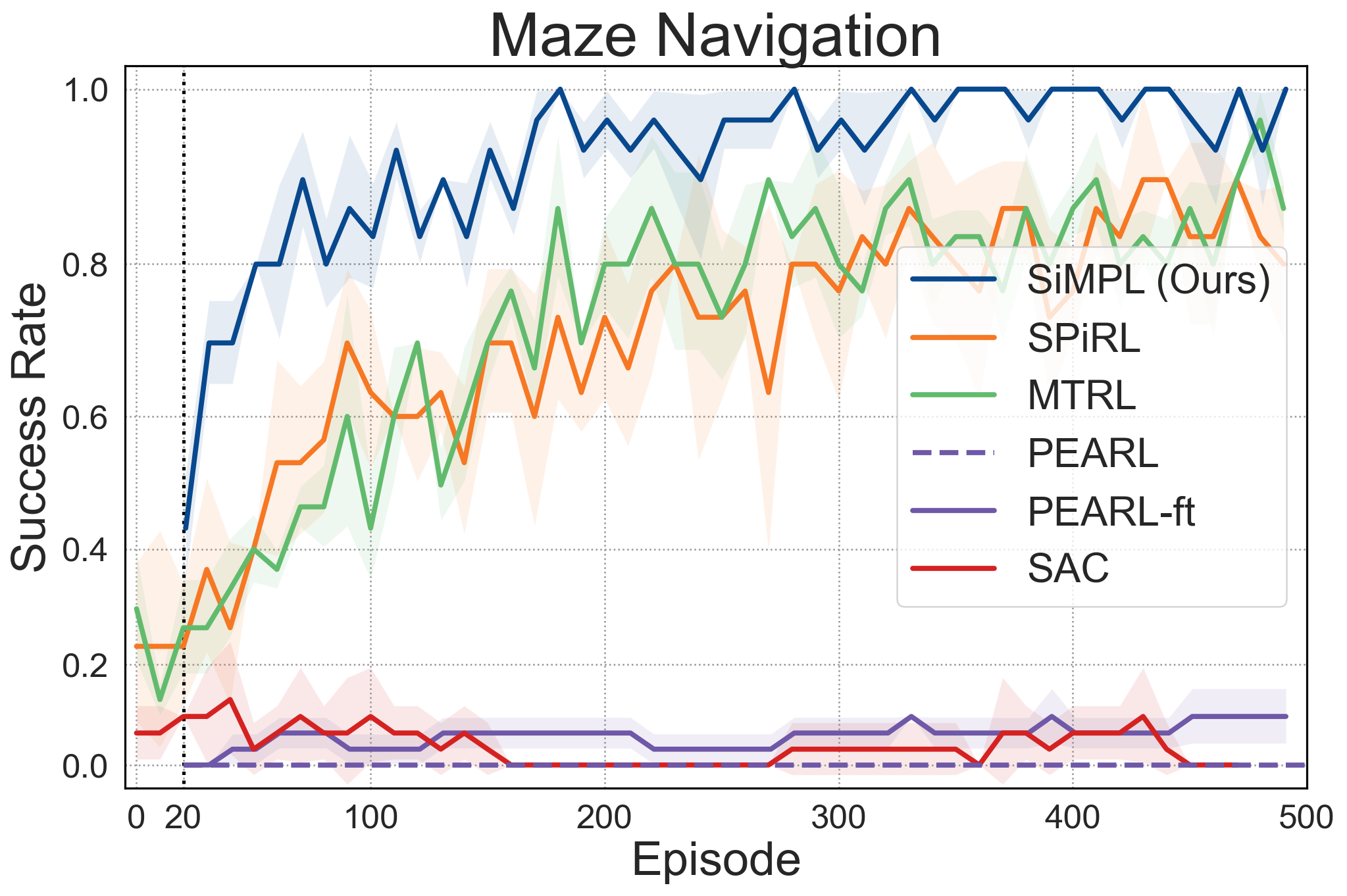}}
\hfill
\subfigure[Kitchen Manipulation]
    {\label{fig:maze_top}\includegraphics[width=0.48\textwidth]{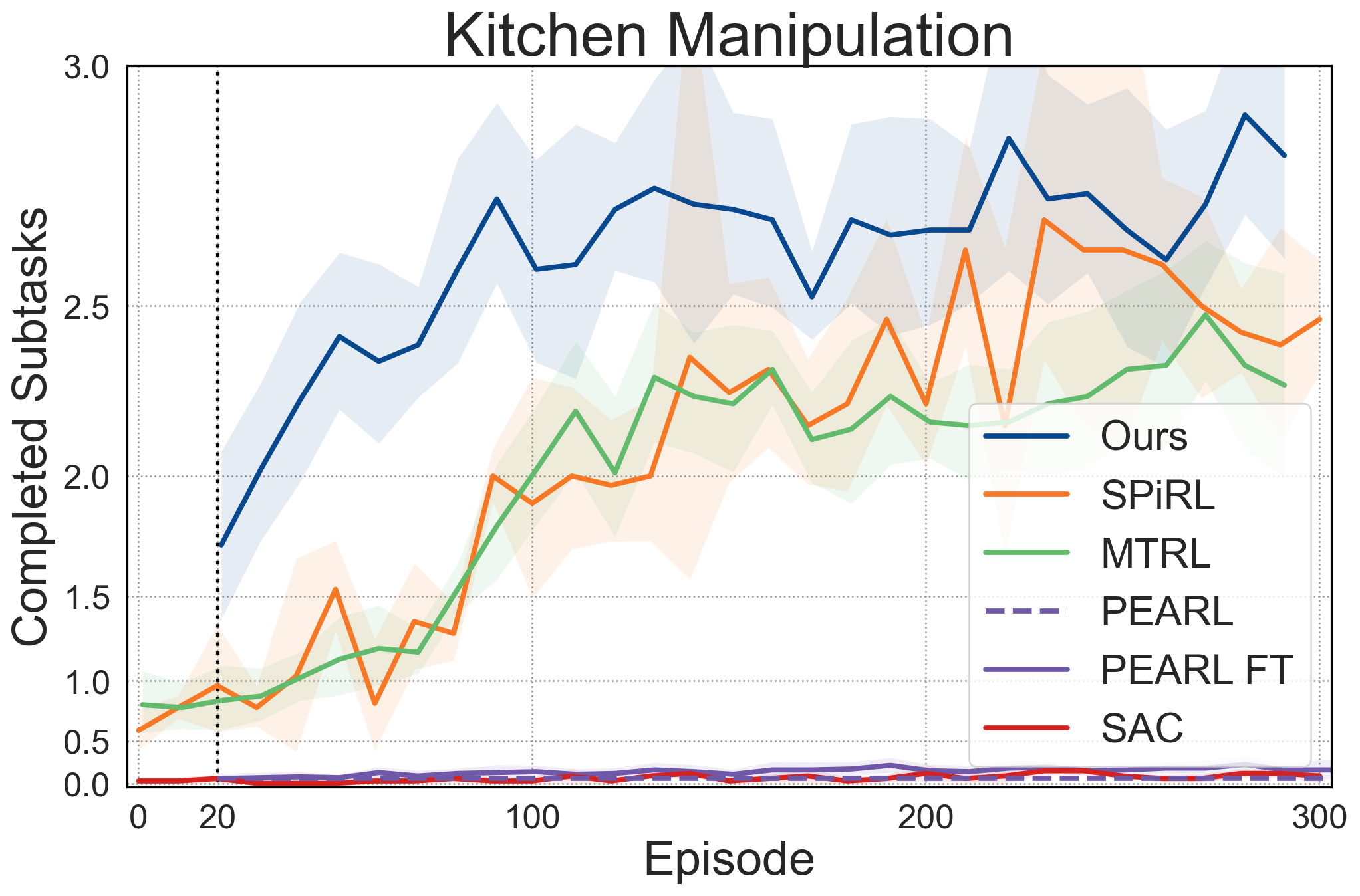}}
\\
\subfigure
    {\label{fig:maze_top}\includegraphics[width=0.6\textwidth]{fig/main_curve_legend.png}}
    \vspace{-0.5cm} 
    }
\includegraphics[width=1\linewidth]{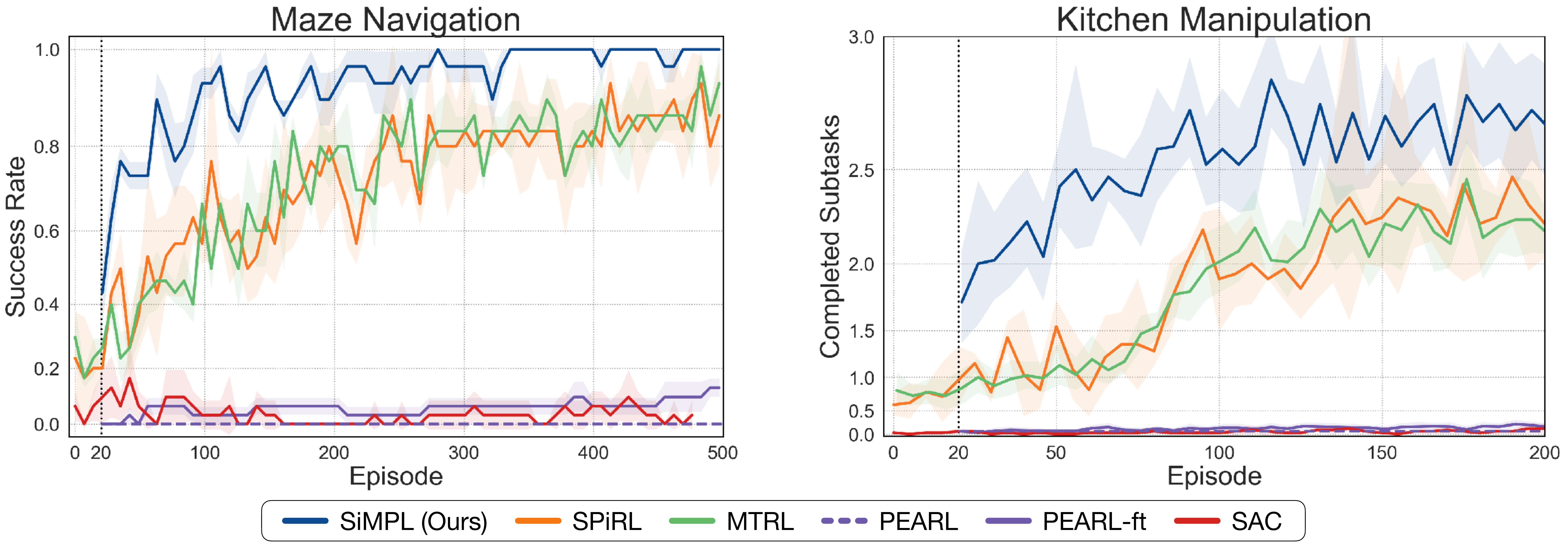}    
    \caption{\small
        \textbf{Target Task Learning Efficiency.}
        \method\\ demonstrates better sample efficiency compared to all the baselines,
        verifying the efficacy of meta-learning on long-horizon tasks
        by leveraging skills and skill prior extracted from an offline dataset.
        For both the two environments, 
        we train each model on each target task with 3 different random seeds.
        \method\\ and PEARL-ft first collect 20 episodes of environment interactions 
        (vertical dotted line) for
        conditioning the meta-trained policy before fine-tuning it on target tasks.
	    \label{fig:main_curve}
}
\end{figure}

%% file: fig/maze_qual.tex
\begin{figure}[t]
    \centering
    \includegraphics[width=1\linewidth]{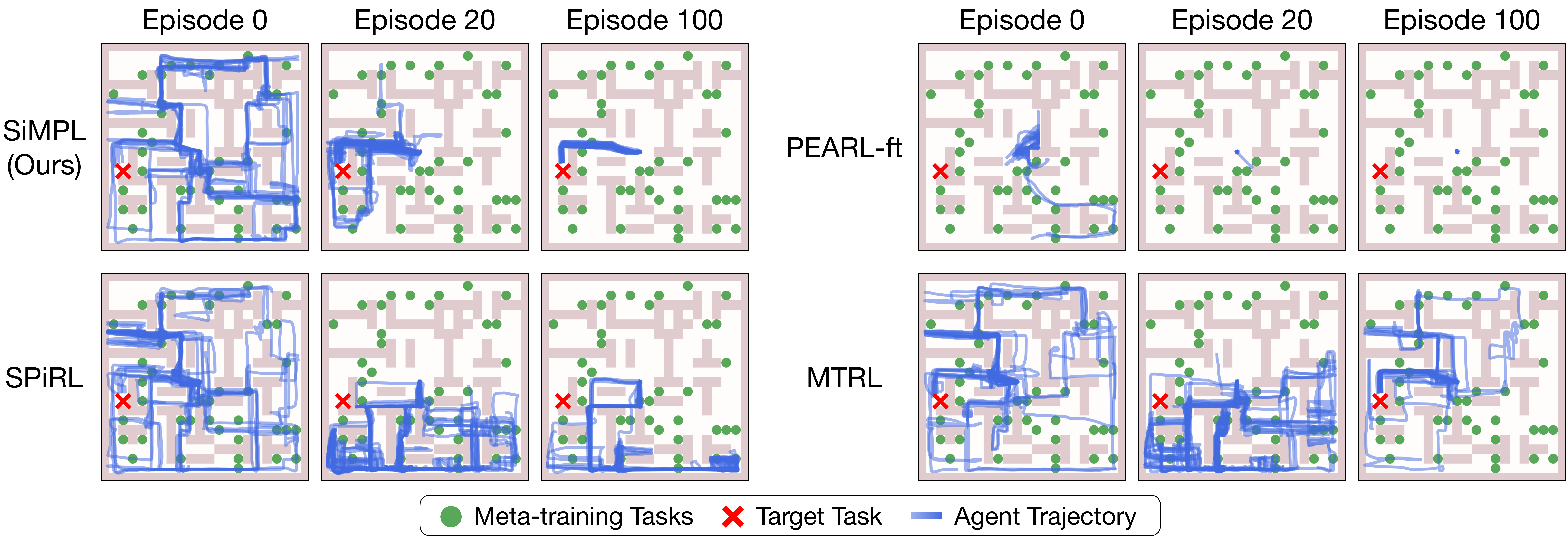}
    \vspace{-0.5cm}
    \caption{\small
        \textbf{Qualitative Results.}
        All the methods that leverage the offline dataset 
        (\ie \method\\, SPiRL, and MTRL) effectively 
        explore the maze in the first episode.
        Then, \method\\ converges with much fewer episodes compared to 
        SPiRL and MTRL.
        In contrast, PEARL-ft is not able to make learning progress.
    }
    \label{fig:maze_qual}
\end{figure}

%% file: sections/05_experiments.tex
\vspacesection{Experiments}
\label{sec:experiments}

Our experiments aim to answer the following questions:
(1)~Can our proposed method learn to efficiently 
solve long-horizon, sparse reward tasks? 
(2)~Is it crucial to utilize offline datasets to achieve this?
(3)~How can we best leverage the training tasks for efficient learning of target tasks?
(4)~How does the training task distribution affect the target task learning?

\vspacesubsection{Experimental Setup}

We evaluate our approach in two challenging continuous control environments: maze navigation and kitchen manipulation environment,
as illustrated in Figure~\ref{fig:env_fig}. 
While meta-RL algorithms are typically evaluated on tasks that span only a few dozen time steps and provide dense rewards~\citep{finn2017model,rothfuss2018promp,rakelly2019efficient, varibad}, our tasks require to learn long-horizon behaviors over hundreds of time steps from sparse reward feedback and thus pose a new challenge to meta-learning algorithms.

\input{fig/env_fig}
\vspacesubsubsection{Maze Navigation}
\label{sec:maze}

\noindent\textbf{Environment.} 
This 2D maze navigation domain based on the maze navigation problem in~\citet{fu2020d4rl}
requires long-horizon control with hundreds of steps for a successful episode and only provides sparse reward feedback upon reaching the goal. 
The observation space of the agent consists of its 2D position and velocity and it acts via planar, continuous velocity commands.

\noindent\textbf{Offline Dataset \& Meta-training / Target Tasks.} Following~\citet{fu2020d4rl} we collect a task-agnostic offline dataset by randomly sampling start-goal locations in the maze and using a planner to generate a trajectory that reaches from start to goal. 
Note that the trajectories are not annotated with any reward or task labels (\ie which start-goal location is used for producing each trajectory).
To generate a set of meta-training and target tasks,
we fix the agent's initial position in the center of the maze and sample 
\SI{40}{}
random goal locations for meta-training 
and another set of \SI{10}{} goals for target tasks.
All meta-training and target tasks use the same sparse reward formulation.
More details can be found in~\mysecref{sec:app_maze_task_dist}.

\vspacesubsubsection{Kitchen Manipulation}
\label{sec:kitchen}

\noindent\textbf{Environment.} 
The FrankaKitchen environment of~\citet{gupta2019relay} requires the agent to control a 7-DoF robot arm via continuous joint velocity commands and complete a sequence of manipulation tasks like opening the microwave or turning on the stove.
Successful episodes span 300-500 steps and the agent is only provided a sparse reward signal upon successful completion of a subtask.

\noindent\textbf{Offline Dataset \& Meta-training / Target Tasks.} We leverage a dataset of 600 human-teleoperated manipulation sequences of~\citet{gupta2019relay} for offline pre-training. 
In each trajectory, the robot executes a sequence of four subtasks. 
We then define a set of 23 meta-training tasks 
and 10 target tasks that in turn require 
the consecutive execution of four subtasks 
(see Figure~\ref{fig:env_fig} for examples).
Note that each task consists of a unique combination of subtasks.
More details can be found in~\mysecref{sec:app_kitchen_task_dist}.

\vspacesubsection{Baselines} 
\label{sec:baselines}
We compare 
\method\\
to prior approaches
in RL, skill-based RL, meta-RL, and multi-task RL.
\begin{itemize}
    \item \textbf{SAC}~\citep{haarnoja2018sac} 
    is a state of the art deep RL algorithm.
    It learns to solve a target task from scratch
    without leveraging the offline dataset nor the meta-training tasks.
    
    \item \textbf{SPiRL}~\citep{pertsch2020spirl} 
    is a method designed to leverage offline data through the transfer of learned skills. 
    It acquires skills and a skill prior from the offline dataset
    but does not utilize the meta-training tasks.
    This investigates the benefits our method can obtain from leveraging the meta-training tasks. 

    \item \textbf{PEARL}~\citep{rakelly2019efficient} 
    is a state of the art off-policy meta-RL algorithm that 
    learns a policy which can quickly adapt to unseen test tasks.
    It learns from the meta-training tasks but does not use the offline dataset.
    This examines the benefits of using learned skills in meta-RL.
    
    \item \textbf{PEARL-ft} 
    demonstrates the performance of 
    a PEARL~\citep{rakelly2019efficient} model further fine-tuned on a target task
    using SAC~\citep{haarnoja2018sac}.
    
    \item 
    \textbf{Multi-task RL (MTRL)}
    is a multi-task RL baseline
    which learns from the meta-training tasks
    by distilling individual policies specialized in each task
    into a shared policy, similar to Distral~\citep{distral}.
    Each individual policy is trained using SPiRL by leveraging skills extracted from the offline dataset. 
    Therefore, it utilizes both the meta-training tasks and offline dataset similar to our method.
    This provides a direct comparison of 
    multi-task learning (MTRL) from the training tasks
    vs. meta-learning using them (ours).
\end{itemize}

More implementation details on the baselines can be found in~\mysecref{sec:app_baseline}.

\vspacesubsection{Results}

We present 
the quantitative results in~\myfig{fig:main_curve} 
and the qualitative results on the maze navigation domain in~\myfig{fig:maze_qual}.
In~\myfig{fig:main_curve}, 
\method\\ demonstrates much better sample efficiency 
for learning the unseen target tasks compared to all the baselines.
Without leveraging the offline dataset and meta-training tasks,
SAC is not able to make learning progress on most of the target tasks. 
While PEARL is first trained on the meta-training tasks,
it still achieves poor performance on the target tasks
and fine-tuning it (PEARL-ft) does not yield significant improvement.
We believe this is because both environments provide only sparse rewards 
yet require the model to exhibit long-horizon and complex behaviors,
which is known to be difficult for meta-RL methods~\citep{mitchell2021offline}.

\input{fig/main_curve}
\input{fig/maze_qual}
On the other hand, 
by first extracting skills and acquiring a skill prior from the offline dataset,
SPiRL's performance consistently improves with more samples from the target tasks.
Yet, it requires significantly more environment interactions 
than our method to solve the target tasks
since the policy is optimized using vanilla RL, 
which is not designed to learn to quickly learn new tasks.
While the multi-task RL (MTRL) baseline first learns a multi-task policy 
from the meta-training tasks,
its sample efficiency is similar to SPiRL on target task learning,
which highlights the strength of our proposed method -- meta-learning from the meta-training tasks for fast target task learning.

Compared to the baselines, our method learns the target tasks much quicker.
Within only a few episodes the policy converges to solve 
more than $80\%$ 
of the target tasks in the maze environment and two out of four subtasks in the kitchen manipulation environment. 
The prior-regularized fine-tuning then continues to improve performance. 
The rapidly increasing performance and the overall faster convergence show the benefits of leveraging meta-training tasks in addition to learning from offline data: 
by first learning to learn how to quickly solve tasks using the extracted skills and the skill prior, 
our policy can efficiently solve the target tasks.

The qualitative results presented in~\myfig{fig:maze_qual} 
show that all the methods that leverage the offline dataset 
(\ie \method\\, SPiRL, and MTRL) effectively 
explore the maze in the first episode.
Then, \method\\ converges with much fewer episodes compared to 
SPiRL and MTRL,
underlining the effectiveness of meta-training.
In contrast, PEARL-ft is not able to make learning progress,
justifying the necessity of employing offline datasets
for acquiring long-horizon, complex behaviors.

\Skip{
\input{fig/maze_encoding}

\noindent\textbf{Task Encoding Visualization.}
}

\vspacesubsection{Meta-training Task Distribution Analysis}
In this section, we aim to investigate the effect of the meta-training task distribution on 
our skill-based meta-training and target task learning phases.
Specifically, we examine the effect of
(1) the number of tasks in the meta-training task distribution
and 
(2) the alignment between a meta-training task distribution and target task distribution.
We conduct experiments and analyses in the maze navigation domain.
More details on task distributions can be found in~\mysecref{sec:app_maze_task_dist}.

\noindent\textbf{Number of meta-training tasks.}
To investigate how the number of meta-training tasks affects the performance of
our method,
we train our method
with fewer numbers meta-training tasks (\ie 10 and 20)
and evaluate it with the same set of target tasks.
The quantitative results presented in~\myfig{fig:maze_density} suggest that 
even with sparser meta-training task distributions 
(\ie fewer numbers of meta-training tasks),
\method\\ is still more sample efficient compared to 
the best-performing baseline (\ie SPiRL).

\input{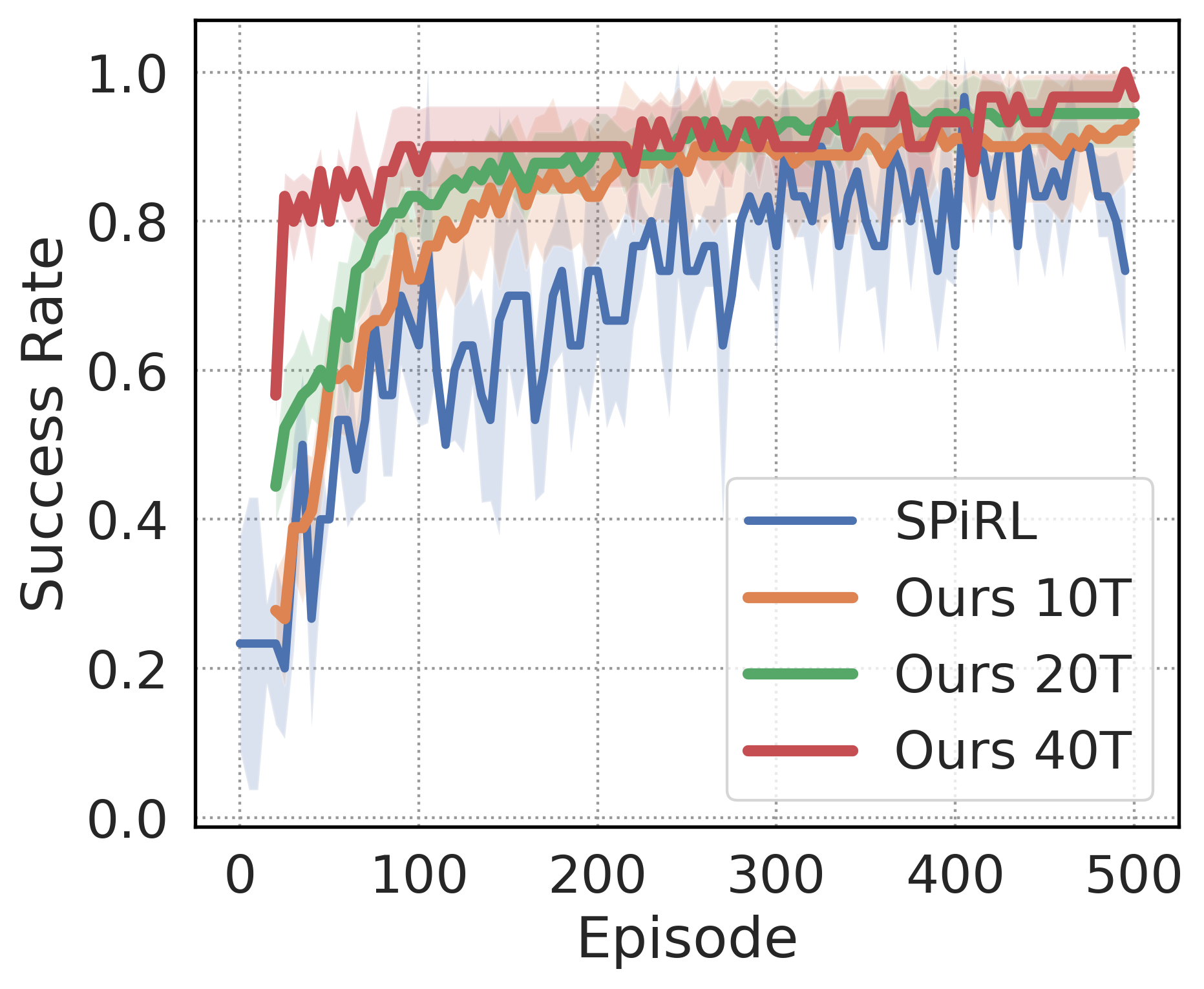}

\noindent\textbf{Meta-train / test task alignment.}
We aim to examine if a model trained on a meta-training task distribution
that aligns better/worse with the target tasks
would yield improved/deteriorated performance.
To this end, 
we create biased meta-training / test task distributions: we create a meta-train set by sampling goal locations
from only the top 25\% portion of the maze ($\mathcal{T}_{\textsc{Train-Top}}$).
To rule out the effect of the density of the task distribution,
we sample 10 (\ie 40 $\times$ 25\%) meta-training tasks. Then, we create two target task distributions that have good and bad alignment with this meta-training distribution respectively by sampling \SI{10}{} target tasks from the top 25\% portion of the maze
($\mathcal{T}_{\textsc{Target-Top}}$) and \SI{10}{} target tasks from the bottom 25\% portion of the maze ($\mathcal{T}_{\textsc{Target-Bottom}}$).

\myfig{fig:maze_top} and \myfig{fig:maze_bottom} 
present the target task learning efficiency for models trained 
with good task alignment 
(meta-train on $\mathcal{T}_{\textsc{Train-Top}}$, 
learn target tasks from $\mathcal{T}_{\textsc{Target-Top}}$) 
and bad task alignment 
(meta-train on $\mathcal{T}_{\textsc{Train-Top}}$, 
learn target tasks from $\mathcal{T}_{\textsc{Target-Bottom}}$), 
respectively. 
The results demonstrate that \method\\ can achieve improved performance when trained on a better aligned meta-training task distribution.
On the other hand, not surprisingly, \method\\ and MTRL perform slightly worse compared to SPiRL 
when trained with misaligned meta-training tasks (see~\myfig{fig:maze_bottom}). 
This is expected given that SPiRL does not learn from the misaligned meta-training tasks. 
In summary, 
from~\myfig{fig:task_dist_analysis}, 
we can conclude that
meta-learning from either a diverse task distribution
or a better informed task distribution
can yield improved performance for our method.

\Skip{
\vspacesubsection{Ablation Study}

\input{fig/maze_ablation}

To justify our design choices, 
we ablate the following variations of our method.

\begin{itemize}
    \item \method\\ (share weight) 
    imposes the same KL coefficient $\alpha$ to regularize the high-level policy during meta-training, instead of using two coefficients of differing magnitude based on the size of the conditioning sample set $\tau$ (see Section~\ref{sec:meta-training}).
    \item \method\\ (uniform prior) 
    leverages the offline dataset 
    by learning a low-level skill policy but not a skill prior.
    During the meta-training phase,
    it does not regularize the high-level policy using a learned skill prior;
    instead, it leverages a uniform prior over skills, similar to PEARL.
    This investigates the importance of acquiring a skill prior from the offline dataset 
    and using it to regularize meta-training.
\end{itemize}

The results presented in~\myfig{fig:maze_ablation} 
justify our design choices 
-- both regularizing the two policies using different KL coefficients 
and the skill prior learned from the offline dataset 
are crucial for the superior performance of our method. 
}

%% file: fig/maze_task_dist.tex
\begin{figure}
\centering
\subfigure[Sparser Task Distribution]
    {\label{fig:maze_density}\includegraphics[width=0.32\textwidth]{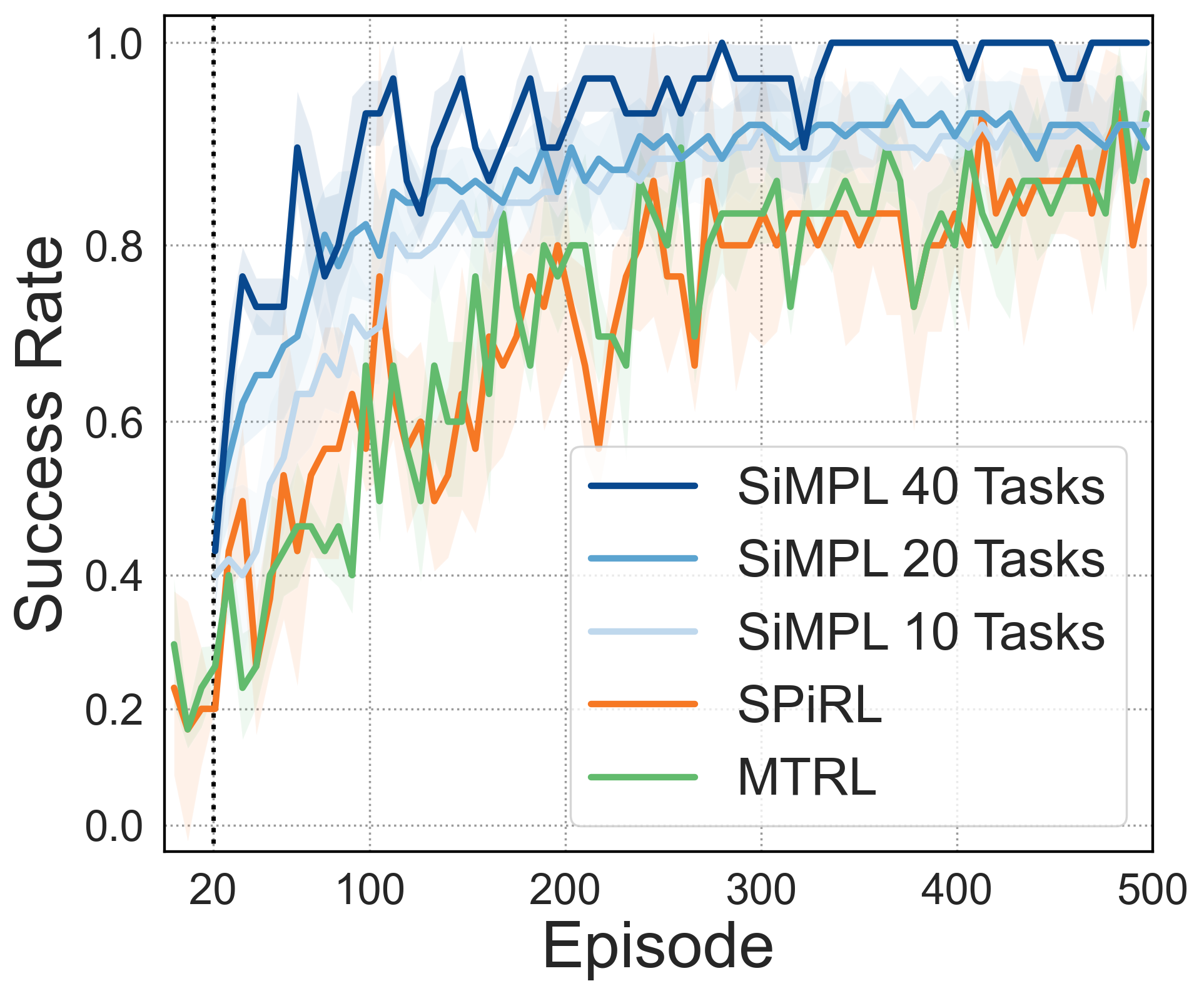}}
\subfigure[$\mathcal{T}_{\textsc{Train-Top}}\rightarrow \mathcal{T}_{\textsc{Target-Top}}$]
    {\label{fig:maze_top}\includegraphics[width=0.32\textwidth]{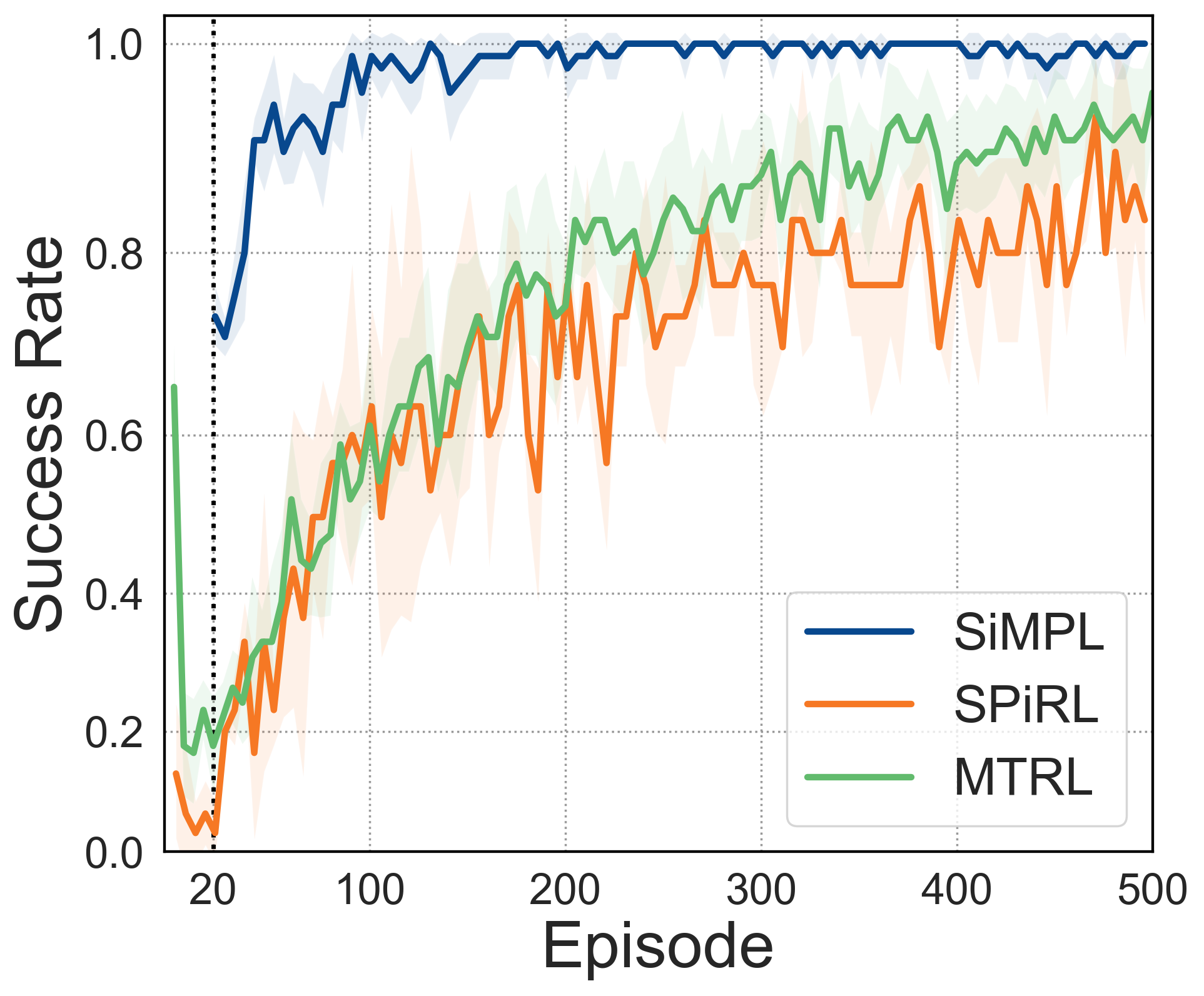}}
\subfigure[$\mathcal{T}_{\textsc{Train-Top}}\rightarrow \mathcal{T}_{\textsc{Target-Bottom}}$]
    {\label{fig:maze_bottom}\includegraphics[width=0.32\textwidth]{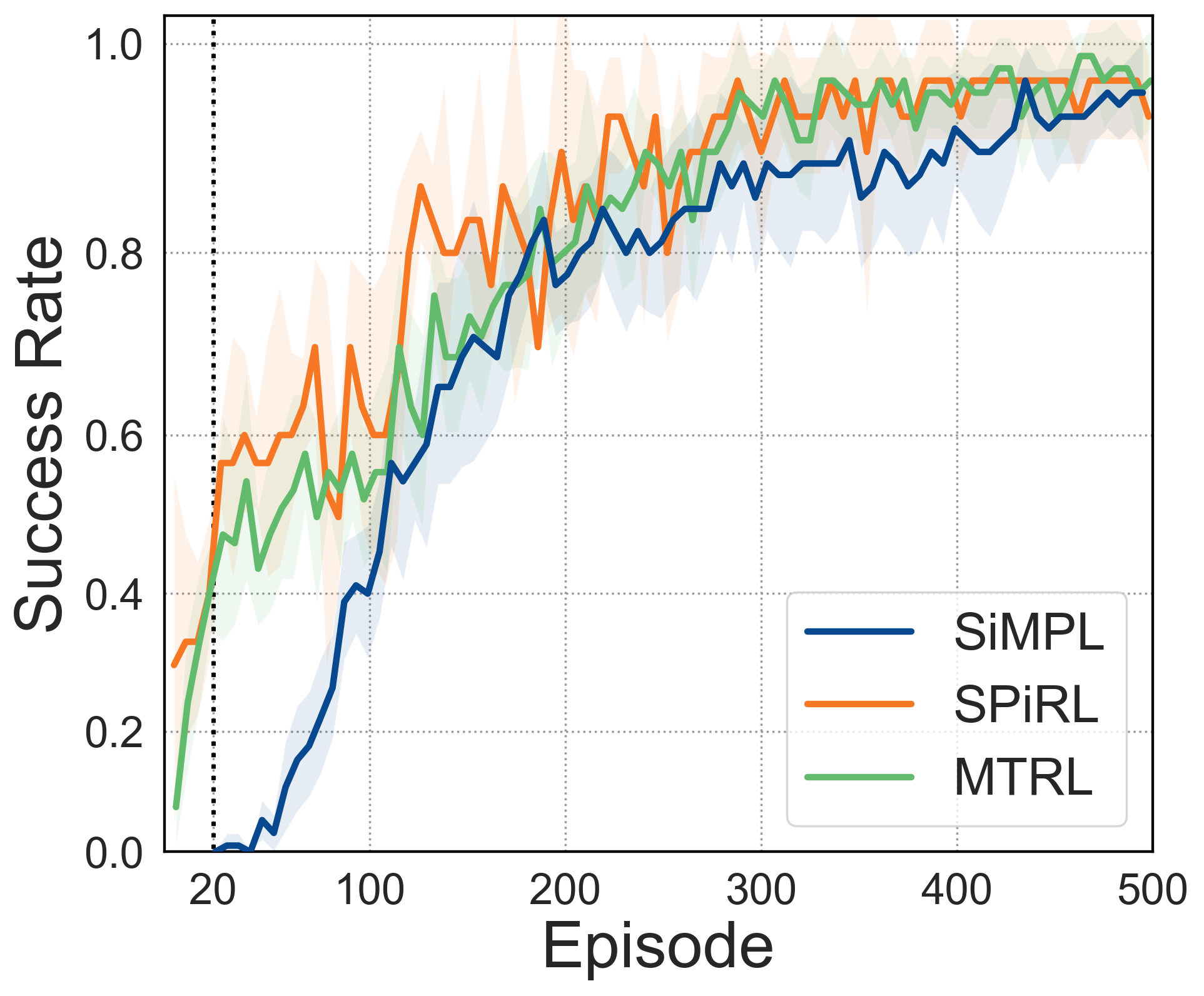}}
    \vspace{-0.3cm} 
\caption{\small
\textbf{Meta-training Task Distribution Analysis.}
(a) With sparser meta-training task distributions 
(\ie fewer numbers of meta-training tasks),
\method\\ still achieves better sample efficiency compared to SPiRL,
highlighting the benefit of leveraging meta-training tasks.
(b) When trained on a meta-training task distribution 
that aligns better with the target task distribution,
\method\\ achieves improved performance.
(c) When trained on a meta-training task distribution 
that is mis-aligned with the target tasks,
\method\\ yields worse performance.
For all the analyses,
we train each model on each target task with 3 different random seeds.
\label{fig:task_dist_analysis}
}
\end{figure}

%% file: fig/maze_ablation.tex
\begin{wrapfigure}{R}{0.37\textwidth}
  \begin{center}
    \includegraphics[width=0.35\textwidth]{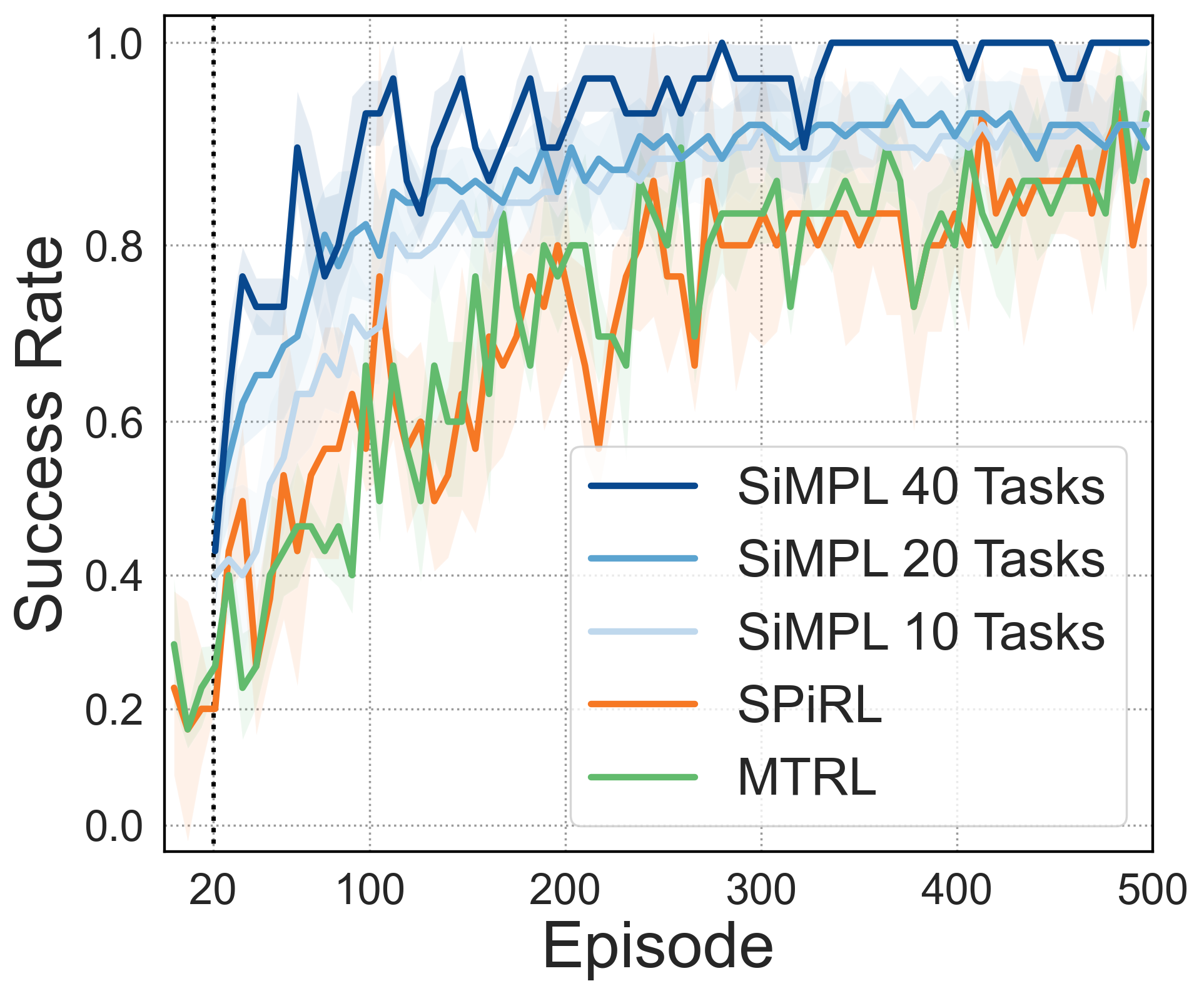}
    \vspace{-0.4cm}    
  \end{center}
  \caption{\small
        \textbf{Ablation Study.}
        The two ablations of our method yield worse performance,
        justifying the effectiveness of
        regularizing with two different KL coefficients during meta-training
        and using the learned skill prior acquired from the offline dataset.
	    \label{fig:maze_ablation}
    }
\end{wrapfigure}

%% file: sections/06_conclusion.tex
\vspacesection{Conclusion}
\label{sec:conclusion}
We propose a skill-based meta-RL method, dubbed \method\\, that can meta-learn on long-horizon tasks by leveraging prior experience in the form of large offline datasets without additional reward and task annotations. Specifically, our method first learns to extracts reusable skills and a skill prior from the offline data.
Then, we propose to meta-trains a high-level policy that leverages these skills for efficient learning of unseen target tasks. 
To effectively utilize learned skills, the high-level policy is regularized by the acquired prior.
The experimental results on challenging continuous control long-horizon navigation and manipulation tasks with sparse rewards demonstrate that our method outperforms the prior approaches in deep RL, skill-based RL, meta-RL, and multi-task RL. In the future, we aim to demonstrate the scalability of our method to high-DoF continuous control problems on real robotic systems to show the benefits of our improved sample efficiency.

%% file: sections/07_ack.tex
\section*{Acknowledgments}
This work was supported by the Engineering Research Center Program through the National Research Foundation of Korea (NRF) funded by the Korean Government MSIT (NRF-2018R1A5A1059921), KAIST-NAVER Hypercreative AI Center, and Institute of Information \& communications Technology Planning \& Evaluation (IITP) grant funded by the Korea government (MSIT)  (No.2019-0-00075, Artificial Intelligence Graduate School Program (KAIST)).
The authors are grateful for the fruitful discussion with the members of USC CLVR lab.

%% file: sections/appendix.tex
\appendix
\section*{Appendix}

\begingroup
\hypersetup{linkcolor=black}

\part{} %
\parttoc[t] %
\endgroup

\ifx\venue\metalearn
   \input{sections_metalearn/appendix_task_dist}
\fi

\ifx\venue\iclr
    \input{sections/appendix_mrl_ablation}
    \input{sections/appendix_few_eps}
    \input{sections/appendix_image_based_maze}
\fi

\section{Extended Related Work}
\label{sec:extended_related_work}

We present an extended discussion of the related work in this section.

\noindent\textbf{Pre-training in Meta-learning.}
Leveraging pre-trained models for improving meta-learning methods 
has been explored in~\cite{bronskill2021memory,dvornik2020selecting,kolesnikov2020big,triantafillou2019meta} 
with a focus on few-shot image classification.
One can also view our proposed framework as a
meta-reinforcement learning method with a pre-training phase.
Specifically, 
in the pre-training phase,
we propose to first 
extract reusable skills and a skill prior from offline datasets
without reward or task annotations in a self-supervised fashion.
Then, our proposed method meta-learns from a set of meta-training tasks,
which significantly accelerates learning 
on unseen target tasks.

\section{Implementation Details on Our Method}
\label{sec:implementation_details}

In this section, we describe the additional implementation details 
on our proposed method.
The details on model architecture is presented in~\mysecref{sec:arch},
followed by the training detailed described in~\mysecref{sec:training}.

\subsection{Model architecture}
\label{sec:arch}

We describe the details on our model architecture in this section.

\subsubsection{Skill Prior}
We followed architecture and learning procedure of~\citet{pertsch2020spirl} 
for learning a low-level skill policy and a skill prior.
Please refer to ~\citet{pertsch2020spirl} for more details
on the architectures for learning skills and skill priors from offline datasets.

\subsubsection{Task Encoder}
Following~\citet{rakelly2019efficient}, 
our task encoder is a permutation invarient neural network.
Specifically, we adopt Set Transformer~\citep{lee2019set}
that consists of layers $[2 \times \mathrm{ISAB}_{32}  \rightarrow \mathrm{PMA}_1 \rightarrow  3 \times \mathrm{MLP}]$ 
for expressive and efficient set encoding.
All the hidden layers are $128$-dimensional and all attention layers have $4$ attention heads.
The encoder takes a set of high-level transitions as input,
where each transition is a vector concatenation of high-level transition tuple.
The output of the encoder is $(\mu_{e}, \sigma_{e})$ 
which are the parameters of Gaussian task posterior
$p(e|c)=\mathcal{N}(e;\mu_{e}, \sigma_{e})$.
We varied task vector dimension $\mathrm{dim}(e)$ depends on task distribution complexity.
$\mathrm{dim}(e)=10$ for Kitchen Manipulation, $\mathrm{dim}(e)=6$ for Maze Navigation with 40 meta-training tasks, and $\mathrm{dim}(e)=5$, otherwise.

\subsubsection{Policy}
\label{sec:policy_arch}
We parameterize our policy with neural network.
We employed $4$-layer MLPs with $256$ hidden units for Maze Navigation, and $6$-layer MLPs with $128$ hidden unit for Kitchen Manipulation experiment.
Instead of direct parameterization of policy, 
the network output is added to skill-prior to make learning more stable.
Specifically, the policy network takes concatenation of $(s, e)$ as input, 
and then outputs residual parameters $(\mu_z, \log \sigma_z)$ to skill-prior distribution $p(z \vert s) = \mathcal{N}(z|\mu_p, \sigma_p)$.
Resulting distribution by this residual parameterization is $\pi(z \vert s) = \mathcal{N}(z \vert \mu_p + \mu_z, \exp(\log \sigma_p + \log \sigma_z))$

\subsubsection{Critic}
The critic network takes concatenation of $s$, $e$, and skill $z$ as input
and outputs an estimation of task-conditioned Q-value $Q(s, z, e)$.
We employ double Q networks~\citep{van2016deep} to mitigate Q-value overestimation.
The architecture of critic follows the policy.

\subsection{Training details}
\label{sec:training}
For all the network updates, 
we used Adam optimizer~\citep{kingma2014adam} 
with a learning rate of $3e-4$, $\beta_1=0.9$, and $\beta_2=0.999$.
We describe the training details of 
the skill-based meta-training phase in~\mysecref{sec:training_phase2} and 
the target task learning phase~\mysecref{sec:training_phase3}.

\subsubsection{Skill-based Meta-training}
\label{sec:training_phase2}
Our meta-training procedure 
is similar to the procedure adopted in~\citep{rakelly2019efficient}.
Encoder and critics networks are updated to minimize MSE between Q-value prediction and target Q value.
Policy network is updated to optimize Equation \ref{eq:ours_objective_metarl} without updating the encoder network.
All network are updated with the average gradients of 20 randomly sampled tasks.
Each batch of gradients is computed from $1024$ and $256$ transitions for Maze Navigation and Kitchen Manipulation experiment, respectively.
We train our models for $10000$, $18000$, and $16000$ episodes for 
the Maze Navigation experiments with $10$, $20$, $40$ meta-training tasks, respectively, and $3450$ episodes for Kitchen Manipulation.

As stated in \mysecref{sec:meta-training}, 
we apply different regularization coefficients depending on 
the size of the conditioning transitions.
In Maze Navigation experiment, we set target KL divergence to $0.1$ for batch that is conditioned on size $4$ transitions and $0.4$ for batch conditioned on size $8192$ transitions.
In Kitchen Manipulation experiment, we set target KL divergence to $0.4$ for batch conditioned with a size $1024$ transitions while KL coefficient for batch conditioned on size $2$ transitions is fixed to $0.3$.

\subsubsection{Target Task Learning}
\label{sec:training_phase3}
We initialize the Q function and the auto-tuning value $\alpha$ with 
the values that learned in the skill-based meta-training phase.
The policy is initialized after observing and encoding $20$ episodes 
obtained from the task unconditioned policy rollouts.
For the target task learning phase,
the target KL $\delta$ is $1$ for Maze Navigation, and $2$ for Kitchen Manipulation experiments.
To compute a gradient step, $256$ high-level transitions are sampled 
from a replay buffer with size $20000$.
Note that we used same setup for baselines that uses SPiRL fine-tuning (SPiRL and MTRL).

\section{Implementation Details on Baselines}
\label{sec:app_baseline}
In this section, we describe the additional implementation details 
on producing the results of the baselines.

\subsection{SAC}
The SAC~\citep{haarnoja2018sac} baseline learns to solve a target task from scratch 
without leveraging the offline dataset nor the meta-training tasks.

We initialize $\alpha$ to $0.1$ and 
set the target entropy to $\mathcal{H} = -\mathrm{dim}(\mathcal{A})$.
To compute a gradient step, $4096$ and $1024$ environment transitions 
are sampled from a replay buffer for Maze Navigation and Kitchen Navigation experiments, respectively.

\subsection{PEARL and PEARL-ft}
PEARL~\citep{rakelly2019efficient} learns from the meta-training tasks
but does not use the offline dataset. 
Therefore,
we directly train PEARL models on the meta-training tasks 
without the phase of learning from offline datasets.
We use gradients averaged from 20 randomly sampled tasks 
where each task gradient is computed by batch sampled from a per-task buffer.
The target entropy is set to $\mathcal{H} = -\mathrm{dim}(\mathcal{A})$ 
and $\alpha$ is initialized to $0.1$.

While the method proposed in~\citet{rakelly2019efficient} 
does not fine-tune on target/meta-testing tasks,
we extend PEARL to be fine-tuned on target tasks for a fair comparison, called PEARL-ft.
Since PEARL does not use learned skills or a skill prior,
the target task learning of PEARL is simply running SAC with task-encoded initialization.
Similar to the target task learning of our method,
we initialize the Q function and entropy coefficient $\alpha$ to 
the value learned during the meta-training phase.
Also, we initialize the policy to the task conditioned policy 
after observing $20$ episodes of experience from the task unconditioned policy rollouts.
The hyperparameters used for fine-tuning are the same as SAC.

\subsection{SPiRL}
Similar to our method,
we initialize the high-level policy to skill-prior 
while fixing low-level policy for target task learning for SPiRL.
$\alpha$ is initialized to $0.01$ 
and we use the same hyperparameters for the SPiRL models as our method.

\subsection{Multi-task RL (MTRL)}
Inspired by Distral~\citep{distral}, 
our multi-task RL baseline is designed to
first learns a set of individual policies, where
each of them is specialized in one task;
then, a shared/multi-task policy is learned by distilling the individual polices.
Since it is inefficient to learn an individual policy from scratch,
we learn each individual policy using SPiRL with learned skills and a skill prior.
Then, we distill the individual policies using the following objective :

\begin{equation}
\label{eq:mtrl_objective}
    \max_{\pi_0} \mathbb{E}_{\mathcal{T} \sim p_{\mathcal{T}}} \bigg[
        \sum_{t}
            \mathbb{E}_{
                (s_t, z_t) \sim \rho_{\pi_0}
            } \big[
                r_{\mathcal{T}}(s_t, z_t) -
                \alpha D_\text{KL}\big(\pi_0(z \vert s_t, e), p(z \vert s_t)\big)
            \big]
    \bigg].
\end{equation}

We use the same setup for $\alpha$ as our method, where $\alpha$
is auto-tuned to satisfy a target KL, $\delta=0.1$ for Maze Navigation and $\delta=0.2$ for Kitchen Manipulation.

While the target task learning phase for MTRL is similar to ours,
except that MTRL is not initialized with a meta-trained Q function and learned $\alpha$.

\section{Meta-training Tasks and Target Tasks.}

In this section, 
we present the meta-training tasks and target tasks
used in the maze navigation domain and the kitchen manipulation domain.

\subsection{Maze Navigation}
\label{sec:app_maze_task_dist}

\begin{figure}
\centering
\subfigure[Meta-training 40 Tasks]
    {\label{fig:maze_40_dist}\includegraphics[width=0.24\textwidth]{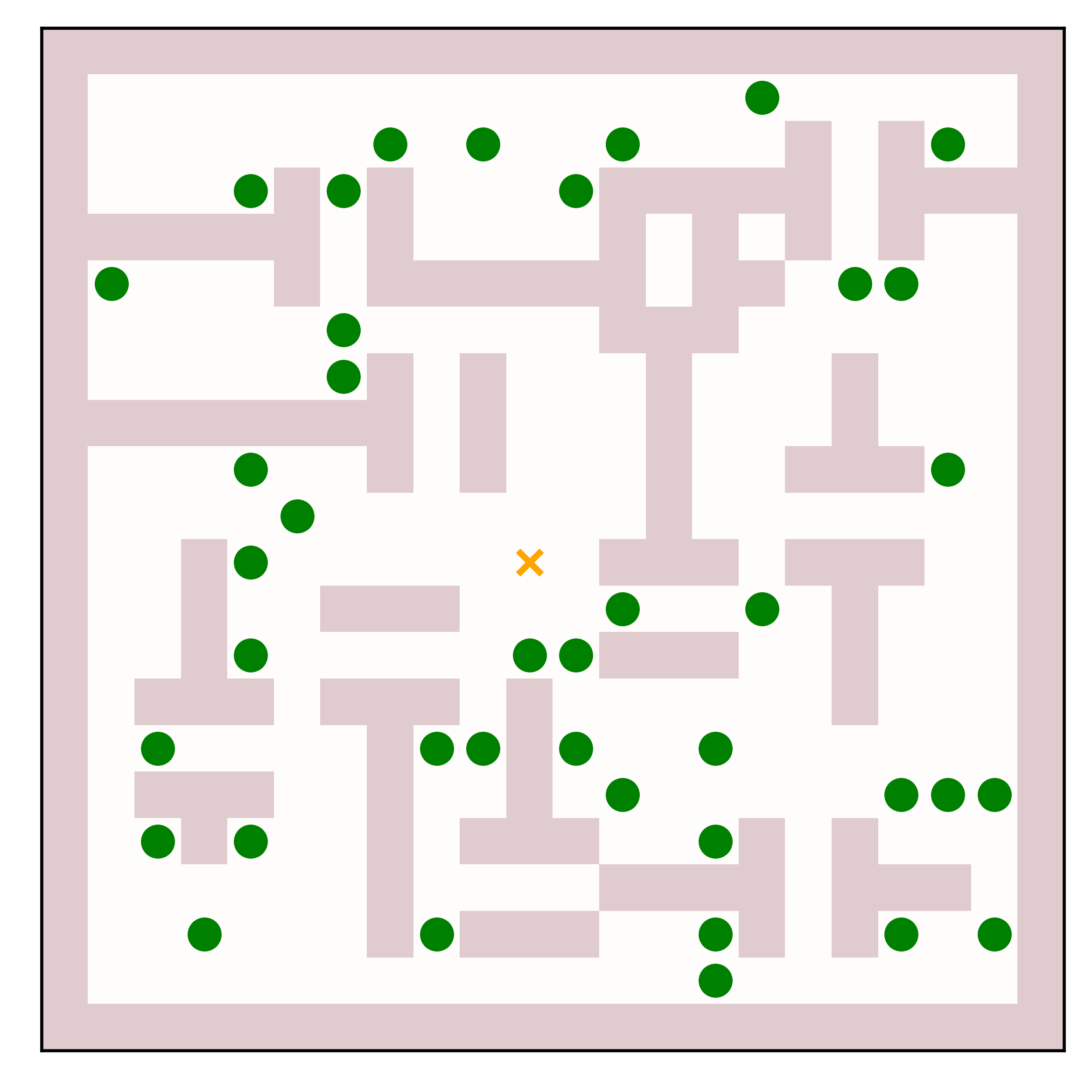}}    
    \hfill
\subfigure[Meta-training 20 Tasks]
    {\label{fig:maze_20_dist}\includegraphics[width=0.24\textwidth]{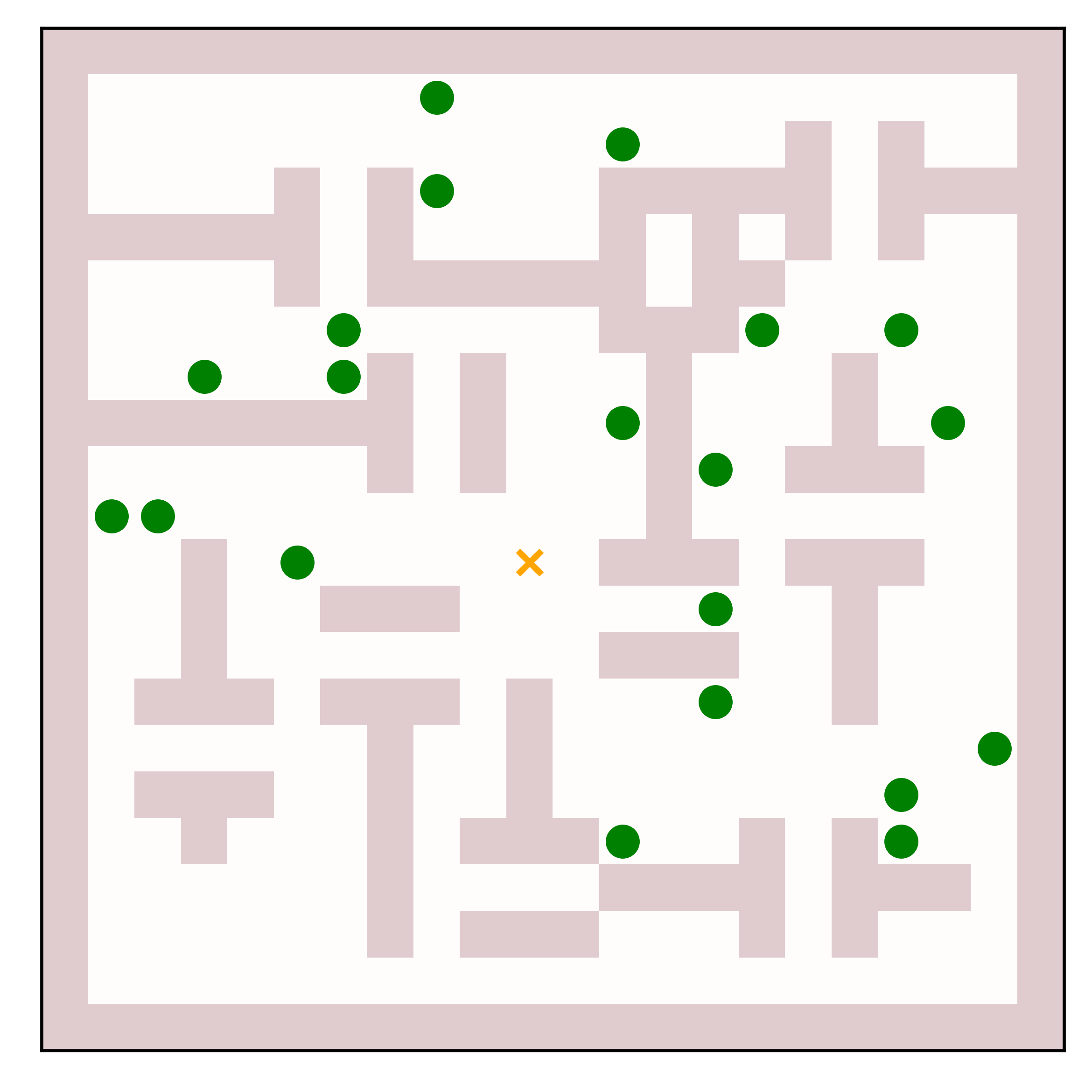}}    
    \hfill
\subfigure[Meta-training 10 Tasks]
    {\label{fig:maze_10_dist}\includegraphics[width=0.24\textwidth]{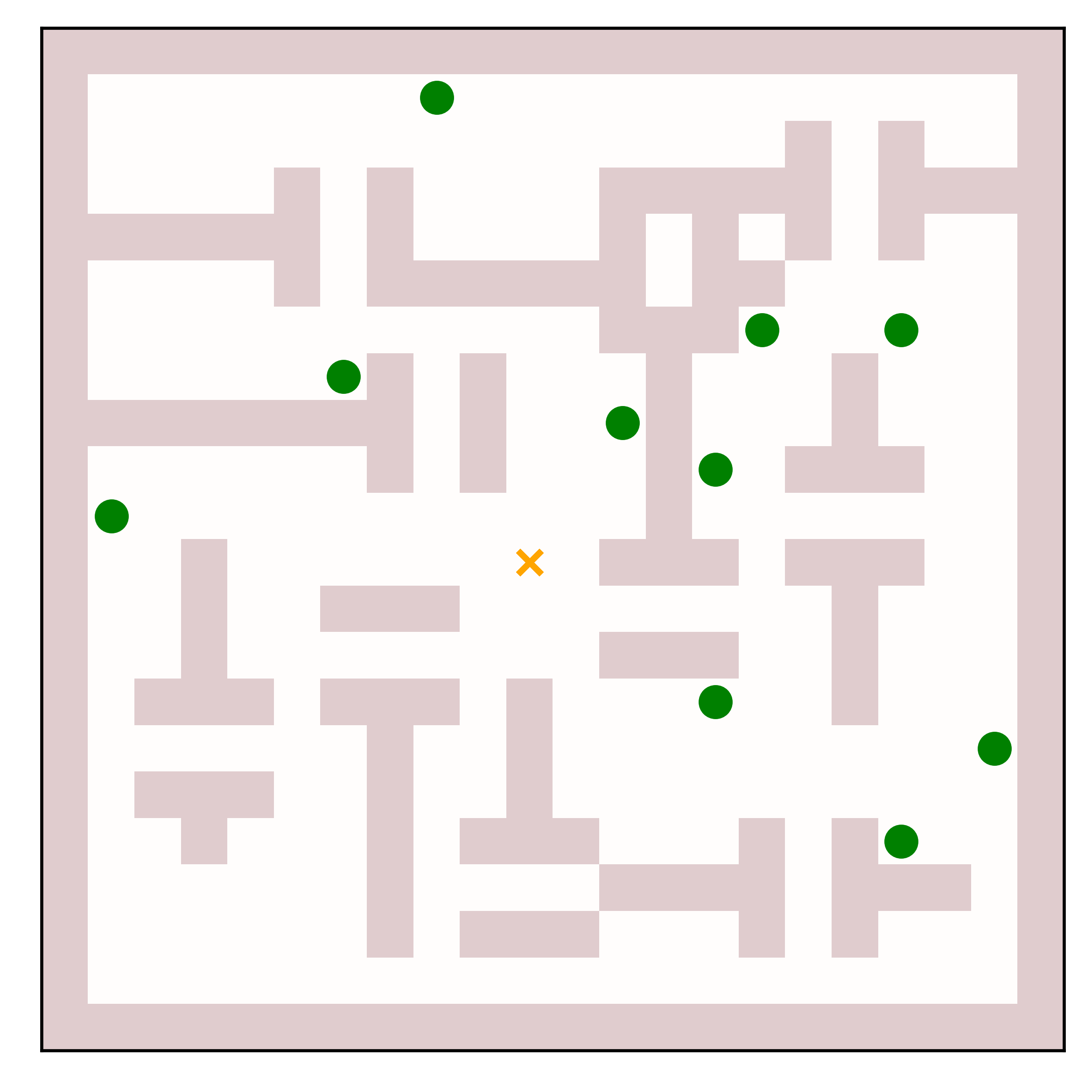}}
    \hfill
\subfigure[Target Tasks]
    {\label{fig:maze_target_dist}\includegraphics[width=0.24\textwidth]{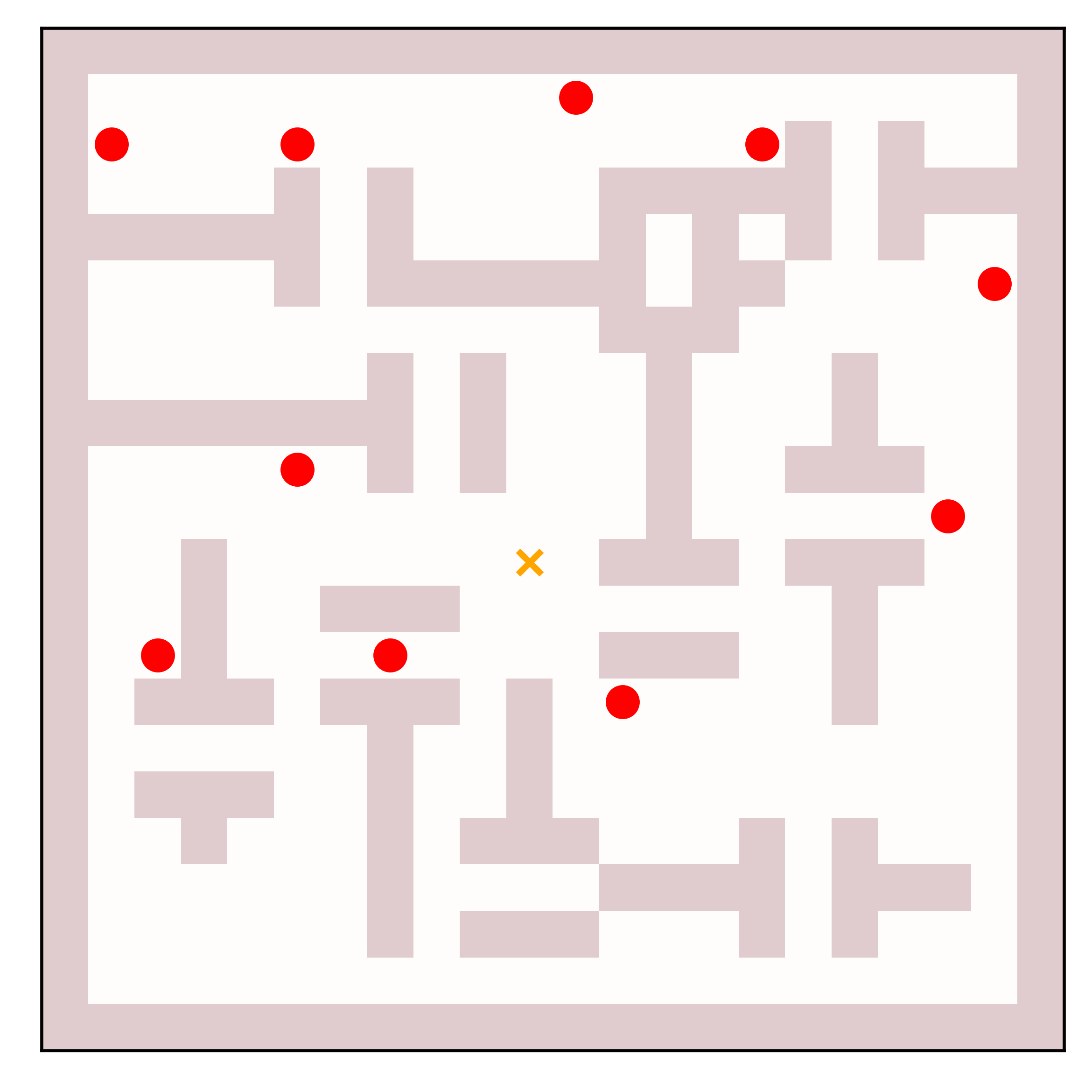}}
\caption{\small
\textbf{Maze Meta-training and Target Task Distributions.}
The green dots represent the goal locations of meta-training tasks
and the red dots represent the goal locations of target tasks.
The yellow cross represent the initial location of the agent.
\label{fig:maze_main_task_dist}
}
\end{figure}

\begin{figure}
\centering
\subfigure[$\mathcal{T}_{\textsc{Train-Top}}$]
    {\label{fig:maze_train_top_dist}\includegraphics[width=0.24\textwidth]{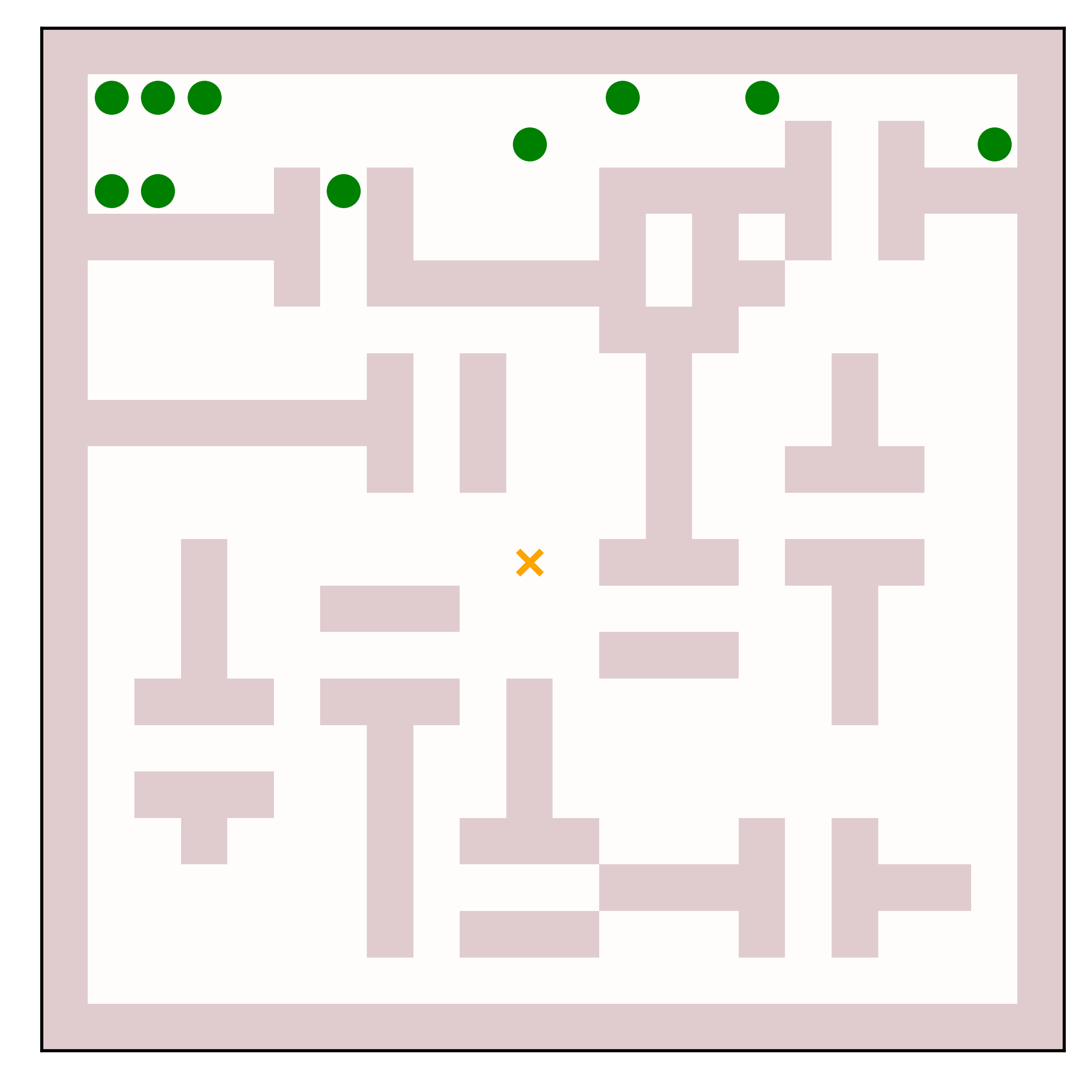}}    
\subfigure[$\mathcal{T}_{\textsc{Target-Top}}$]
    {\label{fig:maze_target_top_dist}\includegraphics[width=0.24\textwidth]{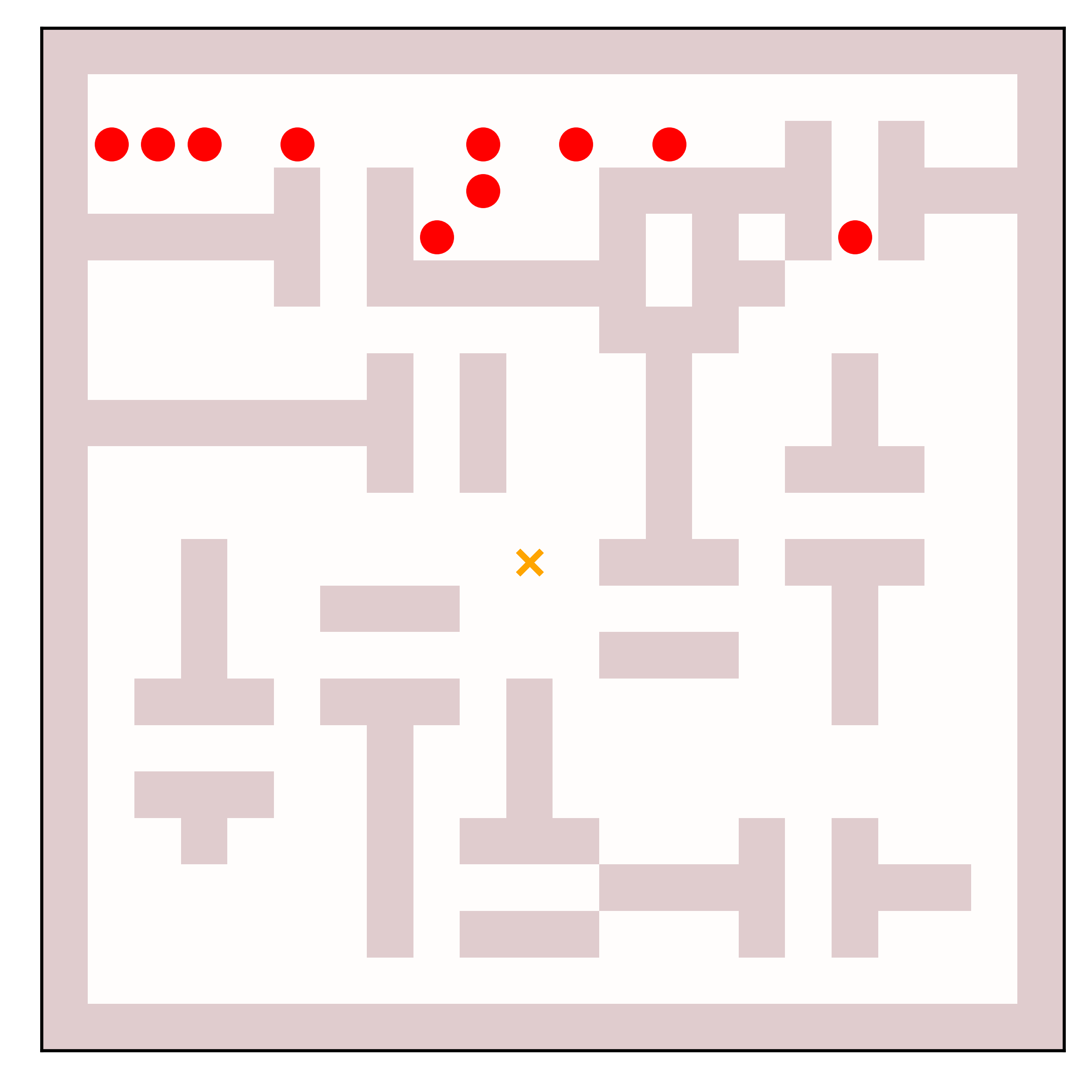}}
\subfigure[$\mathcal{T}_{\textsc{Target-Bottom}}$]
    {\label{fig:maze_target_bottom_dist}\includegraphics[width=0.24\textwidth]{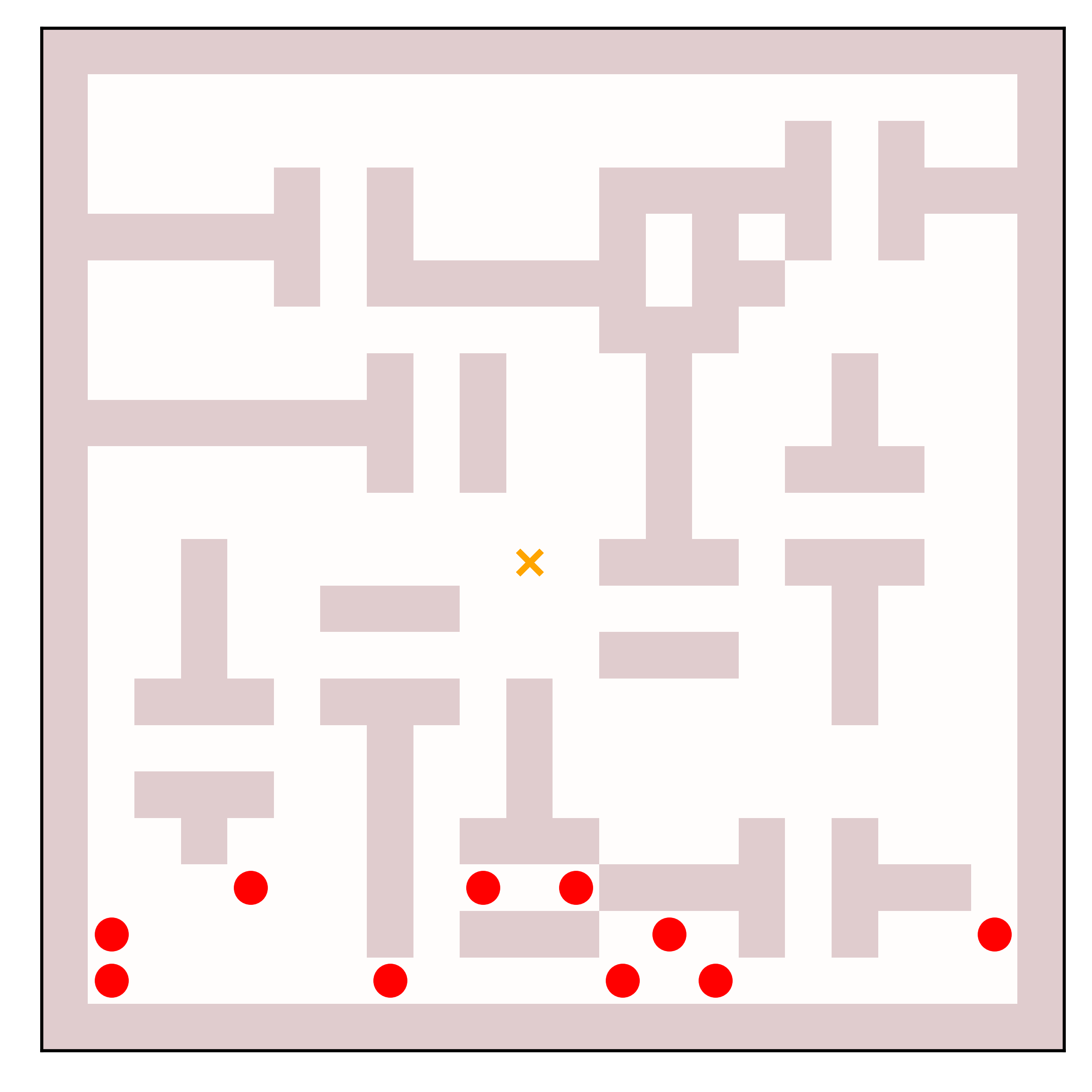}}
\caption{\small
\textbf{Maze Meta-training and Target Task Distributions for Meta-training Task Distribution Analysis.}
The green dots represent the goal locations of meta-training tasks
and the red dots represent the goal locations of target tasks.
The yellow cross represent the initial location of the agent.
\label{fig:maze_task_dist_analysis_dist}
}
\end{figure}

The meta-training tasks and target tasks are visualized in
\myfig{fig:maze_main_task_dist} and \myfig{fig:maze_task_dist_analysis_dist}.

\subsection{Kitchen Manipulation}
\label{sec:app_kitchen_task_dist}

\Skip{
\begin{compactwrapfigure}[16]{R}{0.48\textwidth}
  \begin{center}
    \includegraphics[width=0.45\textwidth]{fig/kitchen_task_resplit.png}
  \end{center}
  \caption{\small
        \textbf{Kitchen Meta-training Tasks and Target Tasks.}
	    \label{fig:kitchen_task_split}
    }
\end{compactwrapfigure}
}

The meta-training tasks are:
\begin{itemize}
\item \texttt{microwave}$\rightarrow$\texttt{kettle}$\rightarrow$\texttt{bottom burner}$\rightarrow$\texttt{slide cabinet}
\item \texttt{microwave}$\rightarrow$\texttt{bottom burner}$\rightarrow$\texttt{top burner}$\rightarrow$\texttt{slide cabinet}
\item \texttt{microwave}$\rightarrow$\texttt{top burner}$\rightarrow$\texttt{light switch}$\rightarrow$\texttt{hinge cabinet}
\item \texttt{kettle}$\rightarrow$\texttt{bottom burner}$\rightarrow$\texttt{light switch}$\rightarrow$\texttt{hinge cabinet}
\item \texttt{microwave}$\rightarrow$\texttt{bottom burner}$\rightarrow$\texttt{hinge cabinet}$\rightarrow$\texttt{top burner}
\item \texttt{kettle}$\rightarrow$\texttt{top burner}$\rightarrow$\texttt{light switch}$\rightarrow$\texttt{slide cabinet}
\item \texttt{microwave}$\rightarrow$\texttt{kettle}$\rightarrow$\texttt{slide cabinet}$\rightarrow$\texttt{bottom burner}
\item \texttt{kettle}$\rightarrow$\texttt{light switch}$\rightarrow$\texttt{slide cabinet}$\rightarrow$\texttt{bottom burner}
\item \texttt{microwave}$\rightarrow$\texttt{kettle}$\rightarrow$\texttt{bottom burner}$\rightarrow$\texttt{top burner}
\item \texttt{microwave}$\rightarrow$\texttt{kettle}$\rightarrow$\texttt{slide cabinet}$\rightarrow$\texttt{hinge cabinet}
\item \texttt{microwave}$\rightarrow$\texttt{bottom burner}$\rightarrow$\texttt{slide cabinet}$\rightarrow$\texttt{top burner}
\item \texttt{kettle}$\rightarrow$\texttt{bottom burner}$\rightarrow$\texttt{light switch}$\rightarrow$\texttt{top burner}
\item \texttt{microwave}$\rightarrow$\texttt{kettle}$\rightarrow$\texttt{top burner}$\rightarrow$\texttt{light switch}
\item \texttt{microwave}$\rightarrow$\texttt{kettle}$\rightarrow$\texttt{light switch}$\rightarrow$\texttt{hinge cabinet}
\item \texttt{microwave}$\rightarrow$\texttt{bottom burner}$\rightarrow$\texttt{light switch}$\rightarrow$\texttt{slide cabinet}
\item \texttt{kettle}$\rightarrow$\texttt{bottom burner}$\rightarrow$\texttt{top burner}$\rightarrow$\texttt{light switch}
\item \texttt{microwave}$\rightarrow$\texttt{light switch}$\rightarrow$\texttt{slide cabinet}$\rightarrow$\texttt{hinge cabinet}
\item \texttt{microwave}$\rightarrow$\texttt{bottom burner}$\rightarrow$\texttt{top burner}$\rightarrow$\texttt{hinge cabinet}
\item \texttt{kettle}$\rightarrow$\texttt{bottom burner}$\rightarrow$\texttt{slide cabinet}$\rightarrow$\texttt{hinge cabinet}
\item \texttt{bottom burner}$\rightarrow$\texttt{top burner}$\rightarrow$\texttt{slide cabinet}$\rightarrow$\texttt{light switch}
\item \texttt{microwave}$\rightarrow$\texttt{kettle}$\rightarrow$\texttt{light switch}$\rightarrow$\texttt{slide cabinet}
\item \texttt{kettle}$\rightarrow$\texttt{bottom burner}$\rightarrow$\texttt{top burner}$\rightarrow$\texttt{hinge cabinet}
\item \texttt{bottom burner}$\rightarrow$\texttt{top burner}$\rightarrow$\texttt{light switch}$\rightarrow$\texttt{slide cabinet}
\end{itemize}

The target tasks are:
\begin{itemize}
\item \texttt{microwave}$\rightarrow$\texttt{bottom burner}$\rightarrow$\texttt{light switch}$\rightarrow$\texttt{top burner}
\item \texttt{microwave}$\rightarrow$\texttt{bottom burner}$\rightarrow$\texttt{top burner}$\rightarrow$\texttt{light switch}
\item \texttt{kettle}$\rightarrow$\texttt{bottom burner}$\rightarrow$\texttt{light switch}$\rightarrow$\texttt{slide cabinet}
\item \texttt{microwave}$\rightarrow$\texttt{kettle}$\rightarrow$\texttt{top burner}$\rightarrow$\texttt{hinge cabinet}
\item \texttt{kettle}$\rightarrow$\texttt{bottom burner}$\rightarrow$\texttt{slide cabinet}$\rightarrow$\texttt{top burner}
\item \texttt{kettle}$\rightarrow$\texttt{light switch}$\rightarrow$\texttt{slide cabinet}$\rightarrow$\texttt{hinge cabinet}
\item \texttt{kettle}$\rightarrow$\texttt{bottom burner}$\rightarrow$\texttt{top burner}$\rightarrow$\texttt{slide cabinet}
\item \texttt{microwave}$\rightarrow$\texttt{bottom burner}$\rightarrow$\texttt{slide cabinet}$\rightarrow$\texttt{hinge cabinet}
\item \texttt{bottom burner}$\rightarrow$\texttt{top burner}$\rightarrow$\texttt{slide cabinet}$\rightarrow$\texttt{hinge cabinet}
\item \texttt{microwave}$\rightarrow$\texttt{kettle}$\rightarrow$\texttt{bottom burner}$\rightarrow$\texttt{hinge cabinet}

\end{itemize}

%% file: sections_metalearn/appendix_task_dist.tex
\section{Meta-training Task Distribution Analysis}
In this section, we aim to investigate the effect of the meta-training task distribution on 
our skill-based meta-training and target task learning phases.
Specifically, we examine the effect of
(1) the number of tasks in the meta-training task distribution
and 
(2) the alignment between a meta-training task distribution and target task distribution.
We conduct experiments and analyses in the maze navigation domain.
More details on task distributions can be found in~\mysecref{sec:app_maze_task_dist}.

\noindent\textbf{Number of meta-training tasks.}
To investigate how the number of meta-training tasks affects the performance of
our method,
we train our method
with fewer numbers meta-training tasks (\ie 10 and 20)
and evaluate it with the same set of target tasks.
The quantitative results presented in~\myfig{fig:maze_density} suggest that 
even with sparser meta-training task distributions 
(\ie fewer numbers of meta-training tasks),
\method\\ is still more sample efficient compared to 
the best-performing baseline (\ie SPiRL).

\input{sections_metalearn/maze_task_dist}

\noindent\textbf{Meta-train / test task alignment.}
We aim to examine if a model trained on a meta-training task distribution
that aligns better/worse with the target tasks
would yield improved/deteriorated performance.
To this end, 
we create biased meta-training / test task distributions: we create a meta-train set by sampling goal locations
from only the top 25\% portion of the maze ($\mathcal{T}_{\textsc{Train-Top}}$).
To rule out the effect of the density of the task distribution,
we sample 10 (\ie 40 $\times$ 25\%) meta-training tasks. Then, we create two target task distributions that have good and bad alignment with this meta-training distribution respectively by sampling \SI{10}{} target tasks from the top 25\% portion of the maze
($\mathcal{T}_{\textsc{Target-Top}}$) and \SI{10}{} target tasks from the bottom 25\% portion of the maze ($\mathcal{T}_{\textsc{Target-Bottom}}$).

\myfig{fig:maze_top} and \myfig{fig:maze_bottom} 
present the target task learning efficiency for models trained 
with good task alignment 
(meta-train on $\mathcal{T}_{\textsc{Train-Top}}$, 
learn target tasks from $\mathcal{T}_{\textsc{Target-Top}}$) 
and bad task alignment 
(meta-train on $\mathcal{T}_{\textsc{Train-Top}}$, 
learn target tasks from $\mathcal{T}_{\textsc{Target-Bottom}}$), 
respectively. 
The results demonstrate that \method\\ can achieve improved performance when trained on a better aligned meta-training task distribution.
On the other hand, not surprisingly, \method\\ and MTRL perform slightly worse compared to SPiRL 
when trained with misaligned meta-training tasks (see~\myfig{fig:maze_bottom}). 
This is expected given that SPiRL does not learn from the misaligned meta-training tasks. 
In summary, 
from~\myfig{fig:task_dist_analysis}, 
we can conclude that
meta-learning from either a diverse task distribution
or a better informed task distribution
can yield improved performance for our method.

%% file: sections_metalearn/maze_task_dist.tex
\begin{figure}[b]
\centering
\subfigure[Sparser Task Distribution]
    {\label{fig:maze_density}\includegraphics[width=0.32\textwidth]{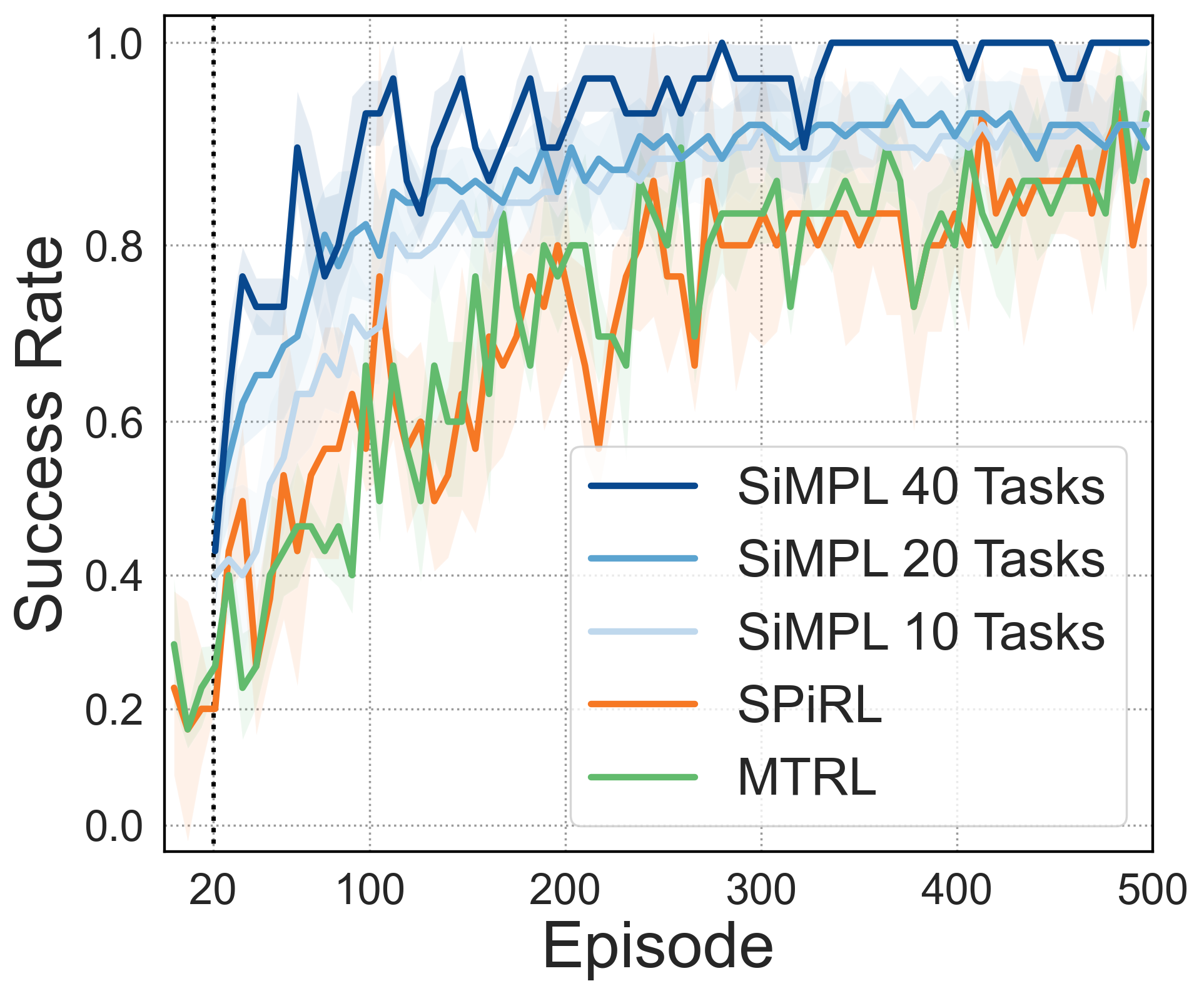}}
\subfigure[$\mathcal{T}_{\textsc{Train-Top}}\rightarrow \mathcal{T}_{\textsc{Target-Top}}$]
    {\label{fig:maze_top}\includegraphics[width=0.32\textwidth]{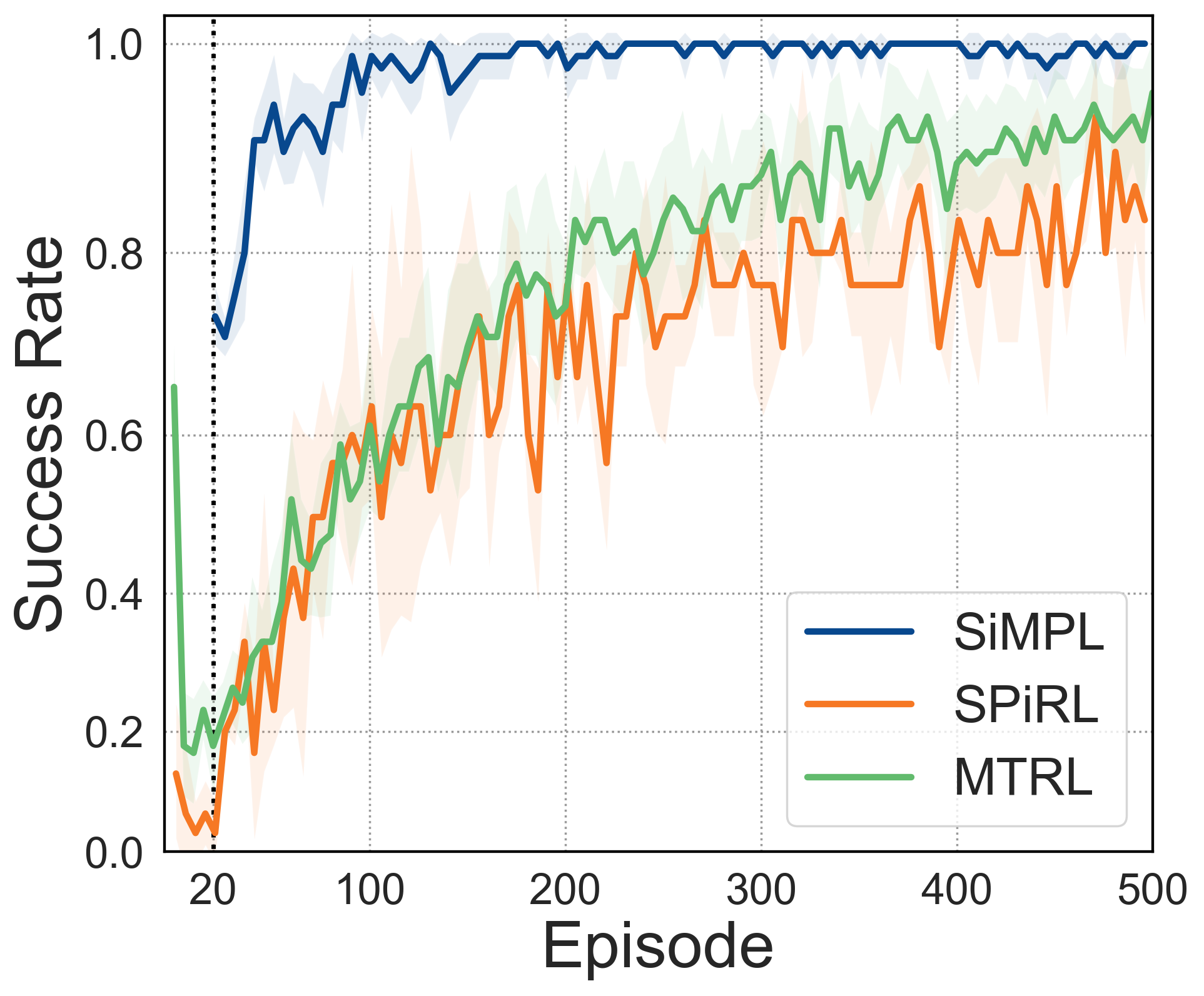}}
\subfigure[$\mathcal{T}_{\textsc{Train-Top}}\rightarrow \mathcal{T}_{\textsc{Target-Bottom}}$]
    {\label{fig:maze_bottom}\includegraphics[width=0.32\textwidth]{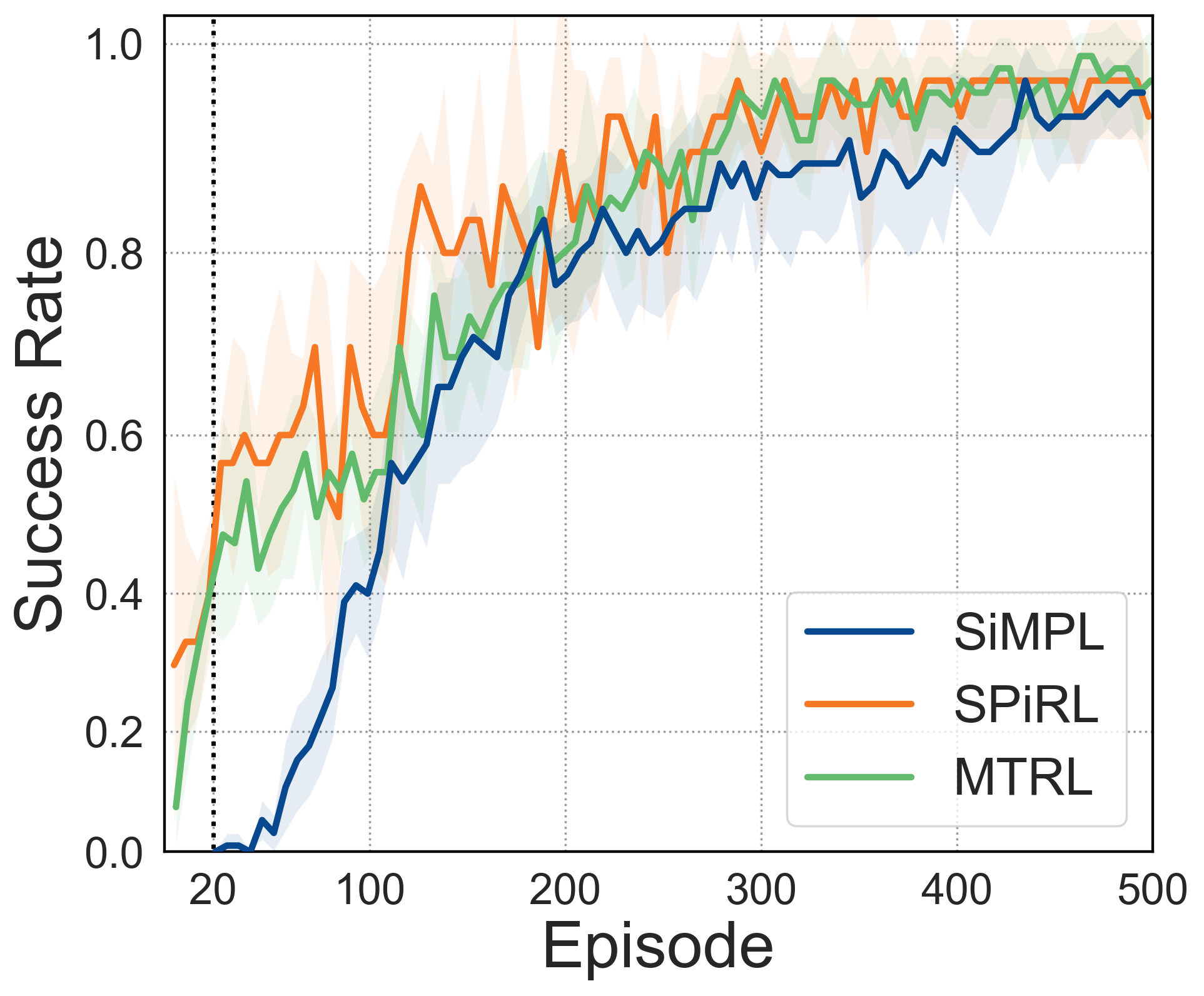}}
    \vspace{-0.3cm} 
\caption{\small
\textbf{Meta-training Task Distribution Analysis.}
(a) With sparser meta-training task distributions 
(\ie fewer numbers of meta-training tasks),
\method\\ still achieves better sample efficiency compared to SPiRL,
highlighting the benefit of leveraging meta-training tasks.
(b) When trained on a meta-training task distribution 
that aligns better with the target task distribution,
\method\\ achieves improved performance.
(c) When trained on a meta-training task distribution 
that is mis-aligned with the target tasks,
\method\\ yields worse performance.
For all the analyses,
we train each model on each target task with 3 different random seeds.
\label{fig:task_dist_analysis}
}
\end{figure}

%% file: sections/appendix_mrl_ablation.tex
\section{Meta-reinforcement Learning Method Ablation}
\label{sec:task_length_ablation}

In this section, we compare the learning efficiency of different meta-RL algorithms with respect to the length of the training tasks. 
Specifically, we hypothesize that our approach \method\\, which extracts temporally extended skills from offline experience, is better suited for learning long-horizon tasks than prior meta-RL algorithms. 
To cleanly investigate the importance of the temporally extended skills vs. the importance of using prior experience we include two additional comparisons to methods 
that leverage prior experience for meta-RL but via flat behavioral cloning instead of through temporally extended skills:

\begin{itemize}
    \item \noindent\textbf{BC+PEARL} first learns a behavior cloning (BC) policy through supervised learning from the offline dataset. 
Then, analogous to our approach \method\\, 
during the meta-training phase,
a task encoder and a meta-learned policy are meta-trained 
with the BC policy constrained SAC objective. 
For fair comparison, 
we use the same residual policy parameterization as described in Section~\ref{sec:policy_arch}.

\item \noindent\textbf{BC+MAML} follows the 
same learning procedure described above,
but uses MAML~\citep{finn2017model} for meta-training instead of PEARL. 
We follow the original learning objective in \citet{finn2017model} (\ie
using REINFORCE~\citep{williams1992simple} for task adaptation, 
and using TRPO~\citep{schulman2017proximal} for meta-policy optimization).
\end{itemize}

\begin{figure}
\centering
\subfigure[$\mathcal{T}_{\textsc{Train-Easy}}$]
    {\label{fig:mrl_task_dist_short}\includegraphics[width=0.24\textwidth]{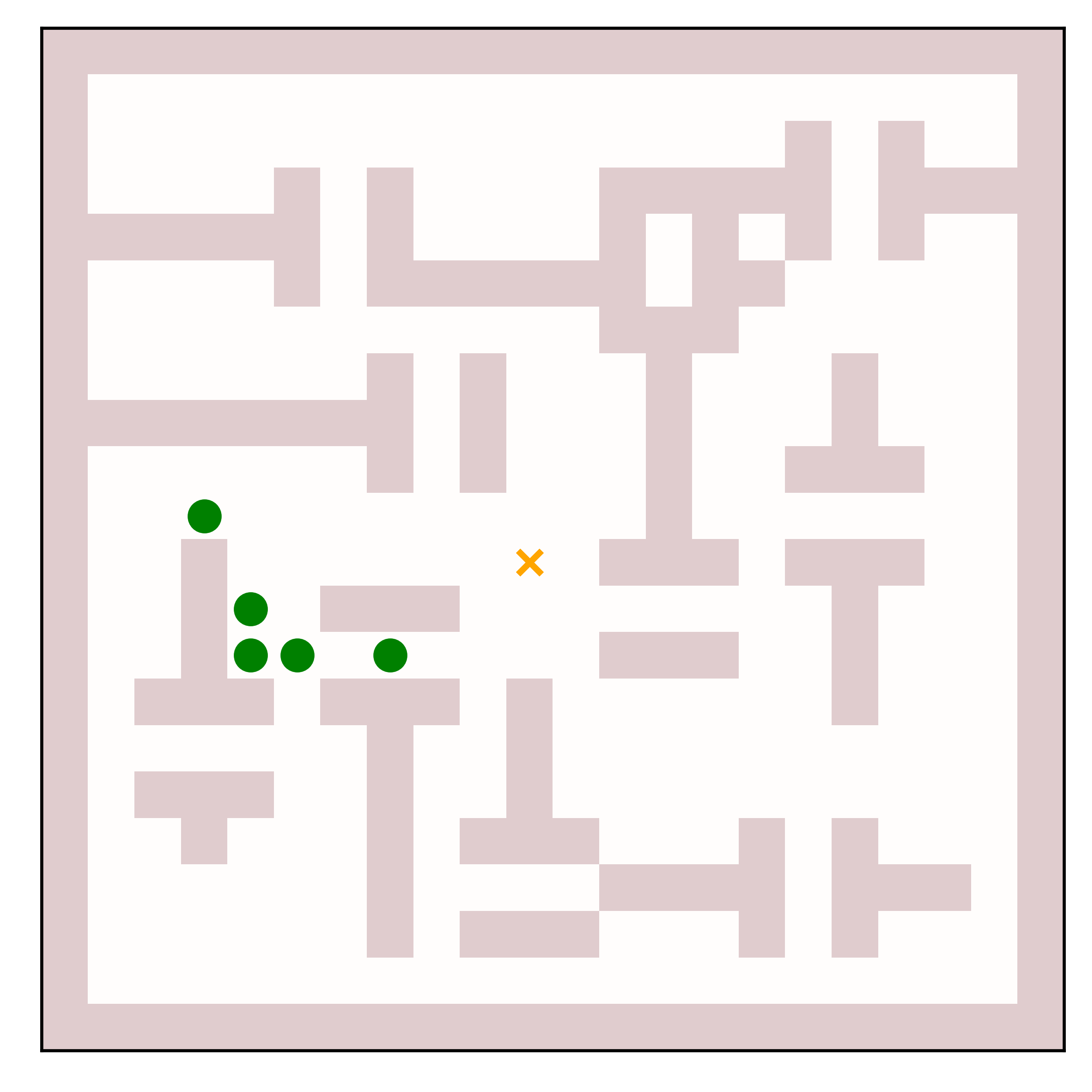}}
\subfigure[$\mathcal{T}_{\textsc{Train-Medium}}$]
    {\includegraphics[width=0.24\textwidth]{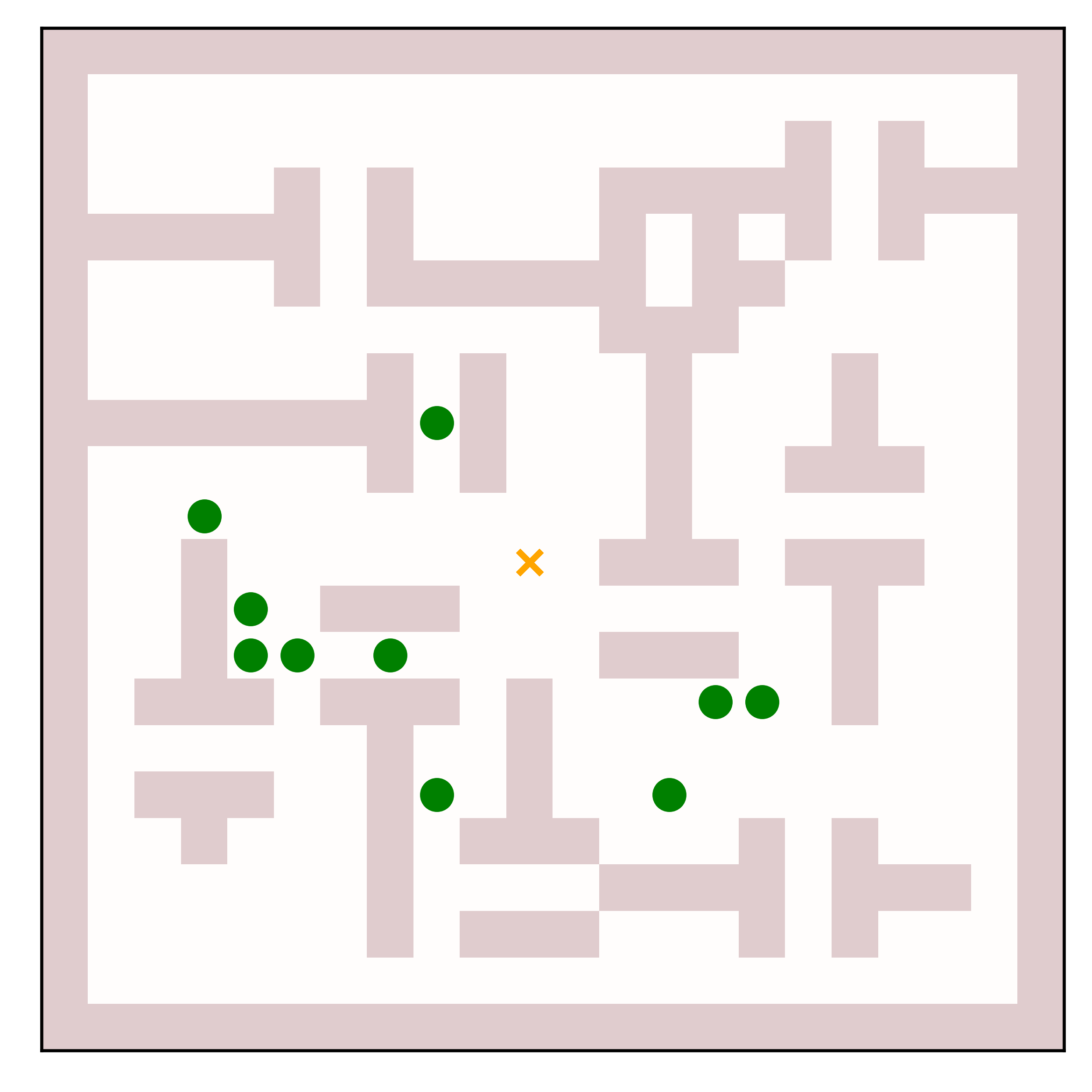}}
\subfigure[$\mathcal{T}_{\textsc{Train-Hard}}$]
    {\includegraphics[width=0.24\textwidth]{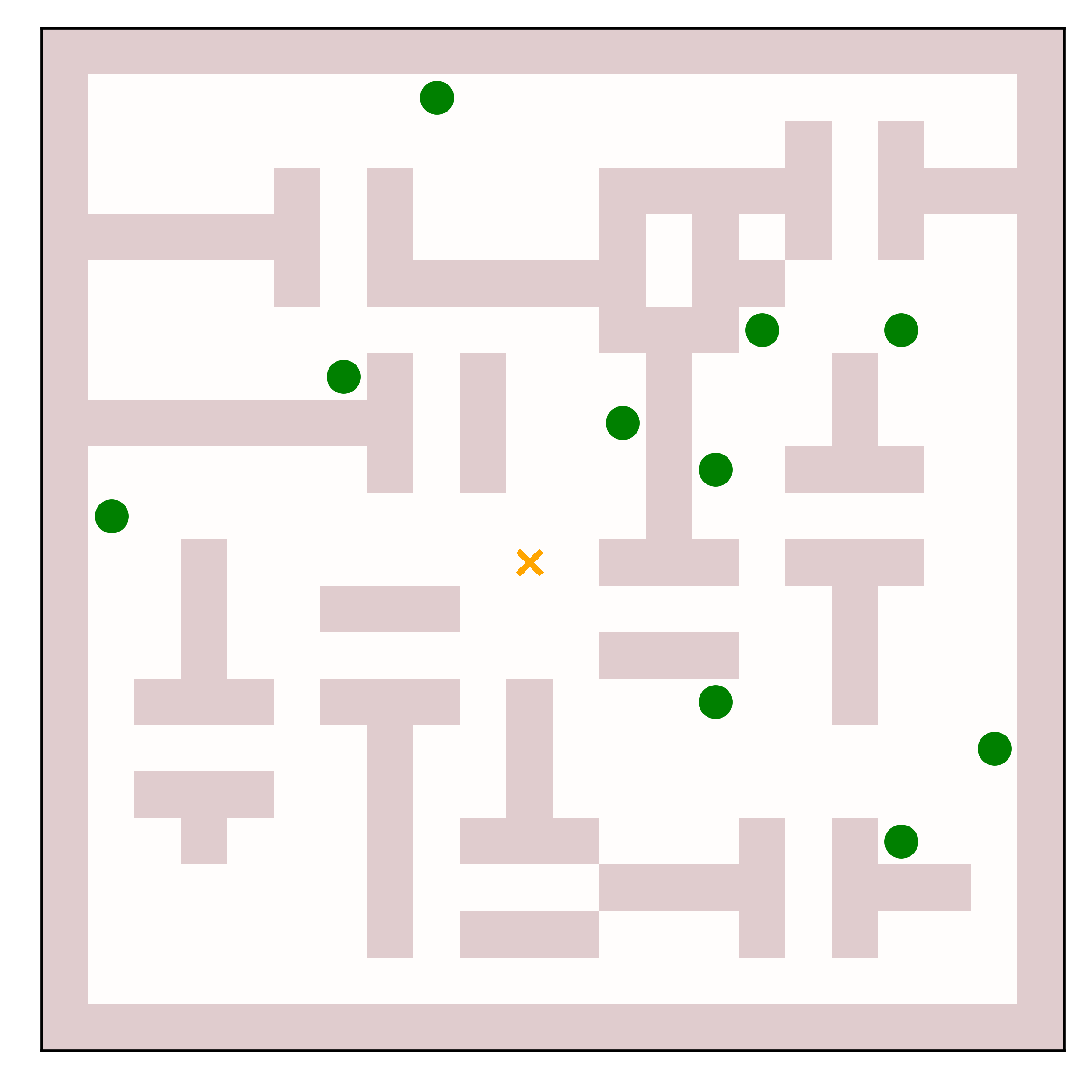}}    
\caption{\small
\textbf{Task Distributions for Task Length Ablation.} We propose three meta-training task distributions of increasing difficulty to compare different meta-RL algorithms: $\mathcal{T}_{\textsc{Train-Easy}}$ uses short-horizon tasks with adjacent goal locations, making exploration easier during meta-training, $\mathcal{T}_{\textsc{Train-Medium}}$ uses similar task horizon but increases the goal position variance, $\mathcal{T}_{\textsc{Train-Hard}}$ contains long-horizon tasks with high variance in goal position and thus is the hardest of the tested task distributions.
\label{fig:mrl_ablation_task_dist}
}
\end{figure}

We compare these methods as well as 
the standard meta-RL approach PEARL~\citep{rakelly2019efficient} 
on three meta-training tasks distributions of increasing complexity in the maze navigation environment (see Figure~\ref{fig:mrl_ablation_task_dist}):
(1) short-range goals with small variance $\mathcal{T}_{\textsc{Train-Easy}}$, 
(2) short-range goals with larger variance $\mathcal{T}_{\textsc{Train-Medium}}$, 
and (3) long-range goals with large variance $\mathcal{T}_{\textsc{Train-Hard}}$, which we used in our original maze experiments.
By increasing variance and length of the tasks in each task distribution, 
we can investigate the learning capability of the meta-RL algorithms.

\begin{figure}
\centering
\subfigure[$\mathcal{T}_{\textsc{Train-Easy}}$]
    {\label{fig:maze_mrl_ablation_short}\includegraphics[width=0.32\textwidth]{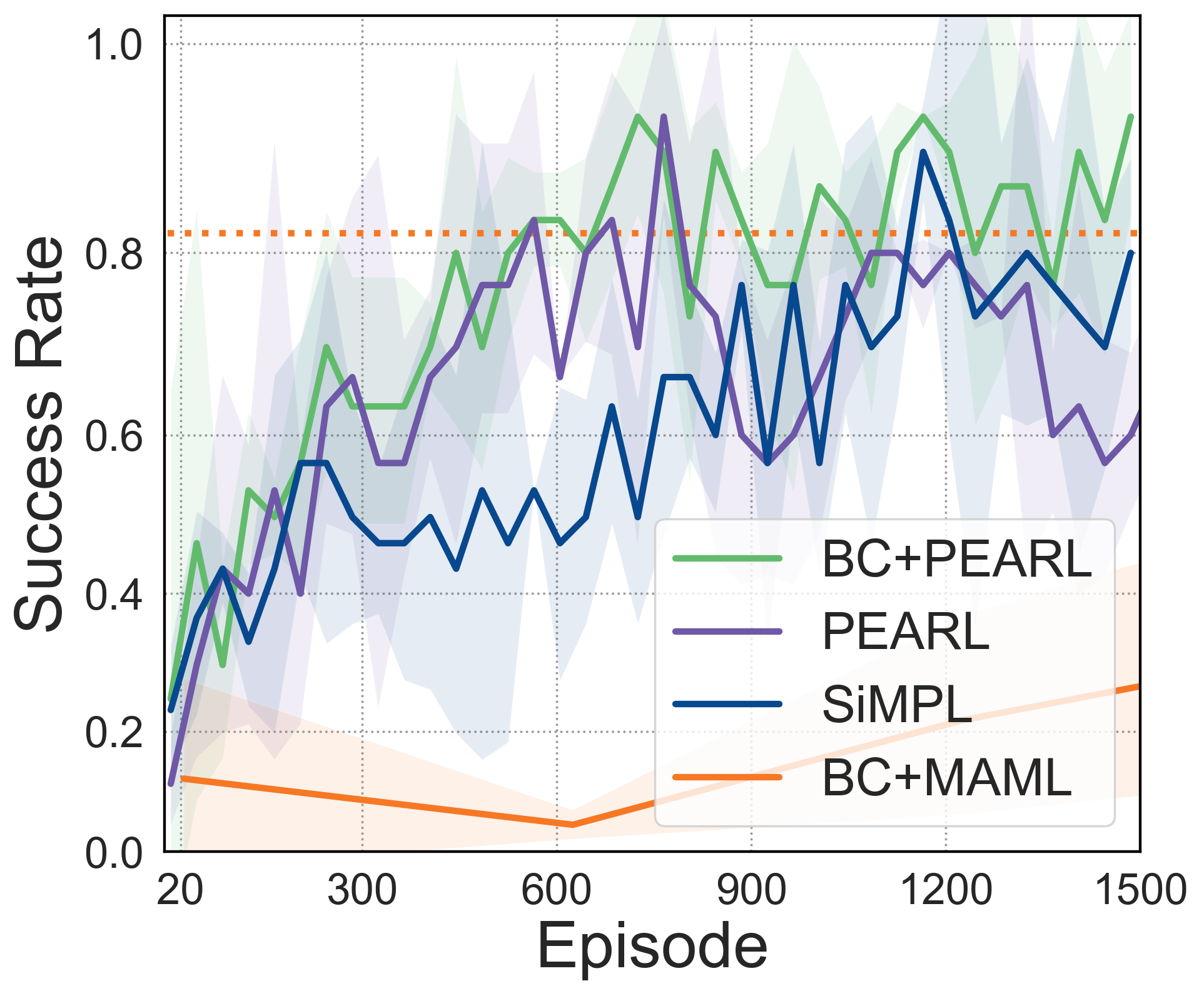}}
\subfigure[$\mathcal{T}_{\textsc{Train-Medium}}$]
    {\label{fig:maze_mrl_ablation_medium}\includegraphics[width=0.32\textwidth]{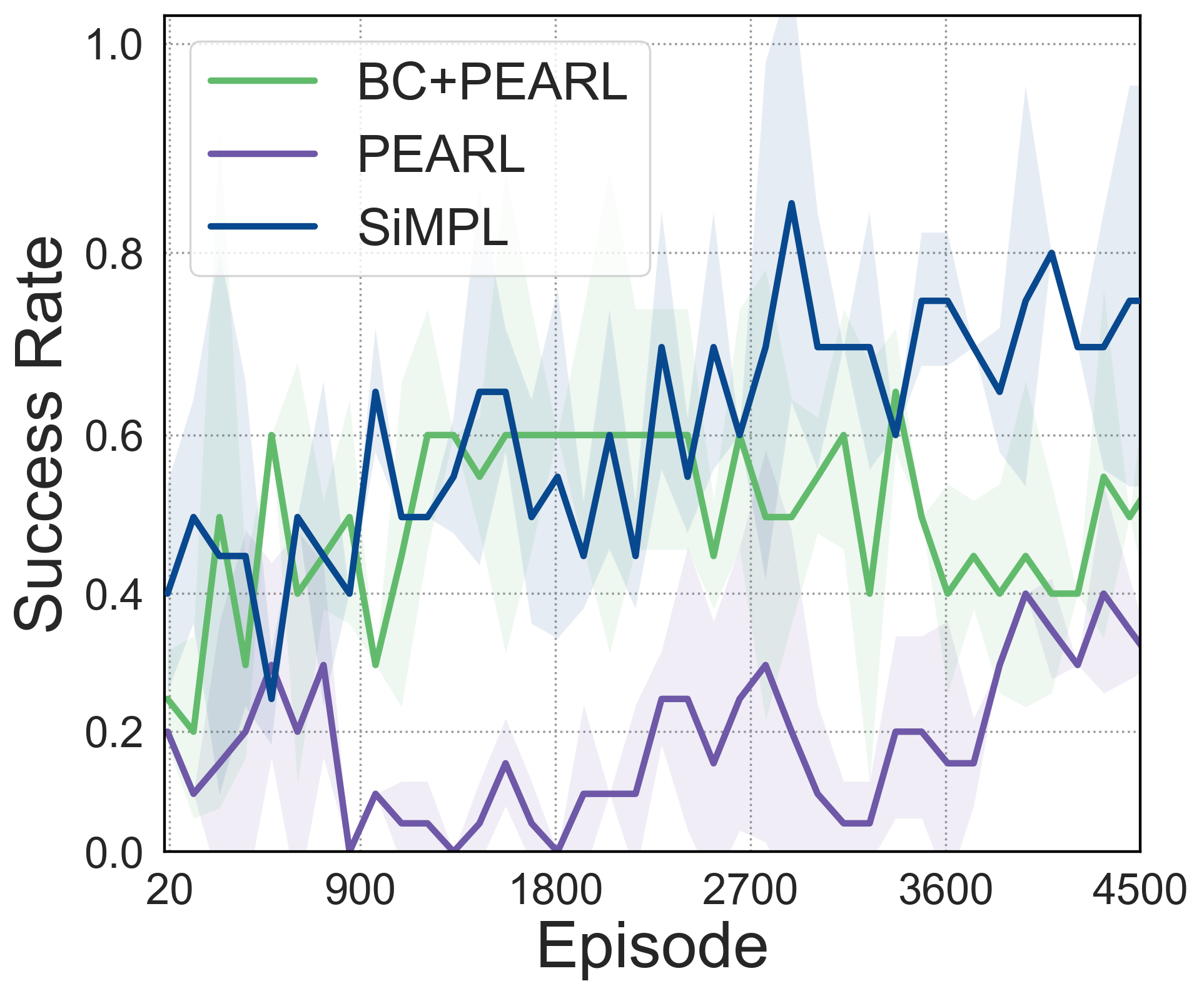}}
\subfigure[$\mathcal{T}_{\textsc{Train-Hard}}$]
    {\label{fig:maze_mrl_ablation_long}\includegraphics[width=0.32\textwidth]{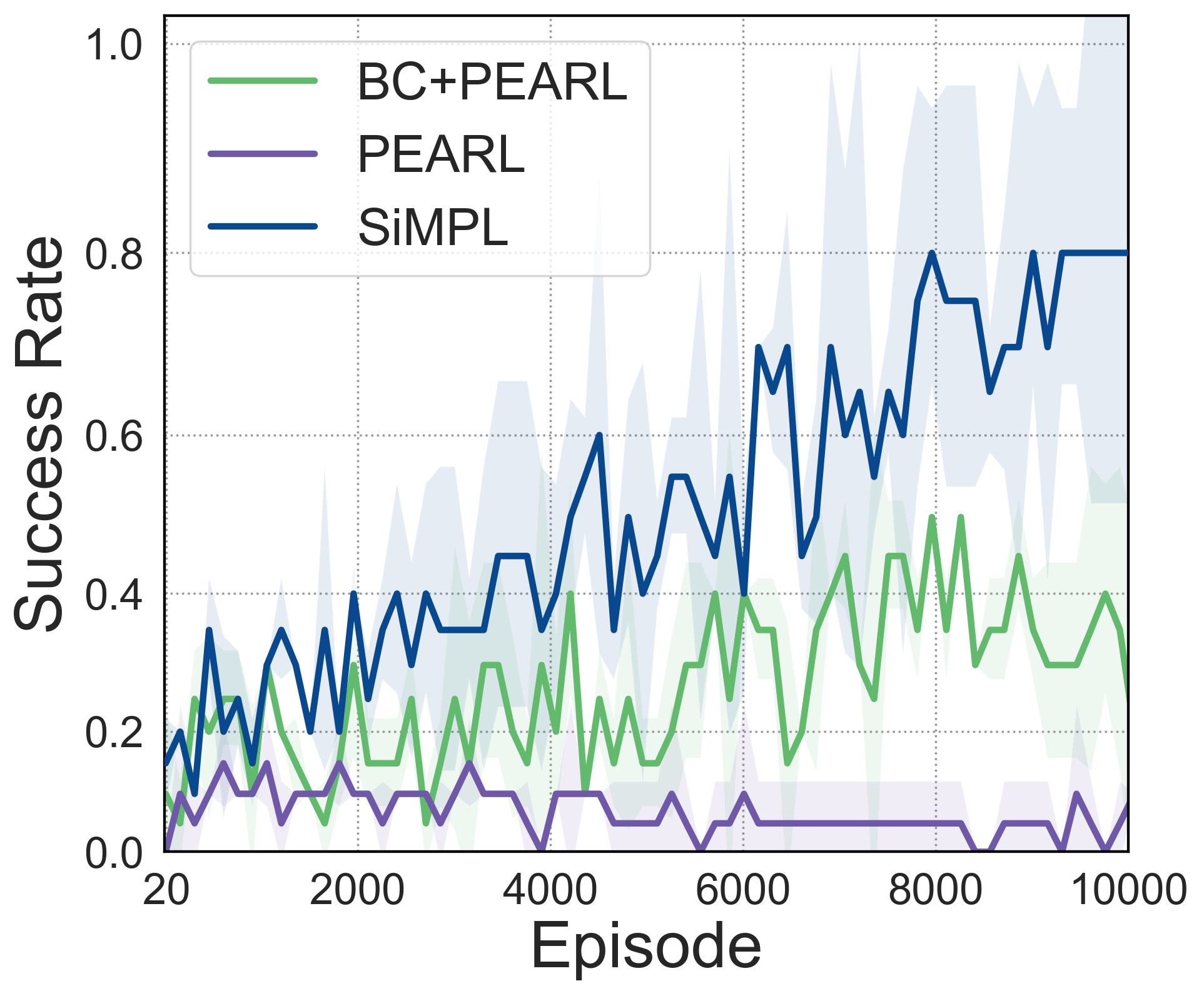}}
    \vspace{-0.3cm} 
\caption{\small
\textbf{Meta-Training Performance for Task Length Ablation.} We find that most meta-learning approaches can solve the simplest task distribution, but using prior experience in BC+PEARL and \method\\ helps for the more challenging distributions (b) and (c). We find that only our approach, which uses the prior data by extracting temporally extended skills, is able to learn the challenging long-horizon tasks efficiently.
\label{fig:mrl_ablation_curve}
}
\end{figure}

\begin{figure}
\centering
\subfigure[PEARL on $\mathcal{T}_{\textsc{Train-Easy}}$]
    {\label{fig:maze_train_short_easy_dist}\includegraphics[width=0.3\textwidth]{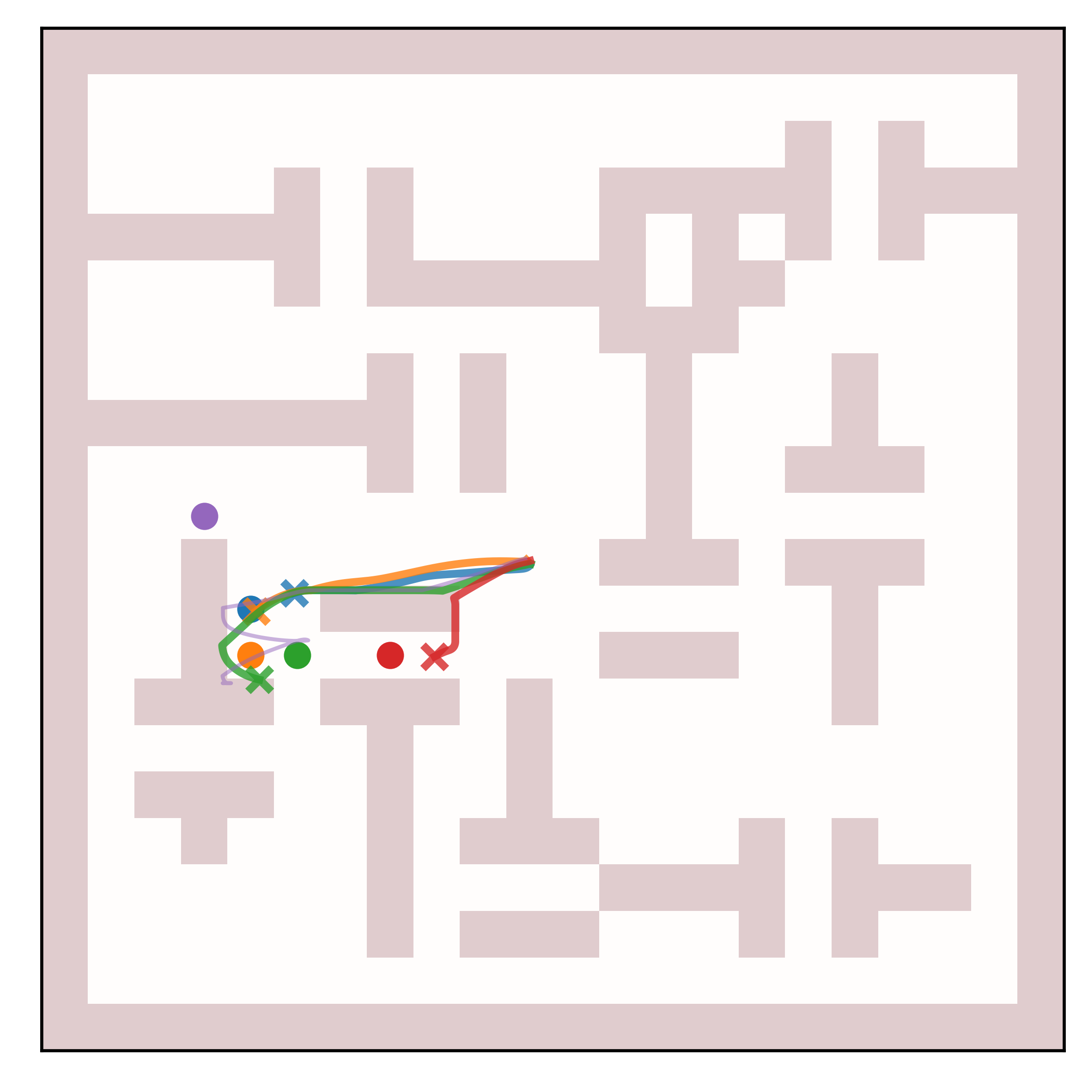}}
\hfill    
\subfigure[BC+PEARL on $\mathcal{T}_{\textsc{Train-Easy}}$]
    {\label{fig:maze_train_short_easy_dist}\includegraphics[width=0.3\textwidth]{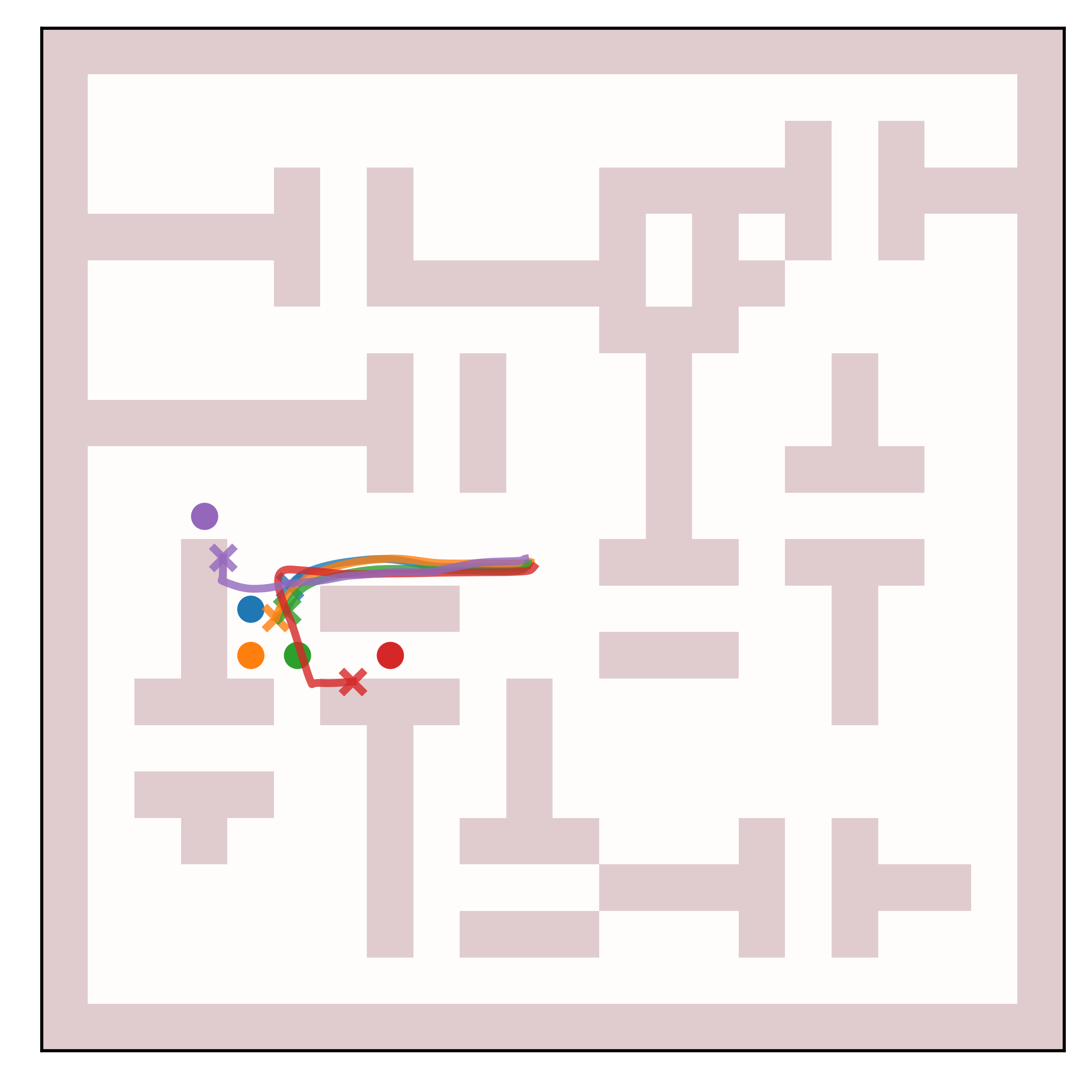}}    
\hfill    
\subfigure[\method\\ on $\mathcal{T}_{\textsc{Train-Easy}}$]
    {\label{fig:maze_train_short_easy_dist}\includegraphics[width=0.3\textwidth]{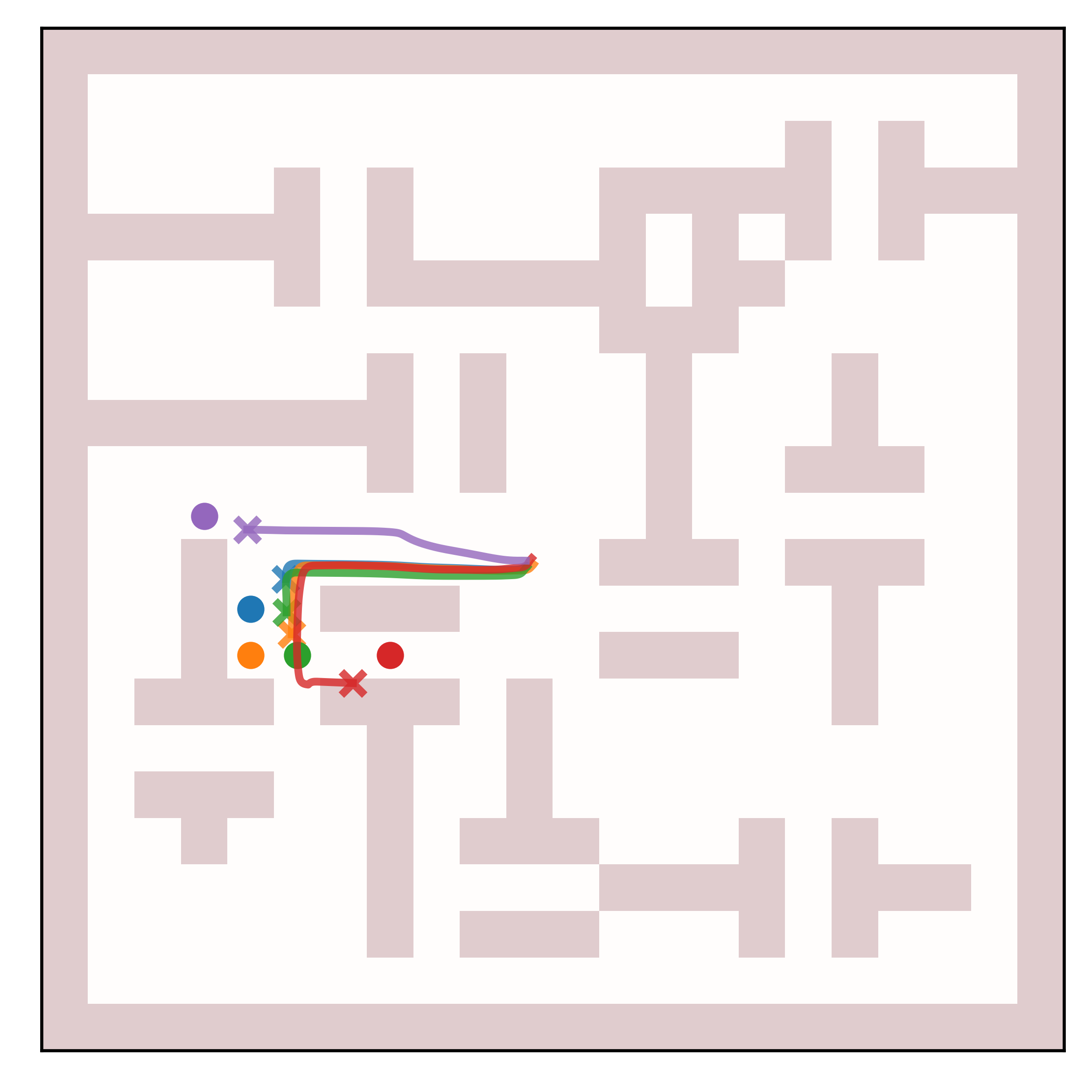}}
\\
\subfigure[PEARL on $\mathcal{T}_{\textsc{Train-Medium}}$]
    {\label{fig:maze_target_short_dist}\includegraphics[width=0.3\textwidth]{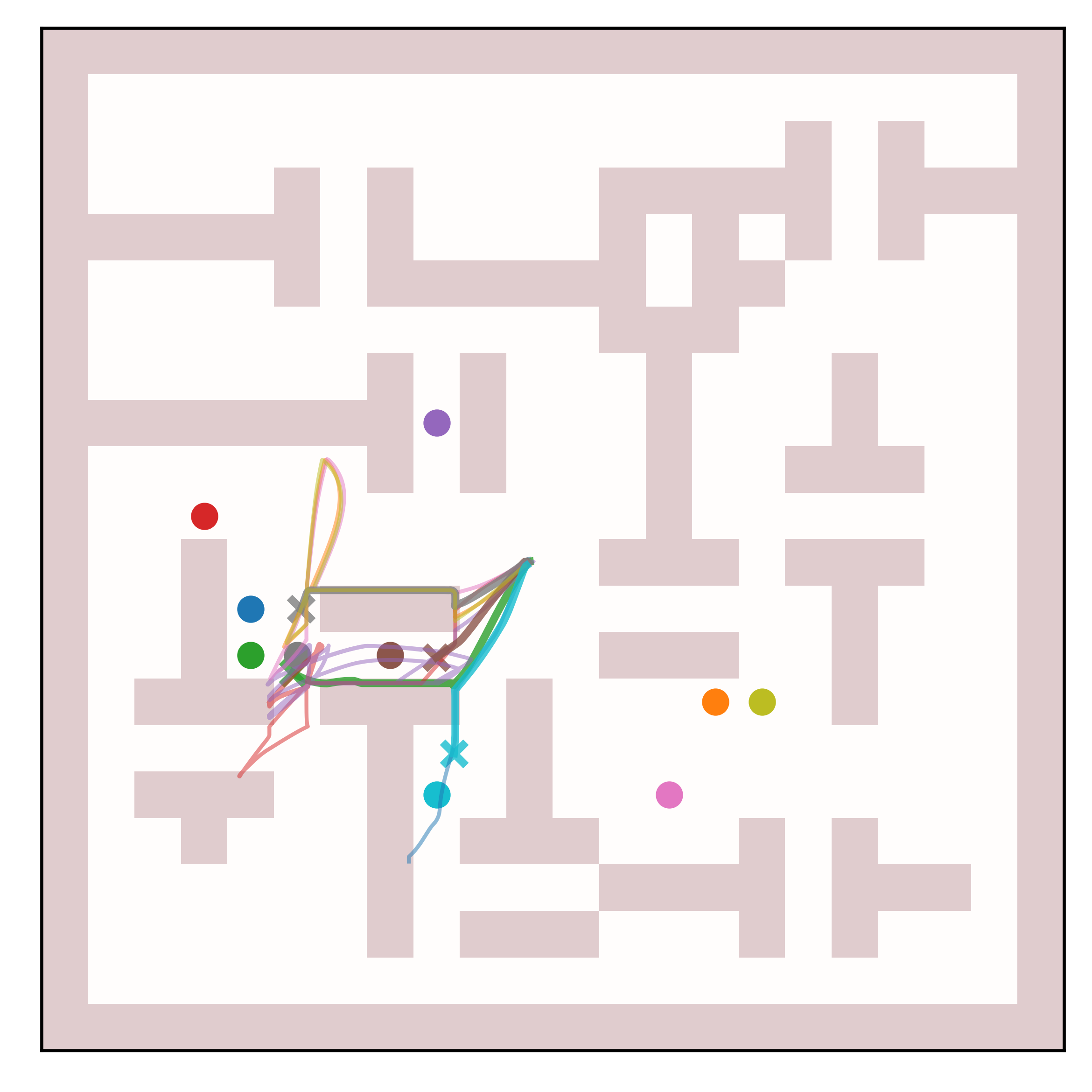}}
\hfill    
\subfigure[BC+PEARL on $\mathcal{T}_{\textsc{Train-Medium}}$]
    {\label{fig:maze_target_short_dist}\includegraphics[width=0.3\textwidth]{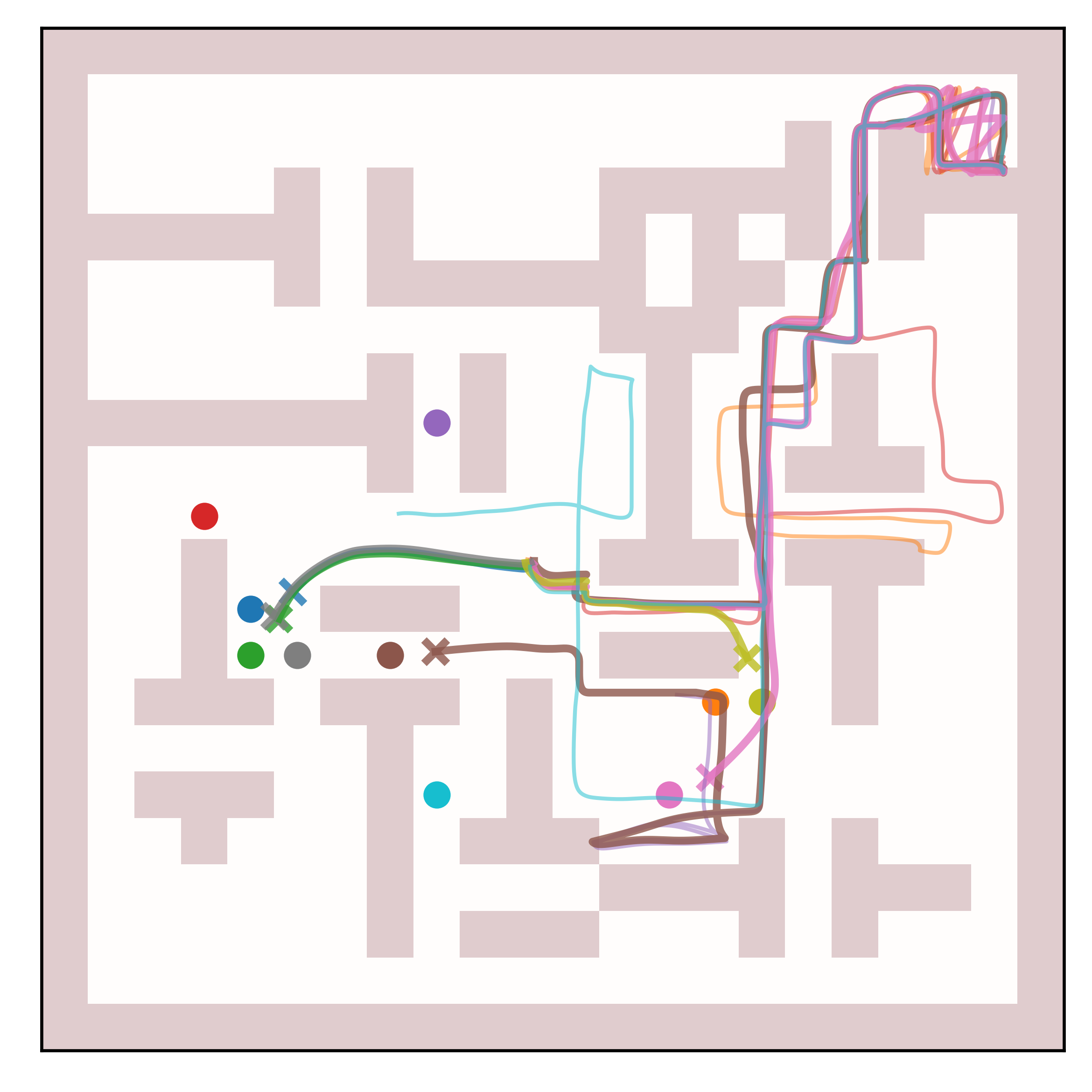}}
\hfill    
\subfigure[\method\\ on $\mathcal{T}_{\textsc{Train-Medium}}$]
    {\label{fig:maze_target_short_dist}\includegraphics[width=0.3\textwidth]{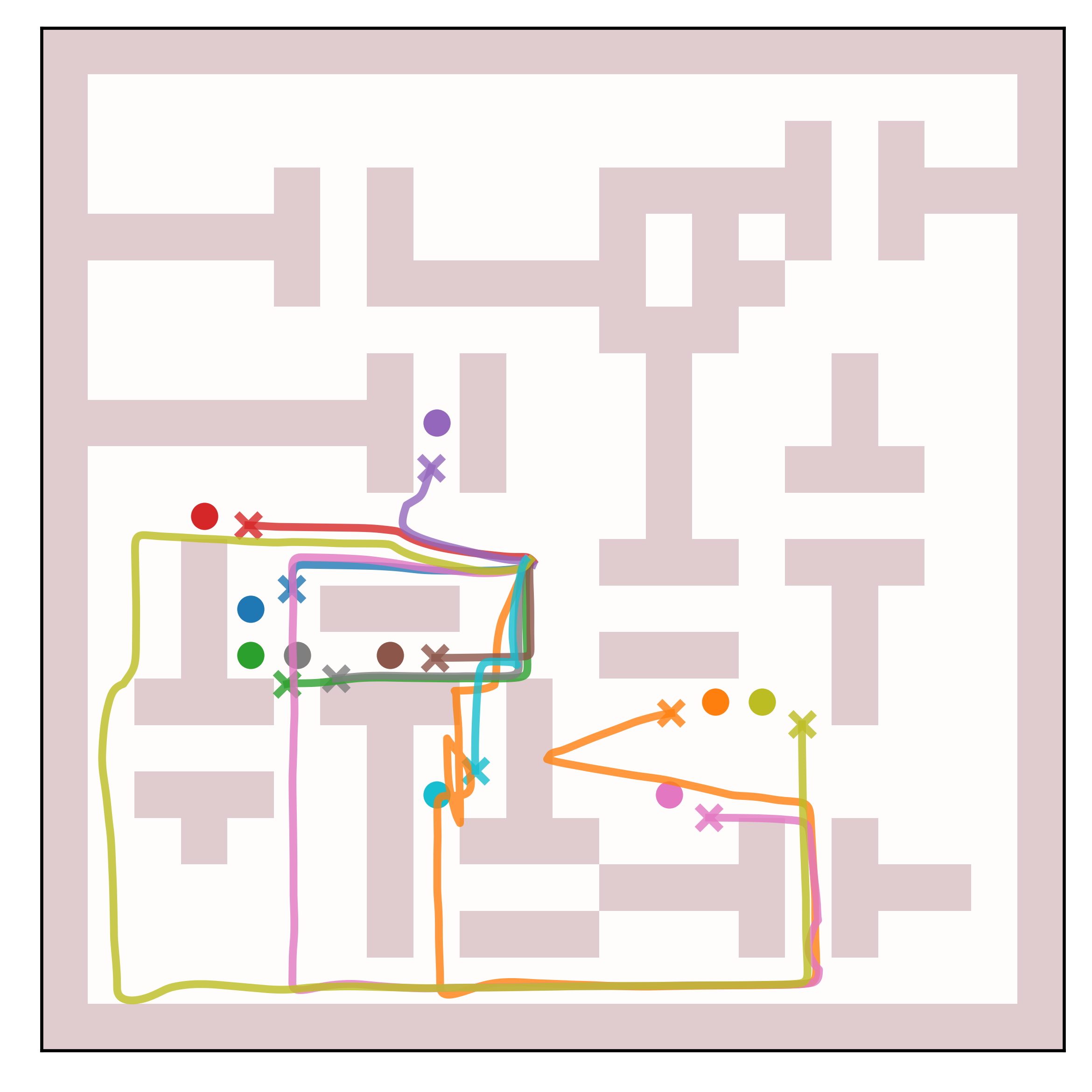}}
\\    
\subfigure[PEARL on $\mathcal{T}_{\textsc{Train-Hard}}$]
    {\label{fig:maze_target_long_dist}\includegraphics[width=0.3\textwidth]{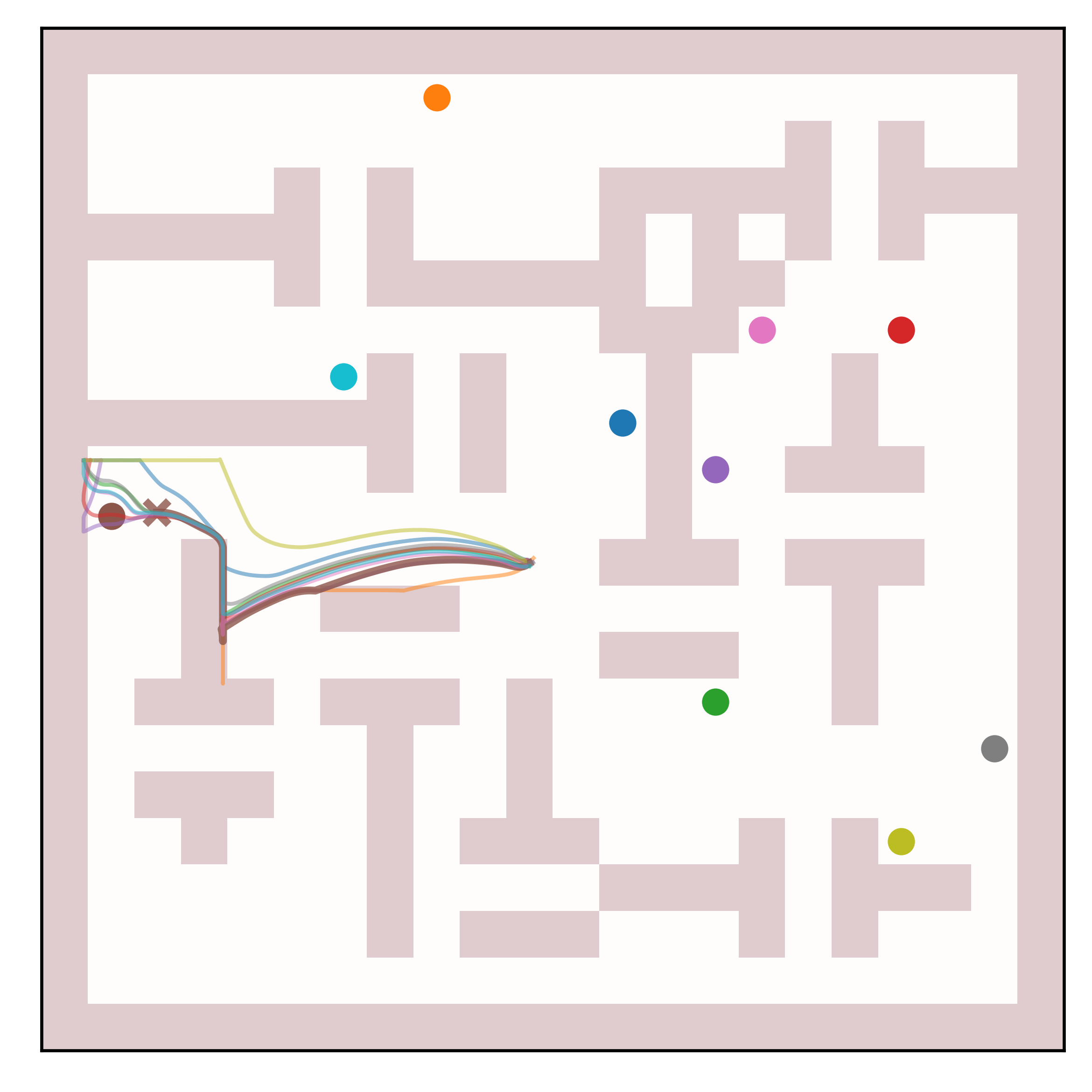}}
\hfill
\subfigure[BC+PEARL on $\mathcal{T}_{\textsc{Train-Hard}}$]
    {\label{fig:maze_target_long_dist}\includegraphics[width=0.3\textwidth]{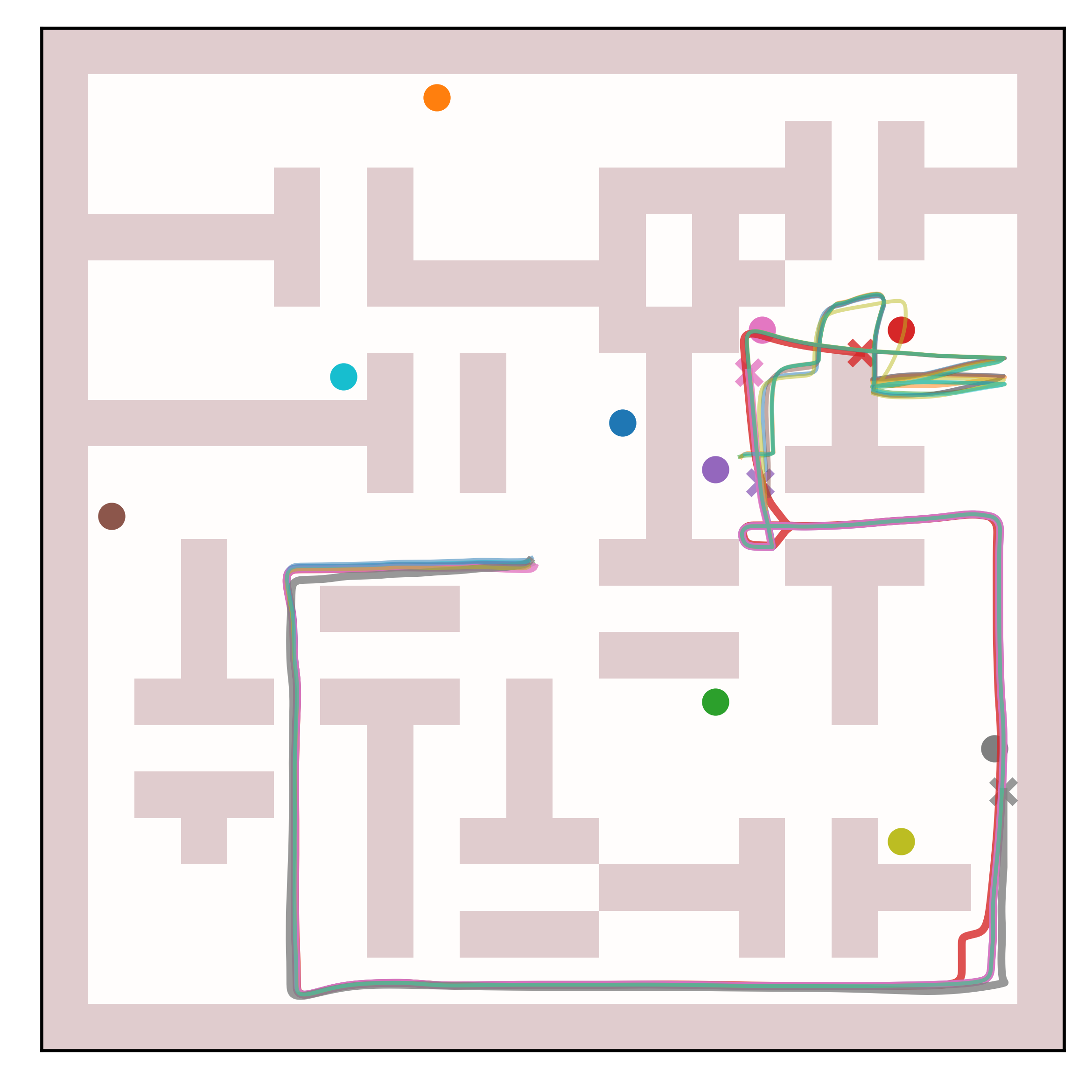}}
\hfill    
\subfigure[\method\\ on $\mathcal{T}_{\textsc{Train-Hard}}$]
    {\label{fig:maze_target_long_dist}\includegraphics[width=0.3\textwidth]{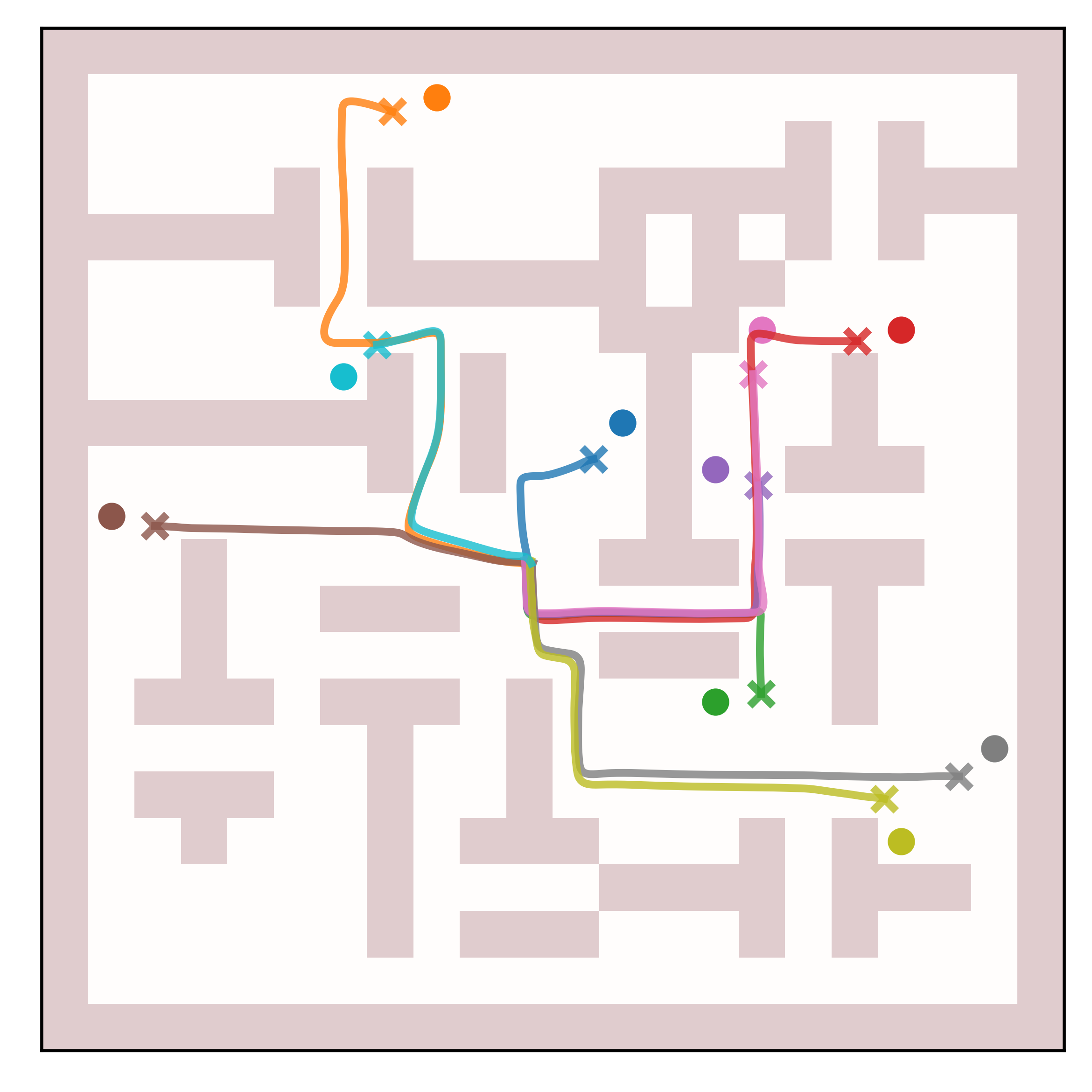}}    
\caption{\small
\textbf{Qualitative Result of Meta-reinforcement Learning Method Ablation.}
\textbf{Top.} All the methods can learn to solve 
short-horizon tasks $\mathcal{T}_{\textsc{Train-Easy}}$.
\textbf{Middle.} On medium-horizon tasks
$\mathcal{T}_{\textsc{Train-Medium}}$,
PEARL struggles at exploring further,
while BC+PEARL exhibits more consistent exploration yet still fails to solve some of the tasks.
\method\\ can explore well and solve all the tasks.
\textbf{Bottom.} On long-horizon tasks$\mathcal{T}_{\textsc{Train-Hard}}$,
PEARL falls into a local minimum,
focusing only on one single task on the left.
BC+PEARL explores slightly better and can solve a few more tasks.
\method\\ can effectively 
learn all the tasks.
\label{fig:maze_task_dist_shorter}
}
\end{figure}

We present the quantitative results in Figure~\ref{fig:mrl_ablation_curve} and the corresponding qualitative analysis in Figure~\ref{fig:maze_task_dist_shorter}. 
On the simplest task distribution we find that all approaches can learn to solve the tasks efficiently, 
except for BC+MAML. 
While the latter also learns to solve the task eventually (see performance upon convergence as dashed orange line in \myfig{fig:maze_mrl_ablation_short}) 
it uses on-policy meta-RL and thus requires substantially more environment interactions during meta-training. 
We thus only consider the more sample efficient BC+PEARL off-policy meta-RL method in the remaining comparisons.

On the more complex task distributions $\mathcal{T}_{\textsc{Train-Medium}}$ and $\mathcal{T}_{\textsc{Train-Hard}}$, 
we find that using prior data for meta-learning is generally beneficial: 
both BC+PEARL and \method\\ learn more efficiently on the task distribution of medium difficulty $\mathcal{T}_{\textsc{Train-Medium}}$, as shown in \myfig{fig:maze_mrl_ablation_medium},
since the policy pre-trained from offline data allows for more efficient exploration during meta-training. 
Importantly, on the hardest task distribution $\mathcal{T}_{\textsc{Train-Hard}}$, 
as shown in \myfig{fig:maze_mrl_ablation_long},
which consists exclusively of long-horizon tasks, 
we find that only \method\\ is able to effectively learn, 
highlighting the importance of leveraging the offline data via temporally extended skills 
instead of flat behavioral cloning. 
This supports our intuition that the abstraction provided by skills is particularly beneficial for meta-learning on long-horizon tasks.

%% file: sections/appendix_few_eps.tex
\section{Learning Efficiency on Target Tasks with Few Episodes of Experience}
\label{sec:data_efficiency}

In this section, 
we examine the data efficiency of the compared methods on the target tasks, 
specifically when provided with only a \textit{few} (<20) episodes of online interaction with an unseen target task. 
Being able to learn new tasks this quickly is a major strength of meta-RL approaches~\citep{finn2017model,rakelly2019efficient}. 
We hypothesize that our skill-based meta-RL algorithm \method\\ can learn similarly fast, 
even on long-horizon, sparse-reward tasks.

\begin{figure}
\centering
\includegraphics[width=1\linewidth]{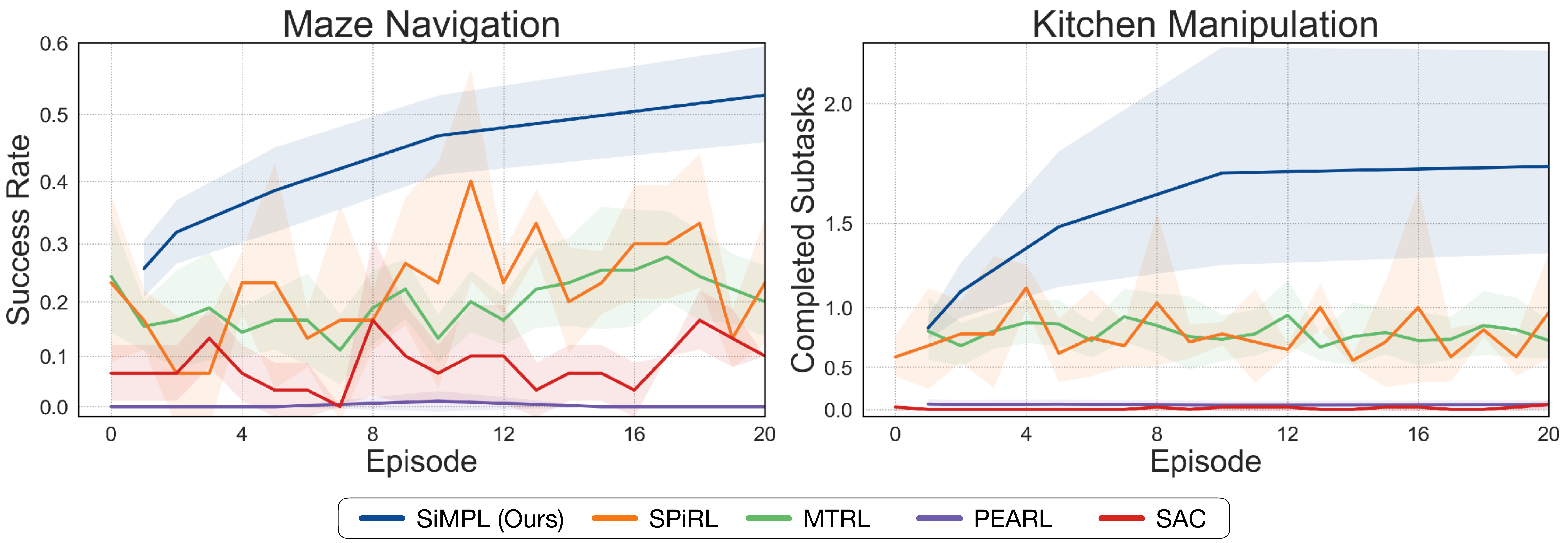} 
\caption{\small
\textbf{Performance with few episodes of target task interaction.} We find that our skill-based meta-RL approach SiMPL is able to learn complex, long-horizon tasks within few episodes of online interaction with a new task while prior meta-RL approaches and non-meta-learning baselines require many more interactions or fail to learn the task altogether.}
\label{fig:few_episode_eval}
\end{figure}

In our original evaluations in Section~\ref{sec:experiments}, 
we used 20 episodes of initial exploration to condition our meta-trained policy.
In Figure~\ref{fig:few_episode_eval}, 
we instead compare performance of different approaches when only provided with very few episodes of online interactions. 
We find that \method\\ learns to solve the unseen tasks substantially faster than all alternative approaches. 
On the kitchen manipulation tasks our approach learns to almost solve two out of four subtasks within a time span equivalent to only a few minutes of real-robot execution time. 
In contrast, prior meta-RL methods struggle at making progress at all on such long-horizon tasks, 
showing the benefit of combining meta-RL with prior offline experience.

%% file: sections/appendix_image_based_maze.tex
\section{Investigating Offline Data vs. Target Domain Shift}
\label{sec:offline_target_domain_shift}
\begin{figure}
\centering
\subfigure[$\mathcal{T}_{\textsc{Train-Image-based}}$]
    {\label{fig:image_based_maze_train_task_dist}\includegraphics[width=0.26\textwidth]{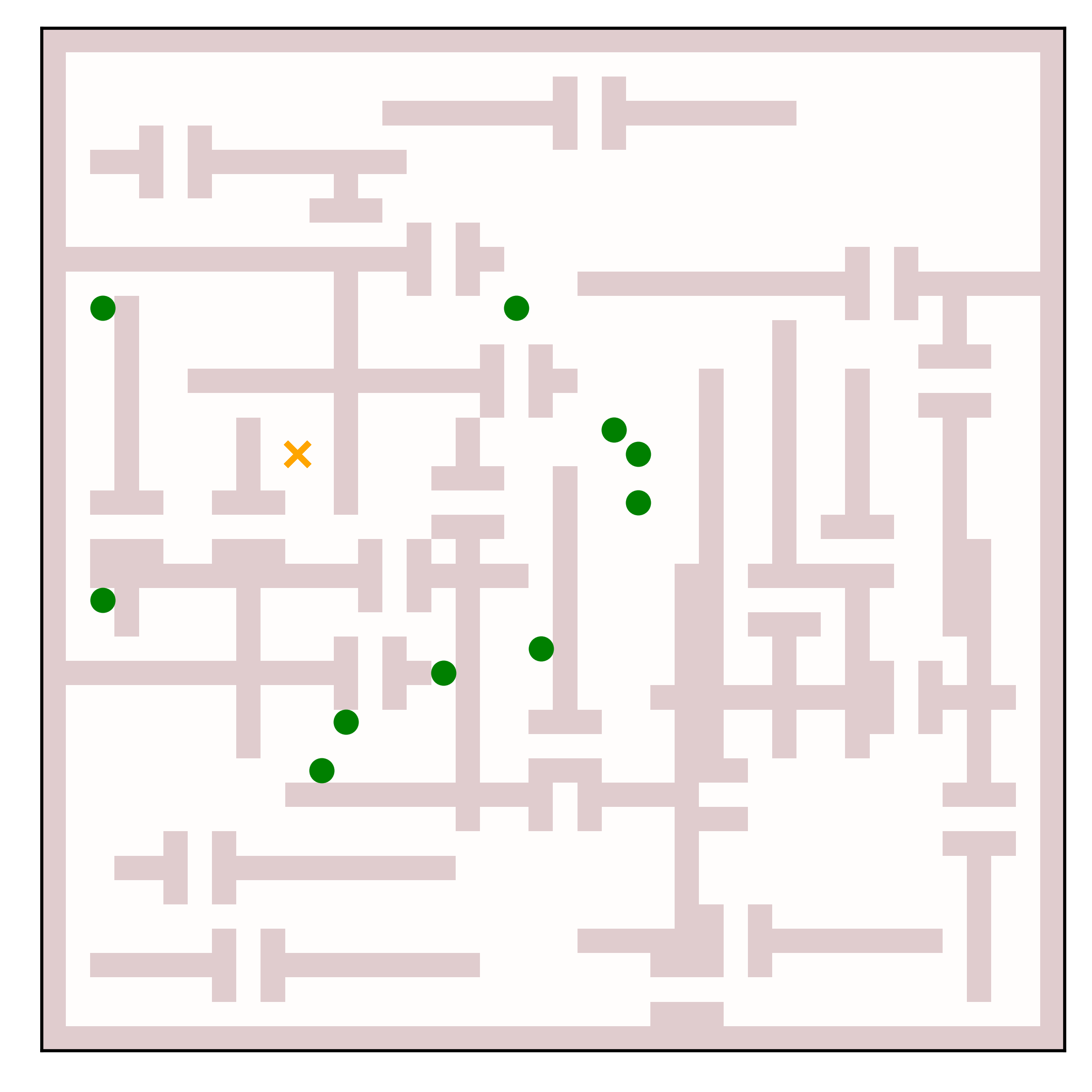}}
\subfigure[$\mathcal{T}_{\textsc{Target-Image-based}}$]
    {\label{fig:image_based_maze_target_task_dist}\includegraphics[width=0.26\textwidth]{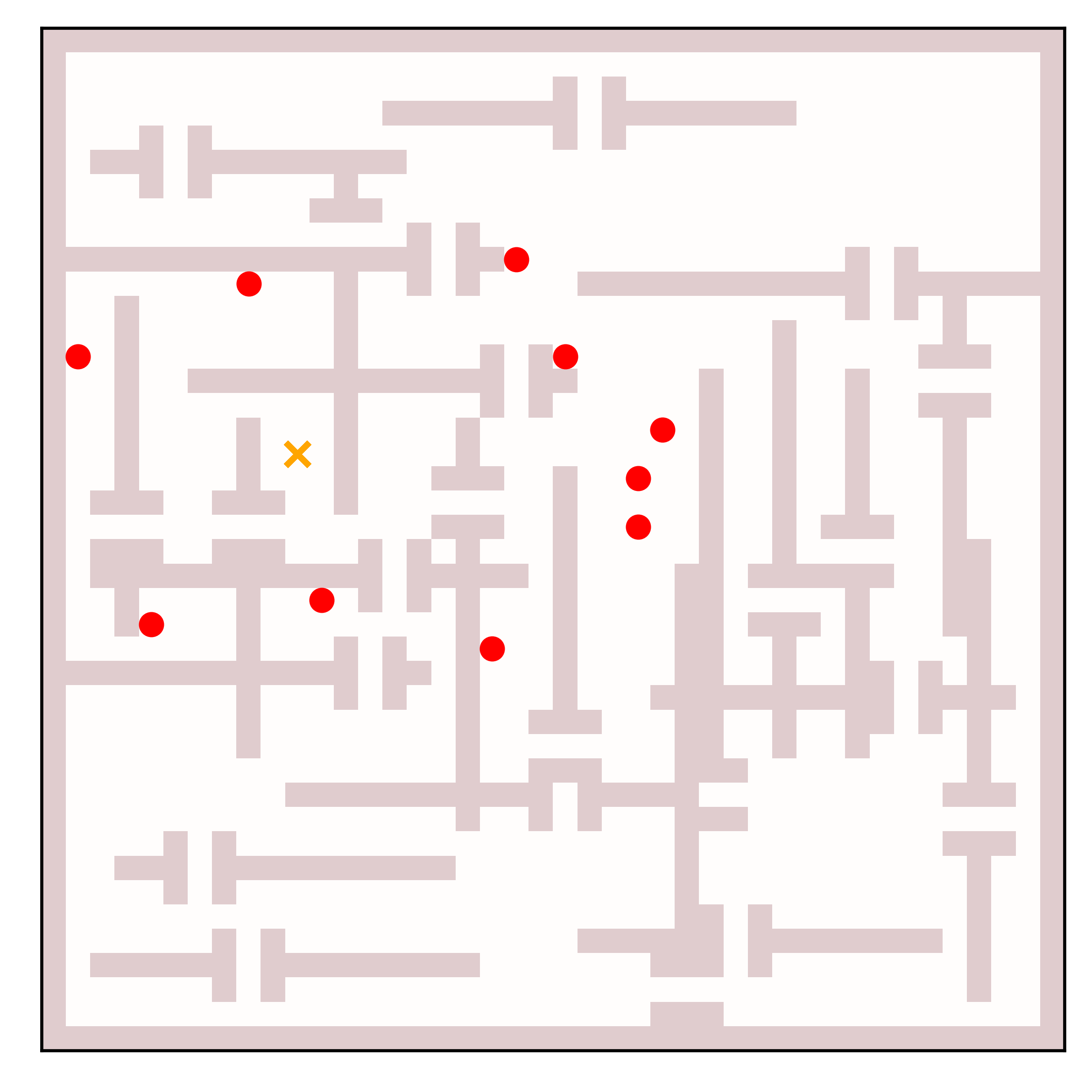}}
\subfigure[Target Task Learning Efficiency]
    {\label{fig:image_based_maze_curve}\includegraphics[width=0.43\textwidth]{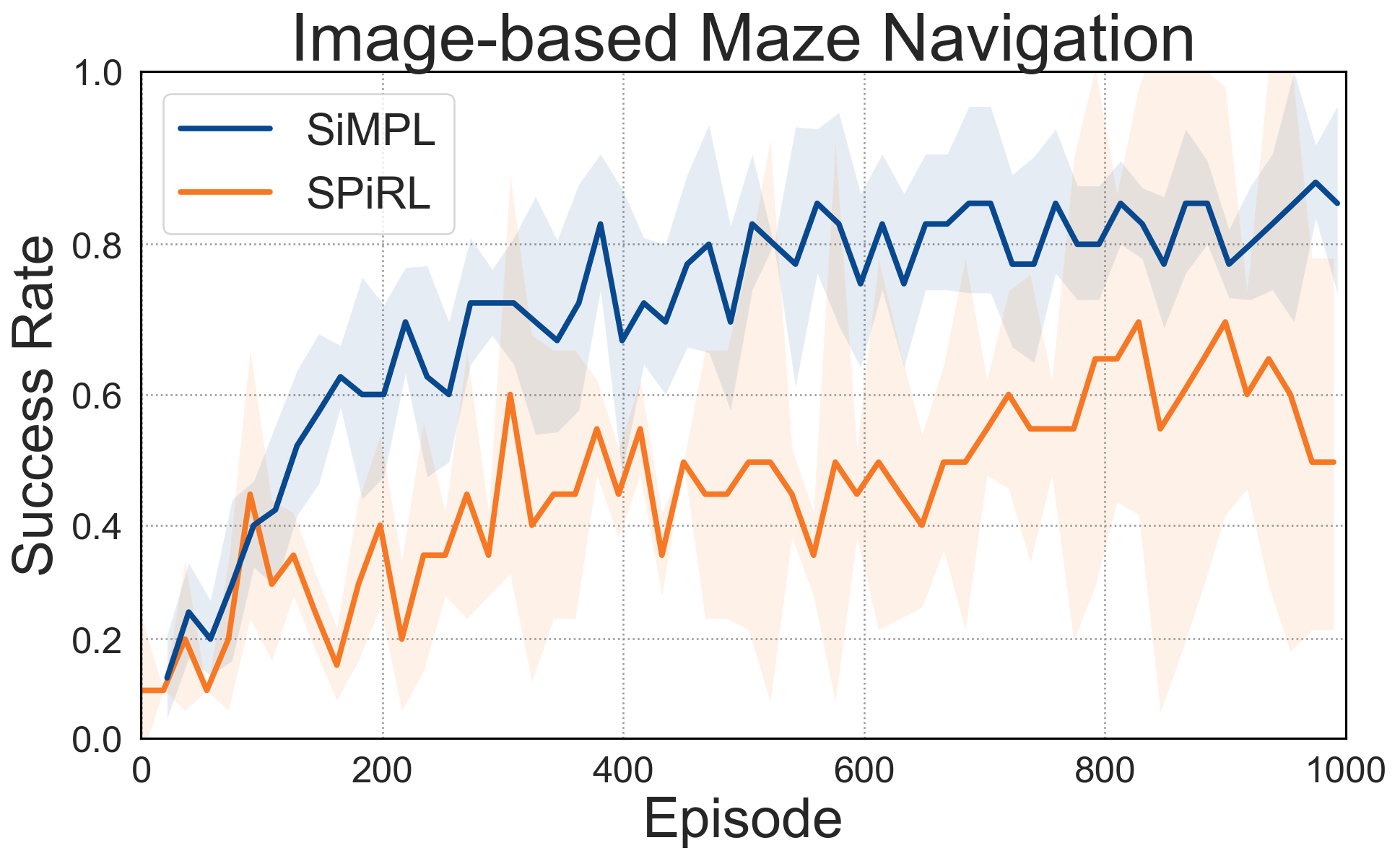}}    
\caption{\small
\textbf{Image-Based Maze Navigation with Distribution Shift.}
\textbf{(a-b)}: Meta-training and target task distributions. The green dots represent the goal locations of meta-training tasks
and the red dots represent the goal locations of target tasks.
The yellow cross represent the initial location of the agent, which is equivalent to the one used in~\citet{pertsch2020spirl}. 
\textbf{(c)}: Performance on the target task. Our approach SiMPL can leverage skills learned from offline data for efficient meta-RL on the maze navigation task and is robust to the domain shift between offline data environments and the target environment.
\label{fig:img_maze_task_dist}
}
\end{figure}

To provide more insights on comparing \method\\ and SPiRL~\citep{pertsch2020spirl},
we evaluate \method\\ in the maze navigation task setup proposed in \citet{pertsch2020spirl}. 
This tests whether our approach can scale to image-based observations: 
\citet{pertsch2020spirl} use $\SI{32}{}\times\SI{32}{}$px observations centered around the agent. 
Even more importantly, it allows us to investigate the robustness of the approach to the domain shifts between the offline pre-training data and the target task: 
we use the maze navigation offline dataset from~\citet{pertsch2020spirl} 
which was collected on \emph{randomly sampled} $\SI{20}{}\times\SI{20}{}$ maze layouts and test on tasks in the unseen, randomly sampled $\SI{40}{}\times\SI{40}{}$ test maze layout from~\citet{pertsch2020spirl}. 
We visualize the meta-training task distribution in~\myfig{fig:image_based_maze_train_task_dist}
and the target task distribution in~\myfig{fig:image_based_maze_target_task_dist}.

We compare the performance of our method to the best-performing baseline, SPiRL~\citep{pertsch2020spirl}, in~\myfig{fig:image_based_maze_curve}. 
Similar to the result presented in \myfig{fig:main_curve}, 
\method\\ can learn the target task faster by combining skills learned from the offline dataset with efficient meta-training. 
This shows that our approach can scale to image-based inputs and 
is robust to substantial domain shifts between the offline pre-training data and the target tasks.

Note that the above results are obtained by comparing our proposed method and SPiRL with the exact same setup used in the SPiRL paper~\citep{pertsch2020spirl}. Specifically, we used the same initial position of the agent as well as sampled the tasks of comparable complexity to the ones used in the SPiRL paper for our evaluation (please see Figure 13 in the SPiRL paper for tasks used in their evaluation). While the used test tasks do not fully cover the entire maze, they are already considerably long-horizon, requiring on average 710 steps until completion while only providing sparse goal-reaching rewards.

To further explore the performance of our proposed method and SPiRL, we have experimented with learning from goals sampled across the entire maze. Yet, SPiRL cannot learn such target tasks and our proposed method consequently does not converge well on the meta-training tasks. This highlights the limitation of skill-based RL methods and can potentially be addressed by learning a more expressive skill prior, \eg using flow models~\citep{dinh2016density}, but this is outside the scope of this work.

%% file: main.bbl
\begin{thebibliography}{56}
\providecommand{\natexlab}[1]{#1}
\providecommand{\url}[1]{\texttt{#1}}
\expandafter\ifx\csname urlstyle\endcsname\relax
  \providecommand{\doi}[1]{doi: #1}\else
  \providecommand{\doi}{doi: \begingroup \urlstyle{rm}\Url}\fi

\bibitem[Ajay et~al.(2021)Ajay, Kumar, Agrawal, Levine, and
  Nachum]{ajay2020opal}
Anurag Ajay, Aviral Kumar, Pulkit Agrawal, Sergey Levine, and Ofir Nachum.
\newblock Opal: Offline primitive discovery for accelerating offline
  reinforcement learning.
\newblock In \emph{International Conference on Learning Representations}, 2021.

\bibitem[Andrychowicz et~al.(2020)Andrychowicz, Baker, Chociej, Jozefowicz,
  McGrew, Pachocki, Petron, Plappert, Powell, Ray, et~al.]{openai2018learning}
OpenAI:~Marcin Andrychowicz, Bowen Baker, Maciek Chociej, Rafal Jozefowicz, Bob
  McGrew, Jakub Pachocki, Arthur Petron, Matthias Plappert, Glenn Powell, Alex
  Ray, et~al.
\newblock Learning dexterous in-hand manipulation.
\newblock \emph{The International Journal of Robotics Research}, 2020.

\bibitem[Bronskill et~al.(2021)Bronskill, Massiceti, Patacchiola, Hofmann,
  Nowozin, and Turner]{bronskill2021memory}
John Bronskill, Daniela Massiceti, Massimiliano Patacchiola, Katja Hofmann,
  Sebastian Nowozin, and Richard Turner.
\newblock Memory efficient meta-learning with large images.
\newblock In \emph{Neural Information Processing Systems}, 2021.

\bibitem[Chebotar et~al.(2021)Chebotar, Hausman, Lu, Xiao, Kalashnikov, Varley,
  Irpan, Eysenbach, Julian, Finn, et~al.]{chebotar2021actionable}
Yevgen Chebotar, Karol Hausman, Yao Lu, Ted Xiao, Dmitry Kalashnikov, Jake
  Varley, Alex Irpan, Benjamin Eysenbach, Ryan Julian, Chelsea Finn, et~al.
\newblock Actionable models: Unsupervised offline reinforcement learning of
  robotic skills.
\newblock \emph{arXiv preprint arXiv:2104.07749}, 2021.

\bibitem[Clavera et~al.(2018)Clavera, Rothfuss, Schulman, Fujita, Asfour, and
  Abbeel]{clavera18a}
Ignasi Clavera, Jonas Rothfuss, John Schulman, Yasuhiro Fujita, Tamim Asfour,
  and Pieter Abbeel.
\newblock Model-based reinforcement learning via meta-policy optimization.
\newblock In \emph{Conference on Robot Learning}, 2018.

\bibitem[Clavera et~al.(2019)Clavera, Nagabandi, Liu, Fearing, Abbeel, Levine,
  and Finn]{clavera2018learning}
Ignasi Clavera, Anusha Nagabandi, Simin Liu, Ronald~S. Fearing, Pieter Abbeel,
  Sergey Levine, and Chelsea Finn.
\newblock Learning to adapt in dynamic, real-world environments through
  meta-reinforcement learning.
\newblock In \emph{International Conference on Learning Representations}, 2019.

\bibitem[Dasari et~al.(2019)Dasari, Ebert, Tian, Nair, Bucher, Schmeckpeper,
  Singh, Levine, and Finn]{dasari2019robonet}
Sudeep Dasari, Frederik Ebert, Stephen Tian, Suraj Nair, Bernadette Bucher,
  Karl Schmeckpeper, Siddharth Singh, Sergey Levine, and Chelsea Finn.
\newblock Robonet: Large-scale multi-robot learning.
\newblock In \emph{Conference on Robot Learning}, 2019.

\bibitem[Dinh et~al.(2017)Dinh, Sohl-Dickstein, and Bengio]{dinh2016density}
Laurent Dinh, Jascha Sohl-Dickstein, and Samy Bengio.
\newblock Density estimation using real nvp.
\newblock In \emph{International Conference on Learning Representations}, 2017.

\bibitem[Dorfman et~al.(2021)Dorfman, Shenfeld, and Tamar]{dorfman2020offline}
Ron Dorfman, Idan Shenfeld, and Aviv Tamar.
\newblock Offline meta learning of exploration.
\newblock In \emph{Neural Information Processing Systems}, 2021.

\bibitem[Duan et~al.(2016)Duan, Schulman, Chen, Bartlett, Sutskever, and
  Abbeel]{duan2016rl}
Yan Duan, John Schulman, Xi~Chen, Peter~L Bartlett, Ilya Sutskever, and Pieter
  Abbeel.
\newblock {RL}$^2$: Fast reinforcement learning via slow reinforcement
  learning.
\newblock \emph{arXiv preprint arXiv:1611.02779}, 2016.

\bibitem[Dvornik et~al.(2020)Dvornik, Schmid, and Mairal]{dvornik2020selecting}
Nikita Dvornik, Cordelia Schmid, and Julien Mairal.
\newblock Selecting relevant features from a multi-domain representation for
  few-shot classification.
\newblock In \emph{European Conference on Computer Vision}, 2020.

\bibitem[Finn et~al.(2017)Finn, Abbeel, and Levine]{finn2017model}
Chelsea Finn, Pieter Abbeel, and Sergey Levine.
\newblock Model-agnostic meta-learning for fast adaptation of deep networks.
\newblock In \emph{International Conference on Machine Learning}, 2017.

\bibitem[Fu et~al.(2020)Fu, Kumar, Nachum, Tucker, and Levine]{fu2020d4rl}
Justin Fu, Aviral Kumar, Ofir Nachum, George Tucker, and Sergey Levine.
\newblock D4rl: Datasets for deep data-driven reinforcement learning.
\newblock \emph{arXiv preprint arXiv:2004.07219}, 2020.

\bibitem[Gu et~al.(2017)Gu, Holly, Lillicrap, and Levine]{gu2017deep}
Shixiang Gu, Ethan Holly, Timothy Lillicrap, and Sergey Levine.
\newblock Deep reinforcement learning for robotic manipulation with
  asynchronous off-policy updates.
\newblock In \emph{IEEE International Conference on Robotics and Automation},
  2017.

\bibitem[Gulcehre et~al.(2020)Gulcehre, Wang, Novikov, Paine, Colmenarejo,
  Zolna, Agarwal, Merel, Mankowitz, Paduraru, et~al.]{gulcehre2020rl}
Caglar Gulcehre, Ziyu Wang, Alexander Novikov, Tom~Le Paine, Sergio~G{\'o}mez
  Colmenarejo, Konrad Zolna, Rishabh Agarwal, Josh Merel, Daniel Mankowitz,
  Cosmin Paduraru, et~al.
\newblock Rl unplugged: Benchmarks for offline reinforcement learning.
\newblock \emph{arXiv preprint arXiv:2006.13888}, 2020.

\bibitem[Gupta et~al.(2018)Gupta, Mendonca, Liu, Abbeel, and
  Levine]{NEURIPS2018_4de75424}
Abhishek Gupta, Russell Mendonca, YuXuan Liu, Pieter Abbeel, and Sergey Levine.
\newblock Meta-reinforcement learning of structured exploration strategies.
\newblock In \emph{Neural Information Processing Systems}, 2018.

\bibitem[Gupta et~al.(2019)Gupta, Kumar, Lynch, Levine, and
  Hausman]{gupta2019relay}
Abhishek Gupta, Vikash Kumar, Corey Lynch, Sergey Levine, and Karol Hausman.
\newblock Relay policy learning: Solving long-horizon tasks via imitation and
  reinforcement learning.
\newblock In \emph{Conference on Robot Learning}, 2019.

\bibitem[Haarnoja et~al.(2018)Haarnoja, Zhou, Abbeel, and
  Levine]{haarnoja2018sac}
Tuomas Haarnoja, Aurick Zhou, Pieter Abbeel, and Sergey Levine.
\newblock Soft actor-critic: Off-policy maximum entropy deep reinforcement
  learning with a stochastic actor.
\newblock In \emph{International Conference on Machine Learning}, 2018.

\bibitem[Hausman et~al.(2018)Hausman, Springenberg, Wang, Heess, and
  Riedmiller]{hausman2018learning}
Karol Hausman, Jost~Tobias Springenberg, Ziyu Wang, Nicolas Heess, and Martin
  Riedmiller.
\newblock Learning an embedding space for transferable robot skills.
\newblock In \emph{International Conference on Learning Representations}, 2018.

\bibitem[Higgins et~al.(2017)Higgins, Matthey, Pal, Burgess, Glorot, Botvinick,
  Mohamed, and Lerchner]{higgins2017beta}
Irina Higgins, Loic Matthey, Arka Pal, Christopher Burgess, Xavier Glorot,
  Matthew Botvinick, Shakir Mohamed, and Alexander Lerchner.
\newblock beta-{VAE}: Learning basic visual concepts with a constrained
  variational framework.
\newblock In \emph{International Conference on Learning Representations}, 2017.

\bibitem[Humplik et~al.(2019)Humplik, Galashov, Hasenclever, Ortega, Teh, and
  Heess]{humplik2019meta}
Jan Humplik, Alexandre Galashov, Leonard Hasenclever, Pedro~A Ortega, Yee~Whye
  Teh, and Nicolas Heess.
\newblock Meta reinforcement learning as task inference.
\newblock \emph{arXiv preprint arXiv:1905.06424}, 2019.

\bibitem[Kalashnikov et~al.(2021)Kalashnikov, Varley, Chebotar, Swanson,
  Jonschkowski, Finn, Levine, and Hausman]{kalashnikov2021mt}
Dmitry Kalashnikov, Jacob Varley, Yevgen Chebotar, Benjamin Swanson, Rico
  Jonschkowski, Chelsea Finn, Sergey Levine, and Karol Hausman.
\newblock Mt-opt: Continuous multi-task robotic reinforcement learning at
  scale.
\newblock \emph{arXiv preprint arXiv:2104.08212}, 2021.

\bibitem[Kingma \& Ba(2015)Kingma and Ba]{kingma2014adam}
Diederik~P Kingma and Jimmy Ba.
\newblock Adam: A method for stochastic optimization.
\newblock In \emph{International Conference on Learning Representations}, 2015.

\bibitem[Kolesnikov et~al.(2020)Kolesnikov, Beyer, Zhai, Puigcerver, Yung,
  Gelly, and Houlsby]{kolesnikov2020big}
Alexander Kolesnikov, Lucas Beyer, Xiaohua Zhai, Joan Puigcerver, Jessica Yung,
  Sylvain Gelly, and Neil Houlsby.
\newblock Big transfer (bit): General visual representation learning.
\newblock In \emph{European Conference on Computer Vision}, 2020.

\bibitem[Kumar et~al.(2020)Kumar, Zhou, Tucker, and
  Levine]{kumar2020conservative}
Aviral Kumar, Aurick Zhou, George Tucker, and Sergey Levine.
\newblock Conservative q-learning for offline reinforcement learning.
\newblock In \emph{Neural Information Processing Systems}, 2020.

\bibitem[Lee et~al.(2019)Lee, Lee, Kim, Kosiorek, Choi, and Teh]{lee2019set}
Juho Lee, Yoonho Lee, Jungtaek Kim, Adam Kosiorek, Seungjin Choi, and Yee~Whye
  Teh.
\newblock Set transformer: A framework for attention-based
  permutation-invariant neural networks.
\newblock In \emph{International Conference on Machine Learning}, 2019.

\bibitem[Lee et~al.(2018)Lee, Sun, Somasundaram, Hu, and Lim]{lee2018composing}
Youngwoon Lee, Shao-Hua Sun, Sriram Somasundaram, Edward~S Hu, and Joseph~J
  Lim.
\newblock Composing complex skills by learning transition policies.
\newblock In \emph{International Conference on Learning Representations}, 2018.

\bibitem[Levine et~al.(2020)Levine, Kumar, Tucker, and Fu]{levine2020offline}
Sergey Levine, Aviral Kumar, George Tucker, and Justin Fu.
\newblock Offline reinforcement learning: Tutorial, review, and perspectives on
  open problems.
\newblock \emph{arXiv preprint arXiv:2005.01643}, 2020.

\bibitem[Liu et~al.(2021)Liu, Raghunathan, Liang, and Finn]{liu2021decoupling}
Evan~Z Liu, Aditi Raghunathan, Percy Liang, and Chelsea Finn.
\newblock Decoupling exploration and exploitation for meta-reinforcement
  learning without sacrifices.
\newblock In \emph{International Conference on Machine Learning}, 2021.

\bibitem[Lynch et~al.(2020)Lynch, Khansari, Xiao, Kumar, Tompson, Levine, and
  Sermanet]{lynch2020learning}
Corey Lynch, Mohi Khansari, Ted Xiao, Vikash Kumar, Jonathan Tompson, Sergey
  Levine, and Pierre Sermanet.
\newblock Learning latent plans from play.
\newblock In \emph{Conference on Robot Learning}, 2020.

\bibitem[Mandlekar et~al.(2018)Mandlekar, Zhu, Garg, Booher, Spero, Tung, Gao,
  Emmons, Gupta, Orbay, Savarese, and Fei-Fei]{mandlekar2018roboturk}
Ajay Mandlekar, Yuke Zhu, Animesh Garg, Jonathan Booher, Max Spero, Albert
  Tung, Julian Gao, John Emmons, Anchit Gupta, Emre Orbay, Silvio Savarese, and
  Li~Fei-Fei.
\newblock Roboturk: A crowdsourcing platform for robotic skill learning through
  imitation.
\newblock In \emph{Conference on Robot Learning}, 2018.

\bibitem[Merel et~al.(2020)Merel, Tunyasuvunakool, Ahuja, Tassa, Hasenclever,
  Pham, Erez, Wayne, and Heess]{merel2019reusable}
Josh Merel, Saran Tunyasuvunakool, Arun Ahuja, Yuval Tassa, Leonard
  Hasenclever, Vu~Pham, Tom Erez, Greg Wayne, and Nicolas Heess.
\newblock Catch \& carry: Reusable neural controllers for vision-guided
  whole-body tasks.
\newblock \emph{ACM Transactions on Graphics}, 2020.

\bibitem[Mitchell et~al.(2021)Mitchell, Rafailov, Peng, Levine, and
  Finn]{mitchell2021offline}
Eric Mitchell, Rafael Rafailov, Xue~Bin Peng, Sergey Levine, and Chelsea Finn.
\newblock Offline meta-reinforcement learning with advantage weighting.
\newblock In \emph{International Conference on Machine Learning}, 2021.

\bibitem[Nagabandi et~al.(2019)Nagabandi, Finn, and Levine]{nagabandi2018deep}
Anusha Nagabandi, Chelsea Finn, and Sergey Levine.
\newblock Deep online learning via meta-learning: Continual adaptation for
  model-based rl.
\newblock In \emph{International Conference on Learning Representations}, 2019.

\bibitem[Pertsch et~al.(2020)Pertsch, Lee, and Lim]{pertsch2020spirl}
Karl Pertsch, Youngwoon Lee, and Joseph~J. Lim.
\newblock Accelerating reinforcement learning with learned skill priors.
\newblock In \emph{Conference on Robot Learning}, 2020.

\bibitem[Pertsch et~al.(2021)Pertsch, Lee, Wu, and Lim]{pertsch2021skild}
Karl Pertsch, Youngwoon Lee, Yue Wu, and Joseph~J. Lim.
\newblock Demonstration-guided reinforcement learning with learned skills.
\newblock In \emph{Conference on Robot Learning}, 2021.

\bibitem[Pong et~al.(2021)Pong, Nair, Smith, Huang, and
  Levine]{pong2021offline}
Vitchyr~H Pong, Ashvin Nair, Laura Smith, Catherine Huang, and Sergey Levine.
\newblock Offline meta-reinforcement learning with online self-supervision.
\newblock \emph{arXiv preprint arXiv:2107.03974}, 2021.

\bibitem[Rakelly et~al.(2019)Rakelly, Zhou, Finn, Levine, and
  Quillen]{rakelly2019efficient}
Kate Rakelly, Aurick Zhou, Chelsea Finn, Sergey Levine, and Deirdre Quillen.
\newblock Efficient off-policy meta-reinforcement learning via probabilistic
  context variables.
\newblock In \emph{International Conference on Machine Learning}, 2019.

\bibitem[Rothfuss et~al.(2019)Rothfuss, Lee, Clavera, Asfour, and
  Abbeel]{rothfuss2018promp}
Jonas Rothfuss, Dennis Lee, Ignasi Clavera, Tamim Asfour, and Pieter Abbeel.
\newblock Pro{MP}: Proximal meta-policy search.
\newblock In \emph{International Conference on Learning Representations}, 2019.

\bibitem[Schaal et~al.(2005)Schaal, Peters, Nakanishi, and
  Ijspeert]{schaal2005motion}
Stefan Schaal, Jan Peters, Jun Nakanishi, and Auke Ijspeert.
\newblock Learning movement primitives.
\newblock In Paolo Dario and Raja Chatila (eds.), \emph{Robotics Research},
  2005.

\bibitem[Schulman et~al.(2017)Schulman, Wolski, Dhariwal, Radford, and
  Klimov]{schulman2017proximal}
John Schulman, Filip Wolski, Prafulla Dhariwal, Alec Radford, and Oleg Klimov.
\newblock Proximal policy optimization algorithms.
\newblock \emph{arXiv preprint arXiv:1707.06347}, 2017.

\bibitem[Sharma et~al.(2020)Sharma, Gu, Levine, Kumar, and
  Hausman]{sharma2019dynamics}
Archit Sharma, Shixiang Gu, Sergey Levine, Vikash Kumar, and Karol Hausman.
\newblock Dynamics-aware unsupervised discovery of skills.
\newblock In \emph{International Conference on Learning Representations}, 2020.

\bibitem[Siegel et~al.(2020)Siegel, Springenberg, Berkenkamp, Abdolmaleki,
  Neunert, Lampe, Hafner, and Riedmiller]{siegel2020keep}
Noah~Y Siegel, Jost~Tobias Springenberg, Felix Berkenkamp, Abbas Abdolmaleki,
  Michael Neunert, Thomas Lampe, Roland Hafner, and Martin Riedmiller.
\newblock Keep doing what worked: Behavioral modelling priors for offline
  reinforcement learning.
\newblock In \emph{International Conference on Learning Representations}, 2020.

\bibitem[Sun(2022)]{sun2022program}
Shao-Hua Sun.
\newblock \emph{Program-Guided Framework for Interpreting and Acquiring Complex
  Skills with Learning Robots}.
\newblock PhD thesis, University of Southern California, 2022.

\bibitem[Teh et~al.(2017)Teh, Bapst, Czarnecki, Quan, Kirkpatrick, Hadsell,
  Heess, and Pascanu]{distral}
Yee Teh, Victor Bapst, Wojciech~M. Czarnecki, John Quan, James Kirkpatrick,
  Raia Hadsell, Nicolas Heess, and Razvan Pascanu.
\newblock Distral: Robust multitask reinforcement learning.
\newblock In \emph{Neural Information Processing Systems}, 2017.

\bibitem[Triantafillou et~al.(2020)Triantafillou, Zhu, Dumoulin, Lamblin, Evci,
  Xu, Goroshin, Gelada, Swersky, Manzagol, et~al.]{triantafillou2019meta}
Eleni Triantafillou, Tyler Zhu, Vincent Dumoulin, Pascal Lamblin, Utku Evci,
  Kelvin Xu, Ross Goroshin, Carles Gelada, Kevin Swersky, Pierre-Antoine
  Manzagol, et~al.
\newblock Meta-dataset: A dataset of datasets for learning to learn from few
  examples.
\newblock In \emph{International Conference on Learning Representations}, 2020.

\bibitem[Van~Hasselt et~al.(2016)Van~Hasselt, Guez, and Silver]{van2016deep}
Hado Van~Hasselt, Arthur Guez, and David Silver.
\newblock Deep reinforcement learning with double q-learning.
\newblock In \emph{AAAI Conference on Artificial Intelligence}, 2016.

\bibitem[Vuorio et~al.(2018)Vuorio, Sun, Hu, and Lim]{vuorio2018toward}
Risto Vuorio, Shao-Hua Sun, Hexiang Hu, and Joseph~J Lim.
\newblock Toward multimodal model-agnostic meta-learning.
\newblock \emph{arXiv preprint arXiv:1812.07172}, 2018.

\bibitem[Vuorio et~al.(2019)Vuorio, Sun, Hu, and Lim]{vuorio2019multimodal}
Risto Vuorio, Shao-Hua Sun, Hexiang Hu, and Joseph~J Lim.
\newblock Multimodal model-agnostic meta-learning via task-aware modulation.
\newblock In \emph{Neural Information Processing Systems}, 2019.

\bibitem[Wang et~al.(2017)Wang, Kurth-Nelson, Tirumala, Soyer, Leibo, Munos,
  Blundell, Kumaran, and Botvinick]{wang2016learning}
Jane~X Wang, Zeb Kurth-Nelson, Dhruva Tirumala, Hubert Soyer, Joel~Z Leibo,
  Remi Munos, Charles Blundell, Dharshan Kumaran, and Matt Botvinick.
\newblock Learning to reinforcement learn.
\newblock In \emph{Annual Meeting of the Cognitive Science Society (CogSci)},
  2017.

\bibitem[Williams(1992)]{williams1992simple}
Ronald~J Williams.
\newblock Simple statistical gradient-following algorithms for connectionist
  reinforcement learning.
\newblock \emph{Machine learning}, 1992.

\bibitem[Yang et~al.(2019)Yang, Caluwaerts, Iscen, Tan, and
  Finn]{yang2019norml}
Yuxiang Yang, Ken Caluwaerts, Atil Iscen, Jie Tan, and Chelsea Finn.
\newblock Norml: No-reward meta learning.
\newblock In \emph{International Conference on Autonomous Agents and Multiagent
  Systems}, 2019.

\bibitem[Yu et~al.(2018)Yu, Finn, Xie, Dasari, Zhang, Abbeel, and
  Levine]{yu2018one}
Tianhe Yu, Chelsea Finn, Annie Xie, Sudeep Dasari, Tianhao Zhang, Pieter
  Abbeel, and Sergey Levine.
\newblock One-shot imitation from observing humans via domain-adaptive
  meta-learning.
\newblock \emph{arXiv preprint arXiv:1802.01557}, 2018.

\bibitem[Yu et~al.(2021)Yu, Kumar, Rafailov, Rajeswaran, Levine, and
  Finn]{yu2021combo}
Tianhe Yu, Aviral Kumar, Rafael Rafailov, Aravind Rajeswaran, Sergey Levine,
  and Chelsea Finn.
\newblock Combo: Conservative offline model-based policy optimization.
\newblock \emph{arXiv preprint arXiv:2102.08363}, 2021.

\bibitem[Zintgraf et~al.(2019)Zintgraf, Shiarli, Kurin, Hofmann, and
  Whiteson]{zintgraf2019fast}
Luisa Zintgraf, Kyriacos Shiarli, Vitaly Kurin, Katja Hofmann, and Shimon
  Whiteson.
\newblock Fast context adaptation via meta-learning.
\newblock In \emph{International Conference on Machine Learning}, 2019.

\bibitem[Zintgraf et~al.(2020)Zintgraf, Shiarlis, Igl, Schulze, Gal, Hofmann,
  and Whiteson]{varibad}
Luisa Zintgraf, Kyriacos Shiarlis, Maximilian Igl, Sebastian Schulze, Yarin
  Gal, Katja Hofmann, and Shimon Whiteson.
\newblock Varibad: A very good method for bayes-adaptive deep rl via
  meta-learning.
\newblock In \emph{International Conference on Learning Representations}, 2020.

\end{thebibliography}
